\pdfoutput=1
\documentclass{article}
\usepackage{amsmath,amsfonts,bm}

\def\eqref#1{equation~\ref{#1}}
\def\1{\bm{1}}

\def\rmO{{\mathbf{O}}}

\def\rmX{{\mathbf{X}}}

\def\ermI{{\textnormal{I}}}

\def\vc{{\bm{c}}}

\def\vg{{\bm{g}}}
\def\vh{{\bm{h}}}

\def\vo{{\bm{o}}}
\def\vp{{\bm{p}}}

\def\vs{{\bm{s}}}

\def\vw{{\bm{w}}}

\def\vy{{\bm{y}}}

\DeclareMathAlphabet{\mathsfit}{\encodingdefault}{\sfdefault}{m}{sl}
\SetMathAlphabet{\mathsfit}{bold}{\encodingdefault}{\sfdefault}{bx}{n}

\def\gB{{\mathcal{B}}}

\def\gD{{\mathcal{D}}}
\def\gE{{\mathcal{E}}}

\def\gG{{\mathcal{G}}}

\def\gL{{\mathcal{L}}}

\def\gN{{\mathcal{N}}}
\def\gO{{\mathcal{O}}}

\def\gV{{\mathcal{V}}}
\def\gW{{\mathcal{W}}}

\def\sR{{\mathbb{R}}}

\DeclareMathOperator*{\argmax}{arg\,max}

\usepackage{microtype}
\usepackage{graphicx}
\usepackage{booktabs} % for professional tables
\usepackage{bbding}
\usepackage{hyperref}
\usepackage{xcolor}
\usepackage{enumitem}
\usepackage{multirow}
\usepackage{colortbl}
\definecolor{Highlight}{rgb}{0.89,0.89,0.94}
\newcommand{\chl}{\cellcolor{Highlight}}
\definecolor{RelativeGain}{HTML}{E0F1F7}
\newcommand{\crg}{\cellcolor{RelativeGain}}
\usepackage[accepted]{icml2021}

\usepackage{fontawesome}
\usepackage{algorithmic}
\usepackage{arydshln}
\usepackage{multirow}
\usepackage{subcaption}

\definecolor{color1}{rgb}{0.1,0.7,0.8} % cyan
\definecolor{color2}{rgb}{0.9,0.1,0.1} % red
\definecolor{color3}{rgb}{0.7,0.3,0.7} % purple
\definecolor{color4}{rgb}{0.3,0.3,0.7} % dark blue
\definecolor{color5}{RGB}{8, 102, 3} % dark green
\definecolor{color6}{rgb}{0.53, 0.66, 0.42} % light green

\newcommand\com[1]{\text{\textcolor{gray}{#1}}}

\definecolor{dark1}{HTML}{315B98}
\definecolor{dark2}{HTML}{7B4399}
\definecolor{dark3}{HSB}{23,175,150}
\newcommand{\graphsgpt}{\textcolor{dark1}{Graphs}\textcolor{dark2}{GPT}}
\newcommand{\encoder}{\textcolor{dark1}{Graph2Seq}}
\newcommand{\decoder}{\textcolor{dark2}{GraphGPT}}

\newcommand{\graphword}{Graph Word}
\newcommand{\graphwords}{Graph Words}
\newcommand{\codebook}{Codebook}

\usepackage{xcolor}
\hypersetup{
    citecolor=black,
    linkcolor=red,  % for internal links (sections, figures, etc.)
    urlcolor=cyan   % for URLs
}

\begin{document}

\twocolumn[
\icmltitle{A Graph is Worth $K$ Words: Euclideanizing Graph using Pure Transformer}



\icmlsetsymbol{equal}{*}

\begin{icmlauthorlist}
  \icmlauthor{Zhangyang Gao}{equal,sch,yyy}
  \icmlauthor{Daize Dong}{equal,sch}
  \icmlauthor{Cheng Tan}{sch,yyy}
  \icmlauthor{Jun Xia}{sch,yyy}
  \icmlauthor{Bozhen Hu}{sch,yyy}
  \icmlauthor{Stan Z. Li}{sch}
\end{icmlauthorlist}

\icmlaffiliation{sch}{Westlake University, Hangzhou, China}
\icmlaffiliation{yyy}{Zhejiang University, Hangzhou, China}

\icmlcorrespondingauthor{Stan Z. Li}{Stan.ZQ.Li@westlake.edu.cn}

\icmlkeywords{Machine Learning, ICML}

\vskip 0.2in
]



\printAffiliationsAndNotice{\icmlEqualContribution} % otherwise use the standard text.

\begin{abstract}
Can we model Non-Euclidean graphs as pure language or even Euclidean vectors while retaining their inherent information? The Non-Euclidean property have posed a long term challenge in graph modeling. Despite recent graph neural networks and graph transformers efforts encoding graphs as Euclidean vectors, recovering the original graph from vectors remains a challenge.
In this paper, we introduce \graphsgpt, featuring an \encoder~encoder that transforms Non-Euclidean graphs into learnable \graphwords~in the Euclidean space, along with a \decoder~decoder that reconstructs the original graph from \graphwords~to ensure information equivalence. We pretrain \graphsgpt~on $100$M molecules and yield some interesting findings:
(1) The pretrained \encoder~excels in graph representation learning, achieving state-of-the-art results on $8/9$ graph classification and regression tasks.
(2) The pretrained \decoder~serves as a strong graph generator, demonstrated by its strong ability to perform both few-shot and conditional graph generation.
(3) \encoder+\decoder~enables effective graph mixup in the Euclidean space, overcoming previously known Non-Euclidean challenges.
(4) The edge-centric pretraining framework \graphsgpt~demonstrates its efficacy in graph domain tasks, excelling in both representation and generation. Code is available at \href{https://github.com/A4Bio/GraphsGPT}{GitHub}.
% (4) The proposed edge-centric \graphsgpt~pretraining framework proves effectiveness in the domain of graphs, marking its success in both representation and generation tasks.
\vspace{-3mm}
\end{abstract}
\vspace{-6mm}
\section{Introduction}

Graphs, inherent to Non-Euclidean data, are extensively applied in scientific fields such as molecular design, social network analysis, recommendation systems, and meshed 3D surfaces \citep{shakibajahromi2024rimeshgnn, zhou2020graph, huang20223dlinker, tan2023target, li2023survey, liu2023gnnrec, xia2022mole, xia2022mole,gao2022pifold, wu2024psc, wu2024mape, tan2023target, gao2022pifold, gao2022simvp, gao2023kw, lin2022conditional}.
% Graphs, as a Non-Euclidean data structure, have extensive applications in scientific domains like molecular design , social network analysis , recommendation systems, and meshed 3D surfaces \citep{shakibajahromi2024rimeshgnn, zhou2020graph, huang20223dlinker, tan2023target, li2023survey, liu2023gnnrec, xia2022mole, xia2022mole,gao2022pifold}.
The Non-Euclidean nature of graphs has inspired sophisticated model designs, including graph neural networks \citep{kipf2016semi,velivckovic2017graph} and graph transformers \citep{ying2021transformers, min2022transformer}.
These models excel in encoding graph structures through attention maps. However, the structural encoding strategies limit the usage of auto-regressive mechanism, thereby hindering pure transformer from revolutionizing graph fields, akin to the success of Vision Transformers (ViT) \citep{dosovitskiy2020image} in computer vision.
We employ pure transformer for graph modeling and address the following open questions:\textit{
\textbf{(1)} How to eliminate the Non-Euclidean nature to facilitate graph representation?
\textbf{(2)} How to generate Non-Euclidean graphs from Euclidean representations? 
\textbf{(3)} Could the combination of graph representation and generation framework benefits from self-supervised pretraining?
}

%%%%% Encoder
We present \encoder, a pure transformer encoder designed to compress the Non-Euclidean graph into a sequence of learnable tokens called \graphwords~in a Euclidean form, where all nodes and edges serve as the inputs and undergo an initial transformation to form \graphwords.
Different from graph transformers \citep{ying2021transformers}, our approach doesn't necessitate explicit encoding of the adjacency matrix and edge features in the attention map. Unlike TokenGT \citep{kim2022pure}, we introduce a \codebook~featuring learnable vectors for graph position encoding, leading to improved training stability and accelerated convergence.
In addition, we employ a random shuffle of the position \codebook,  implicitly augmenting different input orders for the same graph, and offering each position vector the same opportunity of optimization to generalize to larger graphs.

% and allow for generalize to larger graphs.

% ensuring the uniqueness of \graphwords~for identical graphs with different input orders.
% offering each vector the same opportunity for optimization ann generalization to larger graphs not encountered in the training set

%%%%% Decoder
We introduce \decoder, a groundbreaking GPT-style transformer model for graph generation. To recover the Non-Euclidean graph structure, we propose an edge-centric generation strategy that utilizes block-wise causal attention to sequentially generate the graph. Contrary to previous methods \citep{hu2020gpt,shi2019graphaf,peng2022pocket2mol} that generate nodes before predicting edges, the edge-centric technique jointly generates edges and their corresponding endpoint nodes, greatly simplifying the generative space. To align graph generation with language generation, we implement auto-regressive generation using block-wise causal attention, which enables the effective translation of Euclidean representations into Non-Euclidean graph structures.

Leveraging \encoder~encoder and \decoder~decoder, we present \graphsgpt, an integrated end-to-end framework. This framework facilitates a natural self-supervised task to optimize the representation and generation tasks, enabling the transformation between Non-Euclidean and Euclidean data structures.
We pretrain \graphsgpt~on $100$M molecule graphs and comprehensively evaluate it from three perspectives: Encoder, Decoder, and Encoder-Decoder.
The pretrained \encoder~encoder is a strong graph learner for property prediction, outperforming baselines of sophisticated methodologies on $8/9$ molecular classification and regression tasks.
The pretrained \decoder~decoder serves as a powerful structure prior, showcasing both few-shot and conditional generation capabilities.
The \graphsgpt~framework seamlessly connects the Non-Euclidean graph space to the Euclidean vector space while preserving information, facilitating tasks that are known to be challenging in the original graph space, such as graph mixup. The good performance of pretrained \graphsgpt~demonstrates that our edge-centric GPT-style pretraining task offers a simple yet powerful solution for graph learning. In summary, we tame pure transformer to convert Non-Euclidean graph into $K$ learnable \graphwords~, showing the capabilities of \encoder~encoder and \decoder~decoder pretrained through self-supervised tasks, while also paving the way for various Non-Euclidean challenges like graph manipulation and graph mixing in Euclidean latent space.

\section{Related Work}

\paragraph{Graph2Vec.}
Graph2Vec methods create the graph embedding by aggregating node embeddings via graph pooling \cite{lee2019self, ma2019graph, diehl2019edge, ying2018hierarchical}. The node embeddings could be learned by either traditional algorithms \cite{ahmed2013distributed, grover2016node2vec, perozzi2014deepwalk, kipf2016variational, chanpuriya2020infinitewalk, xiao2020vertex}, or deep learning based graph neural networks (GNNs) \cite{kipf2016semi, hamilton2017inductive, wu2019simplifying, chiang2019cluster, chen2018fastgcn, xu2018powerful}, and graph transformers \cite{ying2021transformers, hu2020heterogeneous, dwivedi2020generalization, rampavsek2022recipe, chen2022structure}. These methods are usually designed for specific downstream tasks and can not be used for general pretraining.
\vspace{-3mm}

\paragraph{Graph Transformers.}
The success of extending transformer architectures from natural language processing (NLP) to computer vision (CV) has inspired recent works to apply transformer models in the field of graph learning \cite{ying2021transformers, hu2020heterogeneous, dwivedi2020generalization, rampavsek2022recipe, chen2022structure, wu2021representing, kreuzer2021rethinking, min2022transformer}. To encode the graph prior, these approaches introduce structure-inspired position embeddings and attention mechanisms. For instance, \citet{dwivedi2020generalization,hussain2021edge} adopt Laplacian eigenvectors and SVD vectors of the adjacency matrix as position encoding vectors. \citet{dwivedi2020generalization, mialon2021graphit, ying2021transformers, zhao2021gophormer} enhance the attention computation based on the adjacency matrix. Recently, \citet{kim2022pure} introduced a decoupled position encoding method that empowers the pure transformer as strong graph learner without the needs of expensive computation of eigenvectors and modifications on the attention computation.
\vspace{-3mm}

\paragraph{Graph Self-Supervised Learning.}
The exploration of self-supervised pretext tasks for learning expressive graph representations has garnered significant research interest \citep{wu2021self, liu2022graph, liu2021self, xie2022self}. Contrastive \citep{you2020graph, zeng2021contrastive, qiu2020gcc, zhu2020deep, zhu2021graph, peng2020graph, liu2023hard, liu2023simple, lin2022improving, xia2022simgrace, zou2022multi} and predictive \citep{peng2020self, jin2020self, hou2022graphmae, tian2023heterogeneous, hwang2020self, wang2021self} objectives have been extensively explored, leveraging strategies from the fields of NLP and CV.
However, the discussion around generative pretext tasks \citep{hu2020gpt, zhang2021motif} for graphs is limited, particularly due to the Non-Euclidean nature of graph data, which has led to few instances of pure transformer utilization in graph generation. This paper introduces an innovative approach by framing graph generation as analogous to language generation, thus enabling the use of a pure transformer to generate graphs as a novel self-supervised pretext task.

\begin{figure*}[t]
    \centering
    \includegraphics[width=6.8in]{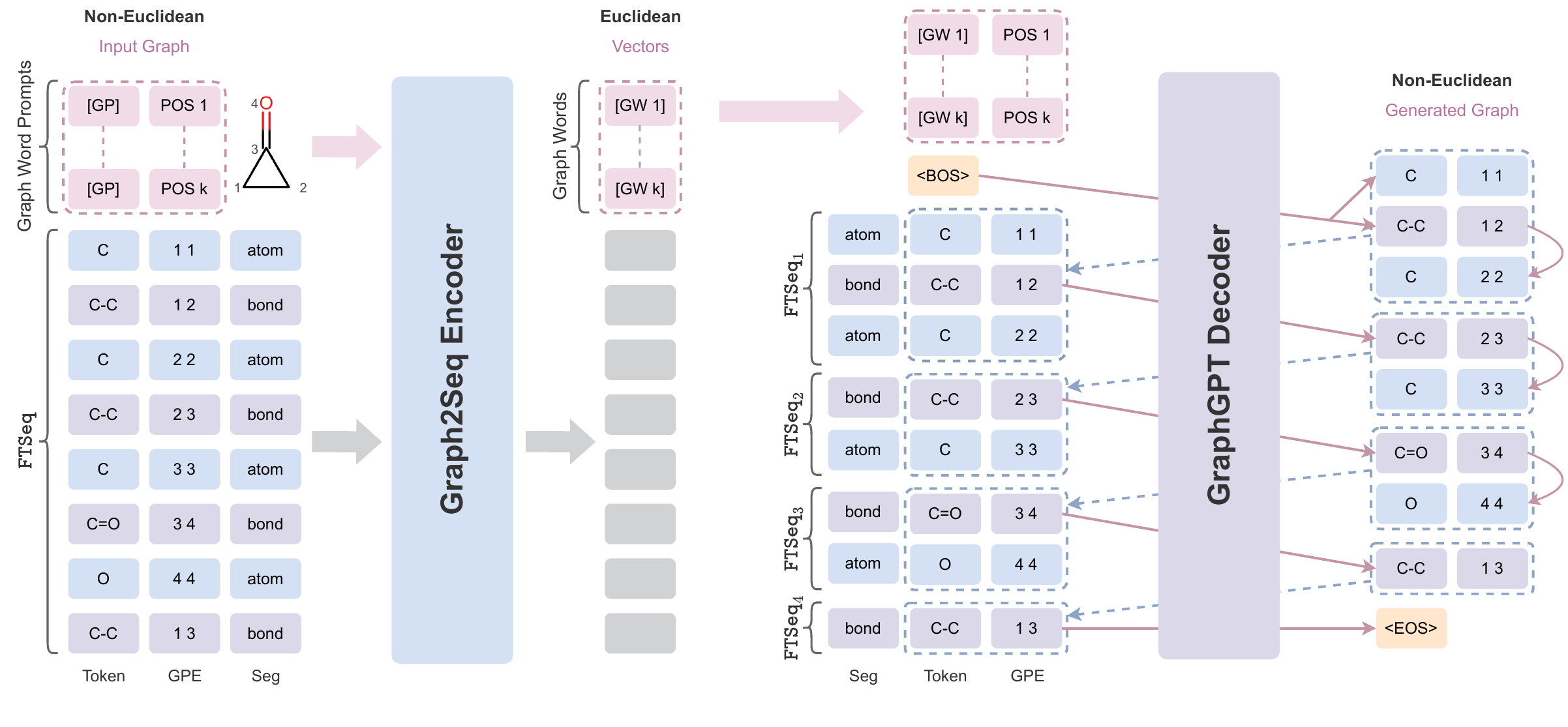}
    \vspace{-5mm}
    \caption{The Overall framework of \graphsgpt. \encoder~encoder transforms the Non-Euclidean graph into Euclidean \graphwords, which are further fed into \decoder~decoder to auto-regressively generate the original Non-Euclidean graph. Both \encoder~and \decoder~employ pure transformer as the structure.} 
    \label{fig:framework}
    \vspace{-3mm}
  \end{figure*}

\vspace{-3mm}
\paragraph{Motivation.} The pure transformer has revolutionized the modeling of texts \citep{devlin2018bert, brown2020language, achiam2023gpt}, images \citep{dosovitskiy2020image, alayrac2022flamingo, dehghani2023scaling, liu2021swin}, and the point cloud \citep{li2023general, yu2022point, pang2022masked} in both representation and generation tasks. However, due to the Non-Euclidean nature, extending transformers to graphs typically necessitates the explicit incorporation of structural information into the attention computation. Such constraint results in following challenges:

\begin{enumerate}
    \vspace{-3mm}
    \item \textbf{Generation Challenge.} When generating new nodes or bonds, the undergone graph structure changes, resulting in a complete update of all graph embeddings from scratch for full attention mechanisms. Moreover, an additional link predictor is required to predict potential edges from a $|\gV| \times |\gV|$ search space.
    \vspace{-1mm}
    \item \textbf{Non-Euclidean Challenge.} Previous methods do not provide Euclidean prototypes to fully describe graphs. The inherent Non-Euclidean nature poses challenges for tasks like graph manipulation and mixing.
    \vspace{-1mm}
    \item \textbf{Representation Challenge.} Limited by the generation challenge, traditional graph self-supervised learning methods have typically focused on reconstructing corrupted sub-features and sub-structures. They overlook of learning from the entire graph potentially limits the ability to capture the global topology.
    
\end{enumerate}
\vspace{-2mm}
To tackle these challenges, we propose \graphsgpt, which uses pure transformer to convert the Non-Euclidean graph into a sequence of Euclidean vectors (\encoder) while ensuring informative equivalence (\decoder). For the first time, we bridge the gap between graph and sequence modeling in both representation and generation tasks.

\vspace{-1mm}
\section{Method}

%%%%%%%%%%%%%%%%%%%%%%%%%%%%%%%%%%%%%
\subsection{Overall Framework}
Figure~\ref{fig:framework} outlines the comprehensive architecture of \textbf{\graphsgpt}, which consists of a \textbf{\encoder}~encoder and a \textbf{\decoder}~decoder. The \encoder~converts Non-Euclidean graphs into a series of learnable feature vectors, named \graphwords. Following this, the \decoder~utilizes these \graphwords~to auto-regressively reconstruct the original Non-Euclidean graph. Both components, the \encoder~and \decoder, incorporate the pure transformer structure and are pretrained via a GPT-style pretext task.

%%%%%%%%%%%%%%%%%%%%%%%%%%%%%%%%%%%%%
% \vspace{-3mm}
\subsection{Graph2Seq Encoder} 
\begin{figure}[H]
   \centering
   \includegraphics[width=3.3in]{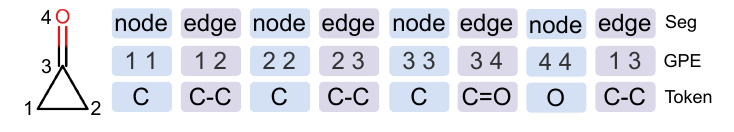}
   \vspace{-6mm}
   \caption{Graph to Flexible Sequence.}
   \vspace{-5mm}
   \label{fig:Graph_to_Flexible_Sequence}
\end{figure}

\vspace{-1mm}
\paragraph{Flexible Token Sequence ($\texttt{FTSeq}$).} Denote $\gG=(\gV, \gE)$ as the input graph, where $\gV=\{v_1, \cdots, v_n\}$ and $\gE=\{e_1,\cdots,e_{n'}\}$ are sets of nodes and edges associated with features $\rmX^{\gV} \in \sR^{n,C}$ and $\rmX^{\gE} \in \sR^{n',C}$, respectively. With a slight abuse of notation, we use $e_i^l$ and $e_i^r$ to represent the left and right endpoint nodes of edge $e_i$. For example, we have $e_1 = (e_1^l, e_1^r)= (v_1, v_2)$ in Figure~\ref{fig:Graph_to_Flexible_Sequence}. Inspired by \cite{kim2022pure}, we flatten the nodes and edges in a graph into a Flexible Token Sequence (\texttt{FTSeq}) consisting of:
% \vspace{-2mm}
\label{sec:permutation_random_shuffle}
\begin{enumerate}
    \vspace{-2mm}
    \item \textbf{Graph Tokens}. The stacked node and edge features are represented by $\rmX = [\rmX^{\gV};\rmX^{\gE}] \in \sR^{n+n',C}$. We utilize a token \codebook~$\gB_t$ to generate node and edge features, incorporating $118+92$ learnable vectors. Specifically, we consider the atom type and bond type, deferring the exploration of other properties, such as the electric charge and chirality, for simplicity. % for future study
    \vspace{-1mm}
    \item \textbf{Graph Position Encodings (GPE)}. The graph structure is implicitly encoded through decoupled position encodings, utilizing a position \codebook~$\gB_p$ comprising $m$ learnable embeddings $\{ \vo_1, \vo_2, \cdots, \vo_m \} \in \sR^{m, d_p}$. The position encodings of node $v_i$ and edge $e_i$ are expressed as $\vg_{v_i} = [\vo_{v_i}, \vo_{v_i}]$ and $\vg_{e_i} = [\vo_{e_i^l}, \vo_{e_i^r}]$, respectively. Notably, $\vg_{v_i}^l = \vg_{v_i}^r = \vo_{v_i}$, $\vg_{e_i}^l = \vo_{e_i^l}$, and $\vg_{e_i}^r = \vo_{e_i^r}$. To learn permutation-invariant features and generalize to larger, unseen graphs, we \textbf{\textit{randomly shuffle}} the position \codebook, giving each vector an equal optimization opportunity.
    \vspace{-1mm}
    \item \textbf{Segment Encodings (Seg)}. We introduce two learnable segment tokens, namely $\texttt{[node]}$ and $\texttt{[edge]}$, to designate the token types within the $\texttt{FTSeq}$.
    \vspace{-1mm}
\end{enumerate}
\vspace{-3mm}
As depicted in Figure~\ref{fig:Graph_to_Flexible_Sequence}, we utilize the Depth-First Search (DFS) algorithm to convert a graph into a flexible token sequence, denoted as $\texttt{FTSeq}=[v_1, e_1, v_2, e_2, v_3, e_3, v_4, e_4]$, where the starting atom matches that in the canonical SMILES. Algorithm~\ref{alg:graph_construction} provides a detailed explanation of our approach. It is crucial to emphasize that the resulting $\texttt{FTSeq}$ remains Non-Euclidean data, as the number of nodes and edges may vary across different graphs.

% \vspace{-3mm}
\label{sec:word_prompts}
\paragraph{Euclidean \graphwords.} Is there a Euclidean representation that can completely describe the Non-Euclidean graph? Given the $\texttt{FTSeq}$ and $k$ graph prompts $[\texttt{[GP]}_1, \texttt{[GP]}_2, \cdots, \texttt{[GP]}_k]$, we use pure transformer to learn a set of \graphwords~$\gW = [\vw_1, \vw_2, \cdots, \vw_k]$:
\begin{equation}
  \gW = \mathrm{Graph2Seq}(\texttt{[GP]}_1, \texttt{[GP]}_2, \cdots, \texttt{[GP]}_k, \texttt{FTSeq}]),
\end{equation}
The token $\texttt{[GP]}_k$ is the sum of a learnable $\texttt{[GP]}$ token and the $k$-th position encoding. The learned \graphwords~$\gW$ are ordered and of fixed length, analogous to a novel graph language created in the latent Euclidean space.
% \vspace{-3mm}

\begin{algorithm}[t]
  \caption{Construction of Flexible Token Sequence}
  \label{alg:graph_construction}
  \begin{algorithmic}[1]
    \REQUIRE
      Canonical SMILES $\texttt{CS}$.
    \ENSURE
      Flexible Token Sequence $\texttt{FTSeq}$.
    \STATE Convert canonical SMILES $\texttt{CS}$ to graph $\gG$.
    \STATE Get the first node $v_1$ in graph $\gG$ by $\texttt{CS}$.
    \STATE Initialize sequence $\texttt{FTSeq}=[v_1]$.
    \FOR{$e_i \operatorname{in} \mathrm{DFS}(\gG,v_1)$}
      \STATE Update sequence $\texttt{FTSeq} \leftarrow [\texttt{FTSeq}, \  e_i]$.
      \IF{$e_i^r \operatorname{not\ in} \texttt{FTSeq}$}
        \STATE Update sequence $\texttt{FTSeq} \leftarrow [\texttt{FTSeq}, \  e_i^r]$.
      \ENDIF
    \ENDFOR{}
  \end{algorithmic}
\end{algorithm}

% \vspace{-3mm}
\paragraph{Graph Vocabulary.} In the context of a molecular system, the complete graph vocabulary for molecules encompasses:
\vspace{-6mm}
\begin{enumerate}
    \item The \graphword~prompts $\texttt{[GP]}$;
    \vspace{-2mm}
    \item Special tokens, including the begin-of-sequence token $\texttt{[BOS]}$, the end-of-sequence token $\texttt{[EOS]}$, and the padding token $\texttt{[PAD]}$;
    \vspace{-2mm}
    \item The dictionary set of atom tokens $\gD_v$ with a size of $\left|\gD_v\right|=118$, where the order of atoms is arranged by their atomic numbers, e.g., $\gD_6$ is the atom C;
    \vspace{-2mm}
    \item The dictionary set of bond tokens $\gD_e$ with a size of $\left|\gD_e\right|=92$, considering the endpoint atom types, e.g., C-C and C-O are different types of bonds even though they are both single bonds.
\end{enumerate}

%%%%%%%%%%%%%%%%%%%%%%%%%%%%%%%%%%%%%
\subsection{GraphGPT Decoder}
\label{sec:decoder}
How to ensure that the learned \graphwords~are information-equivalent to the original Non-Euclidean graph? Previous graph self-supervised learning methods focused on sub-graph generation and multi-view contrasting, which suffer potential information loss due to insufficient capture of the global graph topology. In comparison, we adopt a GPT-style decoder to auto-regressively generate the whole graph from the learned \graphwords~in a edge-centric manner.

\textbf{GraphGPT Formulation.} Given the learned \graphwords~$\gW$ and the flexible token sequence $\texttt{FTSeq}$, the complete data sequence is $[\gW, \texttt{[BOS]}, \texttt{FTSeq}] = [\vw_1, \vw_2, \cdots, \vw_k, \texttt{[BOS]}, v_1, e_1, v_2, \cdots, e_i]$. We define $\texttt{FTSeq}_{1:i}$ as the sub-sequence comprising edges with connected nodes up to $e_i$:
\begin{equation}
\texttt{FTSeq}_{1:i}=
\begin{cases}
    [v_1, e_1,  \cdots, e_i, e_i^r], \ \ \text{if  $e_i^r$ is a new node}\\
    [v_1, e_1,  \cdots, e_i], \ \ \text{otherwise}
\end{cases}.
\end{equation}
In an edge-centric perspective, we assert $e_i^r$ belongs to $e_i$. If $e_i^r$ is a new node, it will be put after $e_i$. Employing \decoder, we auto-regressively generate the complete $\texttt{FTSeq}$ conditioned on $\gW$:
\begin{equation}
      \texttt{FTSeq}_{1:i+1} \xleftarrow{\texttt{FTSeq}_{1:i}} \mathrm{GraphGPT}([\gW, \texttt{[BOS]}, \texttt{FTSeq}_{1:i}]),
\end{equation}
where the notation above the left arrow signifies that the output $\texttt{FTSeq}_{1:i+1}$ corresponds to $\texttt{FTSeq}_{1:i}$.

\paragraph{Edge-Centric Graph Generation.} Nodes and edges are the basic components of a graph. Traditional node-centric graph generation methods divide the problem into two parts: 
\begin{center}
  (1) \textit{Node Generation}; (2) \textit{Link Prediction}.
\end{center}
We argue that node-centric approaches lead to imbalanced difficulties in generating new nodes and edges. For the molecular generation, let $|\gD_v|$ and $|\gD_e|$ denote the number of node and edge types, respectively. Also, let $n$ and $n'$ represent the number of nodes and edges. The step-wise classification complexities for predicting the new node and edge are $\gO(|\gD_v|)$ and $\gO(n \times |\gD_e|)$, respectively. Notably, we observe that $\gO(n \times |\gD_e|) \gg \gO(|\gD_v|)$, indicating a pronounced imbalance in the difficulties of generating nodes and edges. Considering that $\gO(|\gD_v|)$ and $\gO(|\gD_e|)$ are constants, the overall complexity of node-centric graph generation is $\gO(n+ n^2)$.

\begin{figure}[h]
   \centering
   \includegraphics[width=3.2in]{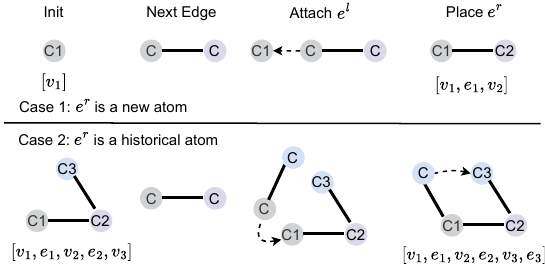}
   \vspace{-2mm}
   \caption{Overview of edge-centric graph generation.}
   \vspace{-4mm}
   \label{fig:generative_pipeline}
\end{figure}

These approaches ignore the basic truism that naturally occurring and chemically valid bonds are sparse: there are only $92$ different bonds (considering the endpoints) among $870$M molecules in the ZINC database \cite{irwin2005zinc}. Given such an observation, we propose the edge-centric generation strategy that decouples the graph generation into:
\begin{center}
  (1) \textit{Edge Generation}; \\ (2) \textit{Left Node Attachment}; (3) \textit{Right Node Placement}.
\end{center}
We provide a brief illustration of the three steps in Figure~\ref{fig:generative_pipeline}. The step-wise classification complexity of generating an edge is $\gO(|\gD_e|)$. Once the edge is obtained, the model automatically infers the left node attachment and right node
placement, relieving the generation from the additional burden of generating atom types and edge connections, resulting in a reduced complexity of $\gO(1)$. With edge-centric generation, we balance the classification complexities of predicting nodes and edge as constants. Notably, the overall generation complexity is reduced to $\gO(n+n')$.

\textit{Next, we introduce the edge-centric generation in detail.}
\paragraph{Step 0: First Node Initialization.} The first node token of $\texttt{FTSeq}$ is generated by:
\begin{equation}
\begin{cases}
    \vh_{v_1} \xleftarrow{\texttt{[BOS]}} \mathrm{GraphGPT}([\gW, \texttt{[BOS]}]) \\
    \vp_{v_1} = \mathrm{Pred}_v(\vh_{v_1}) \\
    v_{1} = \argmax \vp_{v_1} \quad \com{Node Type} \\ % argmax_{2\leq i \leq 119}
    \vg_{v_1} = [\vo_1, \vo_1] \quad \com{GPE}
\end{cases}.
\end{equation}
Here, $\mathrm{Pred}_v(\cdot)$ denotes a linear layer employed for the initial node generation, producing a predictive probability vector $\vp_{v_1} \in \sR^{\left|\gD_v\right|}$. The output $v_1$ corresponds to the predicted node type, and $\vo_1$ represents the node position encoding retrieved from the position \codebook~$\gB_p^\prime$ of the decoder, where we should explicitly note that the encoder \codebook~$\gB_p$ and the decoder \codebook~$\gB_p^\prime$ are not shared.

\paragraph{Step 1: Next Edge Generation.} The edge-centric graph generation method creates the next edge by:
\begin{equation}
\begin{cases}
    \vh_{e_{i+1}} \xleftarrow{e_i} \mathrm{GraphGPT}([\gW, \texttt{[BOS]}, \texttt{FTSeq}_{1:i}])\\
    \vp_{e_{i+1}} = \mathrm{Pred}_e(\vh_{e_{i+1}})\\
    e_{i+1} = \argmax \vp_{e_{i+1}} \quad \com{Edge Type} % argmax_{120\leq i \leq 211}
\end{cases},
\end{equation}
where $\mathrm{Pred}_e$ is a linear layer for the next edge prediction, and $\vp_{e_{i+1}} \in \sR^{\left|\gD_e\right|+1}$ is the predictive probability. $e_{i+1}$ belongs to the set $\gD_e \cup \left\{\texttt{[EOS]}\right\}$, and the generation process will stop if $e_{i+1}=\texttt{[EOS]}$. Note that the edge position encoding $[\vo_{e_{i+1}^l}, \vo_{e_{i+1}^r}]$ remains undetermined. This information will affect the connection of the generated edge to the existing graph, as well as the determination of new atoms, i.e., left atom attachment and right atom placement.

\textbf{Training Token Generation.} The first node and next edge prediction tasks are optimized by the cross entropy loss:
\begin{equation}
    \gL_{\texttt{token}} = -\sum_{i} y_i \cdot \log{p_i}.
\end{equation}

\paragraph{Step 2: Left Node Attachment.} For the newly predicted edge $e_{i+1}$, we further determine how it connects to existing nodes. According to the principles of $\texttt{FTSeq}$ construction, it is required that at least one endpoint of $e_{i+1}$ connects to existing atoms, namely the left atom $e_{i+1}^l$. Given the set of previously generated atoms $\{v_1, v_2, \cdots, v_j\}$ and their corresponding graph position encodings $\rmO_j = [\vo_{v_1}, \vo_{v_2}, \cdots, \vo_{v_j}] \in \sR^{j,C}$ in $\gB_p^\prime$, we predict the position encoding of the left node using a linear layer $\mathrm{PredPos}^l(\cdot)$:
\begin{equation}
    \hat{\vg}_{e_{i+1}}^l = \mathrm{PredPos}^l(\vh_{e_{i+1}}) \in \sR^{1,C}.
\end{equation}
We compute the cosine similarity between $\hat{\vg}_{e_{i+1}}^l$ and $\rmO_j$ by $\vc^l = \hat{\vg}_{e_{i+1}}^l \rmO_j^T \in \sR^t$. The index of existing atoms that $e_{i+1}^l$ will attach to is $u_l = \argmax \vc^l$. This process implicitly infers edge connections by querying over existing atoms, instead of generating all potential edges from scratch. We update the graph position encoding of the left node as:
\begin{equation}
  \vg_{e_{i+1}}^l = \vo_{v_{u_l}} \quad \com{Left Node GPE}.
\end{equation}

\paragraph{Step 3: Right Node Placement.} As for the right node $e_{i+1}^r$, we consider two cases: (1) it connects to one of the existing atoms; (2) it is a new atom. Similar to the step 2, we use a linear layer $\mathrm{PredPos}^r(\cdot)$ to predict the position encoding of the right node:
\begin{equation}
    \hat{\vg}_{e_{i+1}}^r = \mathrm{PredPos}^r(\vh_{e_{i+1}}) \in \sR^{1,C}.
\end{equation}
We get the cosine similarity score $\vc^r = \hat{\vg}_{e_{i+1}}^r \rmO_j^T$ and the index of node with the highest similarity $u_r = \argmax \vc^r$. Given a predefined threshold $\epsilon$, if $\vc_k > \epsilon$, we consider $e_{i+1}$ is connected to $v_{u_r}$, and update:
\begin{equation}
  \vg_{e_{i+1}}^r = \vo_{v_{u_r}} \quad \com{Right Node GPE, Case 1};
\end{equation}
otherwise, $e_{i+1}^r$ is a new atom $v_{j+1}$, and we set:
\begin{equation}
  \vg_{e_{i+1}}^r = \vo_{j+1} \quad \com{Right Node GPE, Case 2}.
\end{equation}
Finally, we update the $\texttt{FTSeq}$ by:
\begin{equation}
  \begin{cases}
      \texttt{FTSeq} \leftarrow  [\texttt{FTSeq}, e_{i+1}] \quad \com{Case 1}\\
      \texttt{FTSeq} \leftarrow [\texttt{FTSeq}, e_{i+1}, v_{j+1}] \quad \com{Case 2}
  \end{cases}.
\end{equation}
By default, we set $\epsilon = 0.5$.
% we randomly choose a new vector from $\{ \vo|  \vo \in \rmO, \vo \notin \rmO_n \}$ and assign it to $\vg_{e_{i+1}}^r$ and $v_{n+1}$

\begin{figure}[H]
   \centering
   \includegraphics[width=2.8in]{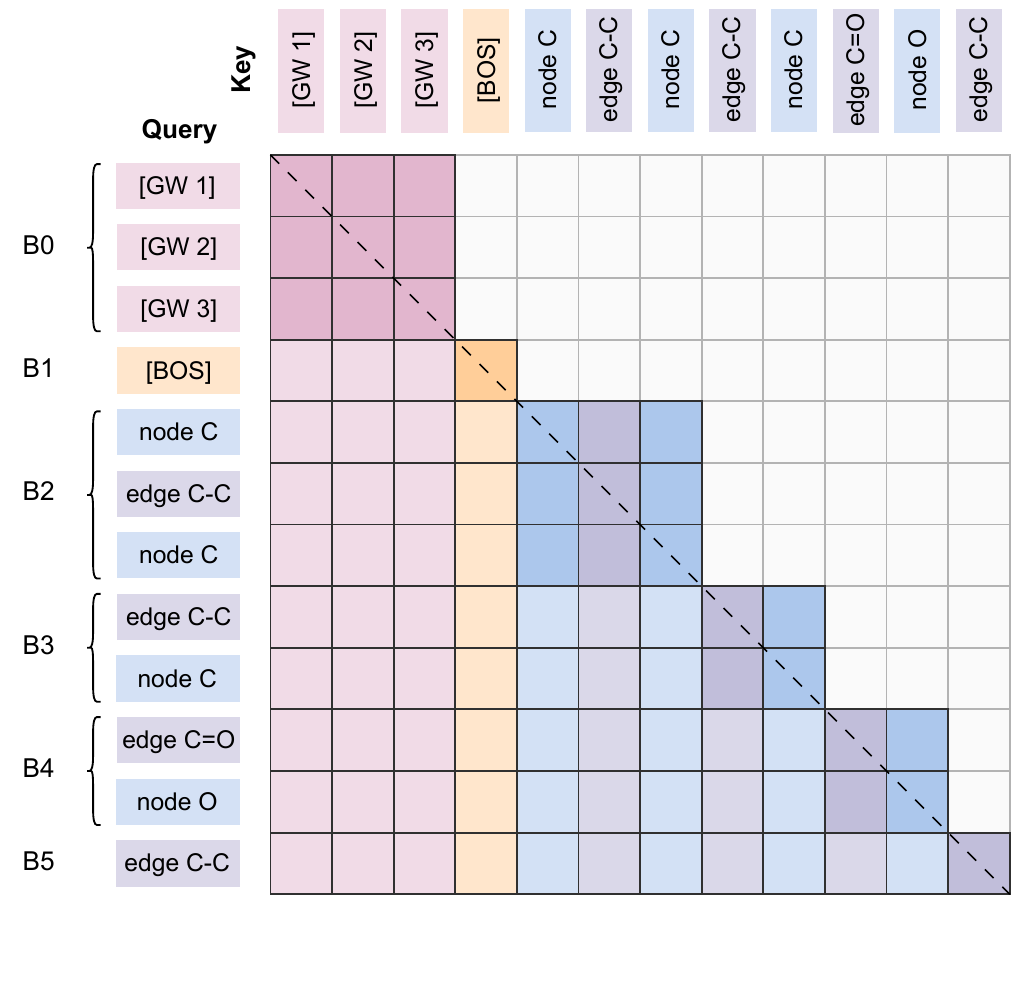}
   \vspace{-4mm}
   \caption{Block-Wise causal attention with grey cells indicating masked positions. \graphwords~contribute to the generation through full attention, serving as prefix prompts.}
   \vspace{-2mm}
   \label{fig:causal_mask}
\end{figure}

\textbf{Training Node Attachment \& Placement.} We adopt a contrastive objective to optimize left node attachment and right node placement problems. Taking left node attachment as an example, given the ground truth $t$, i.e., the index of the attached atom in the original graph, the positive score is $s^+ = e_{i+1}^l \vo_{v_t}^T$, while the negative scores are $\vs^- = |\mathrm{vec}(\rmO \rmO^T)| \in \sR^{|\gB_p^\prime| \times (|\gB_p^\prime|-1)}$, where $\mathrm{vec}(\cdot)$ is a flatten operation while ignoring the diagonal elements. The final contrastive loss is:
\begin{equation}
  \small
    \gL_{\texttt{attach}} = (1-s^+) + \frac{1}{|\gB_p^\prime| \times (|\gB_p^\prime|-1)}\sum s^-.
\end{equation}
\paragraph{Block-Wise Causal Attention.} In our method, node generation is closely entangled with edge generation. Specifically, on its initial occurrence, each node is connected to an edge, creating what we term a block. From the block view, we employ a causal mask for auto-regressive generation. However, within each block, we utilize the full attention. We show the block-wise causal attention in Figure~\ref{fig:causal_mask}.

\section{Experiments}
\subsection{Experiment Settings}
We extensively conduct experiments to assess \graphsgpt, delving into the following questions:
\begin{itemize}[leftmargin=5.5mm]
  \item \textbf{Representation (Q1):}  Can \encoder~effectively learn expressive graph representation through pretraining?
  \item \textbf{Generation (Q2):} Could pretrained \decoder~serve as a strong structural prior model for graph generation?
  \item \textbf{Euclidean \graphwords~(Q3):} What opportunities do the Euclidean \graphwords~offer that were previously considered challenging?
\end{itemize}

\subsection{Datasets}
\paragraph{ZINC (Pretraining).} To pretrain \graphsgpt~, we select the ZINC database \cite{irwin2005zinc} as our pretraining dataset, which contains a total of $870,370,225$ ($870$M) molecules. we randomly shuffle and partition the dataset into training (99.7\%), validation (0.2\%), and test sets (0.1\%). The model does not traverse all the data during pretraining, i.e., a total of about $100$M molecules are used.

\vspace{-1mm}
\paragraph{MoleculeNet (Representation).} \citet{wu2018moleculenet} is a widely-used benchmark dataset for molecular property prediction and drug discovery. It offers a diverse collection of property datasets ranging from quantum mechanics, physical chemistry to biophysics and physiology. Both classification and regression tasks are considered. For rigorous evaluation, we employ standard scaffold splitting, as opposed to random scaffold splitting, for dataset partitioning.

\vspace{-1mm}
\paragraph{MOSES \& ZINC-C (Generation).} For few-shot generation, we evaluate \graphsgpt~on MOSES \cite{polykovskiy2020molecular} dataset, which is designed for benchmarking generative models. Following MOSES, we compute molecular properties (LogP, SA, QED) and scaffolds for molecules collected from ZINC, obtaining ZINC-C. The dataset provides a standardized set of molecules in SMILES format.

\begin{table*}[t]
    \caption{
    Results of molecular property prediction. We report the mean (standard deviation) metrics of 10 runs with standard scaffold splitting (not random scaffold splitting). The \textbf{best} results and the \underline{second best} are highlighted.
    }
    \label{tab:prop_results}
    \setlength{\tabcolsep}{3pt}
    \centering
    \fontsize{7.0pt}{\baselineskip}\selectfont
    \resizebox{1.9 \columnwidth}{!}{
    \begin{tabular}{c l c c c c c c | c c c}
    \toprule
    & & \multicolumn{6}{c|}{ROC-AUC $\uparrow$}  & \multicolumn{3}{c}{RMSD $\downarrow$} \\
    & & Tox21 & ToxCast & Sider & HIV  & BBBP & Bace & ESOL & FreeSolv & Lipo  \\
    \midrule
    & \# Molecules  & 7,831 & 8,575 & 1,427 & 41,127& 2,039 & 1,513 & 1128 & 642 & 4200 \\
    & \# Tasks & 12 & 617 & 27 & 1 & 1 & 1 & 1 & 1 & 1 \\
    \midrule
    \multirow{2}{*}{\rotatebox[origin=c]{90}{No pretrain}}
    & GINs  & 74.6 (0.4) & 61.7 (0.5) &58.2 (1.7)  & 75.5 (0.8) &65.7 (3.3) & 72.4 (3.8) & 1.050 (0.008) & 2.082 (0.082) & 0.683 (0.016) \\
    & \chl Graph2Seq-1W  & \chl 74.0 (0.4)  & \chl 62.6 (0.3) & \chl {66.6 (1.1)}  & \chl 73.6 (3.4) & \chl 68.3 (1.4) & \chl 77.3 (1.2) & \chl 0.953 (0.025) & \chl {1.936 (0.246)} & \chl  0.907 (0.021)\\
    & \crg Relative gain to GIN & \crg -0.8\% & \crg +1.4\% &  \crg+12.6\% & \crg -2.6\% & \crg +3.8\% & \crg +6.3\% & \crg +10.2\% & \crg +7.5\% & \crg -24.7\%\\
    \midrule
    \multirow{20}{*}{\rotatebox[origin=c]{90}{Pretrain}}
    &{InfoGraph}~\citep{sun2020infograph}  & 73.3 (0.6)& 61.8 (0.4)&58.7 (0.6) & 75.4 (4.3) &68.7 (0.6) &74.3 (2.6) & \\
    & {GPT-GNN}~\citep{hu2020gpt-gnn} & 74.9 (0.3) &62.5 (0.4) &58.1 (0.3) &58.3 (5.2)  &64.5 (1.4) &77.9 (3.2) & \\
    & {EdgePred}~\citep{hamilton2017inductive}  &76.0 (0.6)& 64.1 (0.6)& 60.4 (0.7)&64.1 (3.7)   &67.3 (2.4) &77.3 (3.5)  & \\
    & {ContextPred}~\citep{hu2020strategies}  & 73.6 (0.3)& 62.6 (0.6)&59.7 (1.8) &74.0 (3.4)   &70.6 (1.5) &78.8 (1.2) & \\
    & {GraphLoG}~\citep{xu2021self} & 75.0 (0.6)&63.4 (0.6) &59.6 (1.9) &75.7 (2.4)   &68.7 (1.6)&78.6 (1.0)  & \\
    & {G-Contextual}~\citep{rong2020self} & 75.0 (0.6)&62.8 (0.7) &58.7 (1.0) &60.6 (5.2)  &69.9 (2.1) &79.3 (1.1) &\\
    & {G-Motif}~\citep{rong2020self}  & 73.6 (0.7)& 62.3 (0.6)& 61.0 (1.5)& 77.7 (2.7) &66.9 (3.1) &73.0 (3.3) &\\
    & {AD-GCL}~\citep{suresh2021adversarial} & 74.9 (0.4)& 63.4 (0.7)&61.5 (0.9) &77.2 (2.7)   &70.7 (0.3) & 76.6 (1.5)  & \\
    & {JOAO}~\citep{you2021graph}  & 74.8 (0.6)& 62.8 (0.7)&60.4 (1.5) &66.6 (3.1)   &66.4 (1.0) &73.2 (1.6) & 1.120 (0.003) & & 0.708 (0.004)\\
    & {SimGRACE}~\citep{xia2022simgrace}  & 74.4 (0.3)&62.6 (0.7) &60.2 (0.9) &75.5 (2.0)   & 71.2 (1.1)&74.9 (2.0) &\\
    & {GraphCL}~\citep{you2020graph} &75.1 (0.7)& 63.0 (0.4)& 59.8 (1.3)& 77.5 (3.8) & 67.8 (2.4)& 74.6 (2.1)& 0.947 (0.038) & {2.233} (0.261) & 0.739 (0.009)\\
    & {GraphMAE}~\citep{hou2022graphmae}  & 75.2 (0.9) &63.6 (0.3) &60.5 (1.2) &76.5 (3.0)   & {71.2} (1.0) &78.2 (1.5)  & \\
    & 3D InfoMax~\citep{stark20223d} &74.5 (0.7) & 63.5 (0.8)& 56.8 (2.1)& 62.7 (3.3)  &69.1 (1.2) &78.6 (1.9) & \underline{0.894} (0.028) & 2.337 (0.227) & 0.695 (0.012) \\
    & {GraphMVP}~\citep{liu2022pretraining}  &74.9 (0.8)& 63.1 (0.2) &60.2 (1.1)& \underline{79.1} (2.8) & 70.8 (0.5) & {79.3} (1.5)& 1.029 (0.033) & & \underline{0.681} (0.010)\\
    & {MGSSL}~\citep{zhang2021motif} &75.2 (0.6)& 63.3 (0.5)& {61.6} (1.0)& 77.1 (4.5) & 68.8 (0.6)& 78.8 (0.9)&\\
    & {AttrMask}~\citep{hu2020strategies} & 75.1 (0.9) &63.3 (0.6) &60.5 (0.9)& 73.5 (4.3)   &65.2 (1.4) &77.8 (1.8) & 1.100 (0.006) & 2.764 (0.002) & 0.739 (0.003)\\  
    & MolCLR ~\citep{wang2022molecular} & 75.0 (0.2) &  & 58.9 (1.4) & 78.1 (0.5) & \underline{72.2} (2.1) & {82.4} (0.9) & 1.271 (0.040) & 2.594 (0.249) & 0.691 (0.004) \\
    & Graphformer ~\citep{rong2020self}  & 74.3 (0.1) & \textbf{65.4} (0.4) & \underline{64.8} (0.6) & 62.5 (0.9) & 70.0 (0.1) & \underline{82.6} (0.7) & 0.983 (0.090) & \underline{2.176} (0.052) & 0.817 (0.008)\\
    & Mole-BERT ~\citep{xia2023mole} & \underline{76.8} (0.5) & \underline{64.3} (0.2) & {62.8} (1.1)& {78.9} (3.0) & {71.9} (1.6)& {80.8} (1.4)& 1.015 (0.030) &  & \textbf{0.676} (0.017) \\  
    & \crg Relative gain to GIN & \crg +2.9\% & \crg +6.0\% & \crg +11.3\% & \crg +4.8\% & \crg +9.9\% & \crg +14.1\% & \crg +14.9\% & \crg -4.5\% & \crg +1.0\%\\ \midrule
    \multirow{3}{*}{\rotatebox[origin=c]{90}{Pretrain}} 
    & \chl Graph2Seq-1W  & \chl  \textbf{76.9} (0.3)  & \chl  \textbf{65.4} (0.5)  & \chl  \textbf{68.2} (0.9)    & \chl  \textbf{79.4} (3.9) & \chl  \textbf{72.8} (1.5) &  \chl  \textbf{83.4} (1.0) & \chl  \textbf{0.860} (0.024)  & \chl  \textbf{1.797} (0.237) & \chl  0.716 (0.019) \\
    & \crg Relative gain to GIN & \crg +3.1\% & \crg +6.0\% & \crg +17.2\% & \crg +5.2\% & \crg +10.8\% & \crg +15.2\% & \crg +18.1\% & \crg +13.7\% & \crg -4.8\%\\
    & \crg Relative gain to Graph2Seq-1W & \crg +3.9\% & \crg +4.5\% & \crg +2.4\% & \crg +7.9\% & \crg +6.6\% & \crg +7.9\% & \crg +9.8\% & \crg +7.2\% & \crg +21.1\%\\
    % Graph2Seq~(mixup) &  \textbf{77.2}  & \textbf{65.5}  & \textbf{68.9}    & \textbf{79.8} &  \textbf{73.4} & \textbf{85.4} & \textbf{0.843} & \textbf{1.795} & 0.689 \\
    \bottomrule
  \end{tabular}}
  \vspace{-4mm}
  \end{table*}
\vspace{-1mm}
\subsection{Pretraining}
\paragraph{Model Configurations.}
We adopt the transformer as our model structure. Both the \encoder~encoder and the \decoder~decoder consist of $8$ transformer blocks with $8$ attention heads. For all layers, we use Swish \cite{ramachandran2017searching} as the activation function and RMSNorm \cite{zhang2019root} as the normalizing function. The hidden size is set to $512$, and the length of the Graph Position Encoding (GPE) is $128$. The total number parameters of the model is $50$M. Denote $K$ as the number of \graphwords, multiple versions of \graphsgpt, referred to as \graphsgpt-$K$W, were pretrained. We mainly use \graphsgpt-$1$W, while we find that \graphsgpt-$8$W has better encoding-decoding consistency (Section~\ref{sec:rebuttal}, Q2).

\vspace{-1mm}
\paragraph{Training Details.}
The \graphsgpt~model undergoes training for $100$K steps with a global batch size of $1024$ on 8 NVIDIA-A100s, utilizing AdamW optimizer with $0.1$ weight decay, where $\beta_1=0.9$ and $\beta_2=0.95$. The maximum learning rate is $1e^{-4}$ with $5$K warmup steps, and the final learning rate decays to $1e^{-5}$ with cosine scheduling.

\vspace{-1mm}
\subsection{Representation}
\textit{Can \encoder~effectively learn expressive graph representation through pretraining?}

\paragraph{Setting \& Baselines.} We finetune the pretrained \encoder-1W on the MoleculeNet dataset. The learned \graphwords~are input into a linear layer for graph classification or regression. We adhere to standard scaffold splitting (not random scaffold splitting) for rigorous and meaningful comparison. We do not incorporate the 3D structure of molecules for modeling. Recent strong molecular graph pretraining baselines are considered for comparison. 

\textit{We show property prediction results in Table~\ref{tab:prop_results}, finding that:}

\textbf{Pure Transformer is Competitive to GNN.} Without pretraining, \encoder-1W demonstrates a comparable performance to GNN. Specifically, in 4 out of 9 cases, \encoder-1W outperforms GIN with gains exceeding $5\%$, and in another 4 out of 9 cases, it achieves similar performance with an absolute relative gain of less than $5\%$. In addition, pure transformer runs much faster than GNNs, i.e., we finish the pretraining of \graphsgpt~ within 6 hours using 8 A100.

\textbf{GPT-Style Pretraining is All You Need.} Pretrained \encoder~demonstrates a non-trivial improvement over 8 out of 9 datasets when compared to baselines. These results are achieved without employing complex pretraining strategies such as multi-pretext combination and hard-negative sampling, highlighting that GPT-pretraining alone is sufficient for achieving SOTA performance and providing a simple yet effective solution for graph SSL.

\textbf{Graph2Seq Benefits More from GPT-Style Pretraining.} The non-trivial improvement has not been observed by previous GPT-GNN \citep{hu2020gpt-gnn}, which adopts a node-centric generation strategy and GNN architectures. This suggests that the transformer model is more suitable for scaling to large datasets. In addition, previous pretrained transformers without the GPT-style pretraining \citep{rong2020self} perform worse than \encoder. This underscores that generating the entire graph enhances the learning of global topology and results in more expressive representations.
% In summary, pretraining \encoder~is simple, yet the obtained model is quite effective.

\subsection{Generation}
\label{sec:fewshot_generation}
\textit{Could pretrained \decoder~serve as a strong structural prior model for graph generation?}

\paragraph{GraphGPT Generates Novel Molecules with High Validity.} We assess pretrained \decoder-1W on the MOSES dataset through few-shots generation without finetuning. By extracting \graphword~embeddings $\{\vh_i\}_{i=1}^M$ from $M$ training molecules, we construct a mixture Gaussian distribution $p(\vh, s) = \sum_{i=1}^{M}{\gN(\vh_i, s \ermI)}$, where $s$ is the standard variance. We sample $M$ molecules from $p(\vh, s)$ and report the validity, uniqueness, novelty and IntDiv in Table \ref{tab:fewshot_moses}. We observe that \decoder~generates novel molecules with high validity. Without any finetuning, \decoder~outperforms MolGPT on validity, uniqueness, novelty, and diversity. Definition of metrics could be found in the Appendix \ref{app:metric}.

\begin{table}[htbp]
    \caption{Few-shot generation results of \decoder-1W. We use $M=100$K shots and sample the same number of \graphword~embeddings under different variance $s$.}
    \resizebox{1.0 \columnwidth}{!}{
    \begin{tabular}{clccccc}  % $\downarrow$
        \toprule
         & \textbf{Model} & Validity $\uparrow$ & Unique $\uparrow$ & Novelty $\uparrow$ & IntDiv$_1$ $\uparrow$ & IntDiv$_2$ $\uparrow$ \\
        \midrule
        \multirow{9}{*}{\rotatebox[origin=c]{90}{Unconditional}}
         & HMM & 0.076 & 0.567 & \textbf{0.999} & 0.847 & 0.810 \\
         & NGram & 0.238 & 0.922 & 0.969 & 0.874 & 0.864 \\
         & Combinatorial & \textbf{1.0} & 0.991 & 0.988 & 0.873 & 0.867 \\
         & CharRNN & 0.975 & 0.999 & 0.842 & 0.856 & 0.850 \\
         & VAE & 0.977 & 0.998 & 0.695 & 0.856 & 0.850 \\
         & AEE & 0.937 & 0.997 & 0.793 & 0.856 & 0.850 \\
         & LatentGAN & 0.897 & 0.997 & 0.949 & 0.857 & 0.850 \\
         & JT-VAE & \textbf{1.0} & 0.999 & 0.914 & 0.855 & 0.849 \\
         & MolGPT & 0.994 & \textbf{1.0} & 0.797 & 0.857 & 0.851 \\
        \midrule
        \multirow{4}{*}{\rotatebox[origin=c]{90}{Few Shot}}
         % & \decoder-1W$_{100\text{K}}$ & 1.0 & 0.978 & 0.997 & 0.871 & 0.860  \\
         % & \decoder-1W$_{300\text{K}}$ & 1.0 & 0.978 & 0.998 & 0.863 & 0.858  \\
         & \chl GraphGPT-1W$_{s=0.25}$ &\chl \textbf{0.995} &\chl 0.995 &\chl 0.255 &\chl 0.854 &\chl 0.850 \\
         & \chl GraphGPT-1W$_{s=0.5}$ &\chl 0.993 &\chl 0.996 &\chl 0.334 &\chl 0.856 & \chl0.848 \\
         & \chl GraphGPT-1W$_{s=1.0}$ &\chl 0.978 &\chl 0.997 &\chl 0.871 &\chl 0.860 &\chl 0.857 \\
         & \chl GraphGPT-1W$_{s=2.0}$ &\chl 0.972 &\chl \textbf{1.0} &\chl \textbf{1.0} &\chl 0.850 &\chl 0.847\\
        \bottomrule
    \end{tabular}}
    \label{tab:fewshot_moses}
    \vspace{-3mm}
\end{table}

\paragraph{GraphGPT-C is a Controllable Molecule Generator.} Following \citep{bagal2021molgpt}, we finetune \graphsgpt-1W on $100$M molecules from ZINC-C with properties and scaffolds as prefix inputs, obtaining \graphsgpt-1W-C. We access whether the model could generate molecules satisfying specified properties. We present summarized results in Figure \ref{fig:condition_result} and Table \ref{tab:condition_molgpt}, while providing the full results in the appendix due to space limit. The evaluation is conducted using the scaffold ``c1ccccc1'', demonstrating that \decoder~can effectively control the properties of generated molecules. Table \ref{tab:condition_molgpt} further confirms that unsupervised pretraining enhances the controllability and validity of \decoder. More details can be found in Appendix \ref{app:cond_gen}.

\begin{figure}[htbp]
    \centering
    \begin{subfigure}{0.49\linewidth}
        \centering
        \includegraphics[width=1.0\linewidth]{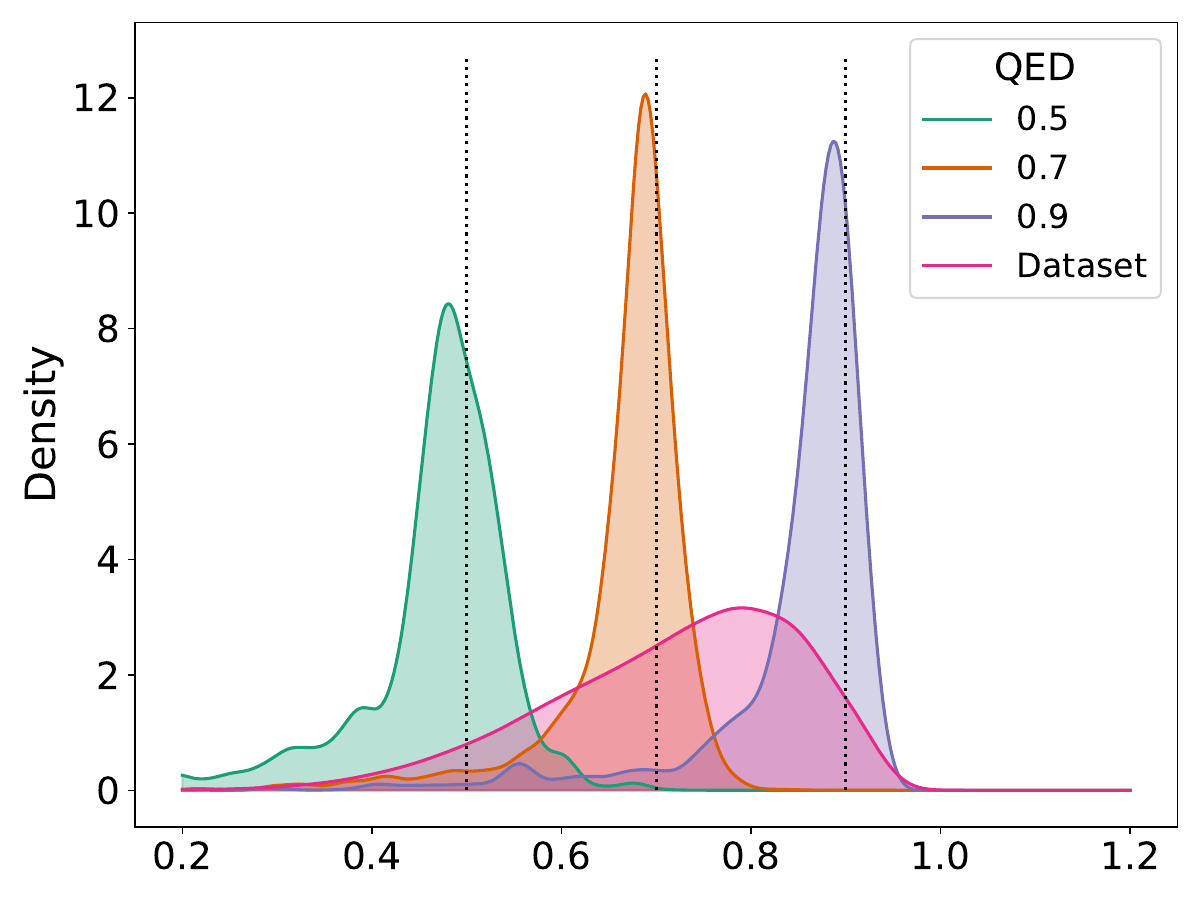}
        \caption{QED}
        \label{fig:condition_qed}
    \end{subfigure}
    \hfill
    \begin{subfigure}{0.49\linewidth}
        \centering
        \includegraphics[width=1.0\linewidth]{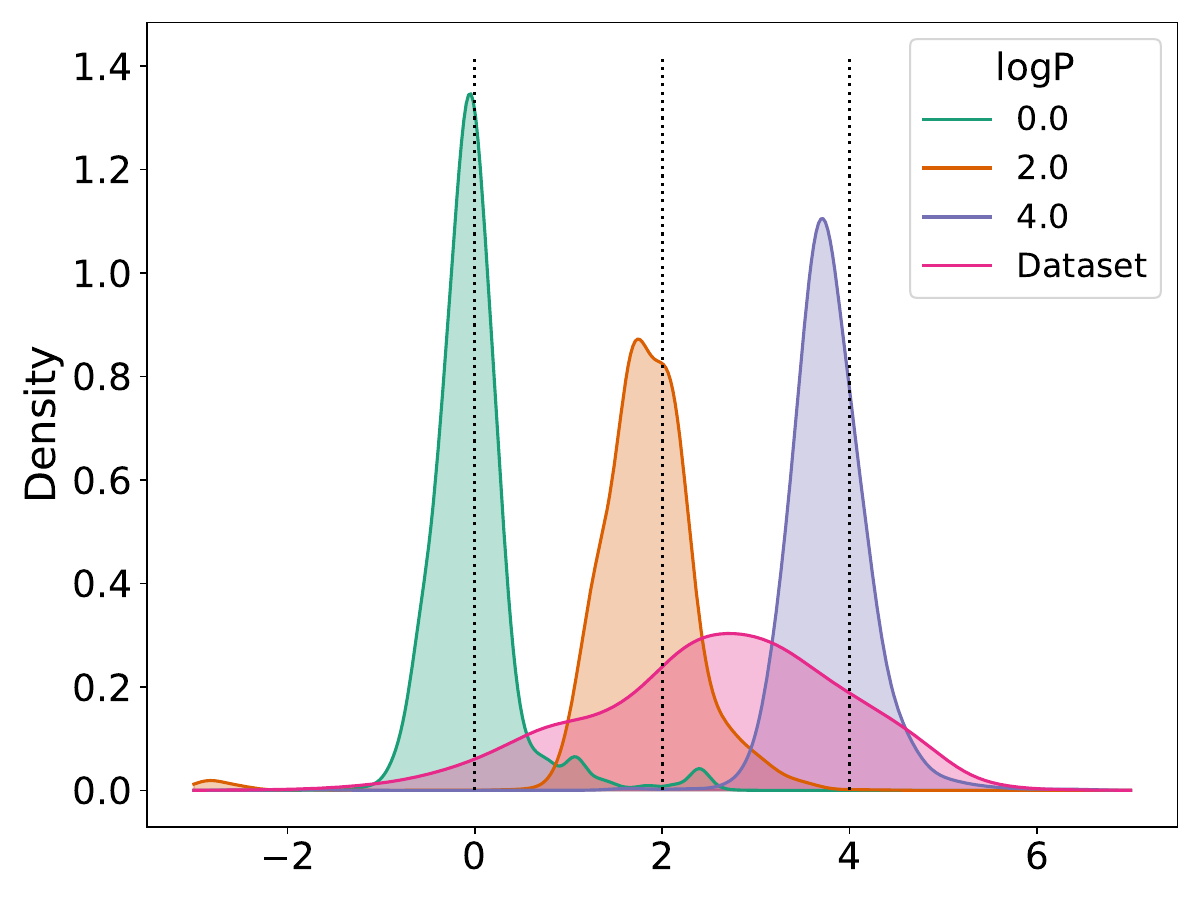}
        \caption{logP}
        \label{fig:condition_logp}
    \end{subfigure}
    \vspace{-3mm}
    \caption{Property distribution of generated molecules on different conditions using \graphsgpt-1W-C.  ``Dataset" denotes the distribution of the training dataset (ZINC-C).}
    \label{fig:condition_result}
    \vspace{-3mm}
\end{figure}

\begin{table}[h]
    \caption{Comparison with MolGPT on different properties. ``MAD" denotes the Mean Absolute Deviation in generated molecule properties compared to the oracle value. ``SD" denotes the Standard Deviation of the generated property.}
    \centering
    \resizebox{0.8 \columnwidth}{!}{
    \begin{tabular}{cclccccc}
        \toprule
         & Pretrain & Metric & QED=0.5 & SA=0.7 & logP=0.0 & \ \ \underline{Avg.}\ \ & \\
         \midrule
        \multirow{3}{*}{\rotatebox[origin=c]{90}{\small{MolGPT}}} &
         & MAD $\downarrow$ & 0.081 & 0.024 & 0.304 & \underline{0.136} & \\
         & \XSolidBrush & SD $\downarrow$ & \textbf{0.065} & \textbf{0.022} & \textbf{0.295} & \underline{\textbf{0.127}} & \\
         & & Validity $\uparrow$ & 0.985 & 0.975 & 0.982 & \underline{0.981} & \\
         \midrule
        \multirow{6}{*}{\rotatebox[origin=c]{90}{\small{GraphGPT-1W-C}}} &
         \chl & \chl MAD $\downarrow$ & \chl 0.041 & \chl 0.012 & \chl 0.103 & \chl \underline{0.052} & \chl \\
         & \chl \XSolidBrush & \chl SD $\downarrow$ &\chl 0.079 & \chl 0.055 &\chl 0.460 & \chl \underline{0.198} & \chl \\
         & \chl & \chl Validity $\uparrow$ &\chl 0.988 & \chl \textbf{0.995} &\chl 0.980 & \chl \underline{0.988} & \chl \\
         \cdashline{2-8}
         & \crg & \crg MAD $\downarrow$ &\crg \textbf{0.032} & \crg \textbf{0.002} & \crg \textbf{0.017} & \crg \underline{\textbf{0.017}} & \crg \\
         & \crg \CheckmarkBold & \crg SD $\downarrow$ &\crg 0.080 & \crg 0.042 & \crg 0.404 & \crg \underline{0.175} & \crg \\
         & \crg & \crg Validity $\uparrow$ &\crg \textbf{0.996} & \crg \textbf{0.995} & \crg \textbf{0.994} &\crg \underline{\textbf{0.995}} & \crg \\
        \bottomrule
    \end{tabular}}
    \label{tab:condition_molgpt}
    \vspace{-5mm}
\end{table}

\subsection{Euclidean \graphwords}
\textit{What opportunities do the Euclidean \graphwords~offer that were previously considered challenging?}

% \paragraph{Graph Mixing.}
For graph classification, let the $i$-th sample be denoted as $(\gG_i, \vy_i)$, where $\gG_i$ and $\vy_i$ represent the graph and one-hot label, respectively. When considering paired graphs $(\gG_i, \vy_i)$ and $(\gG_j, \vy_j)$, and employing a mixing ratio $\lambda$ sampled from the $Beta(\alpha, \alpha)$ distribution, the mixed label is defined as $\vy_{mix} = \lambda \vy_i + (1-\lambda) \vy_j$. However, due to the irregular, unaligned, and Non-Euclidean nature of graph data, applying mixup to get $\gG_{mix}$ is nontrivial. Recent efforts \citep{zhou2020data, park2022graph, wu2022graphmixup, zhang2023mixupexplainer, guo2023interpolating} have attempted to address this challenge by introducing complex hand-crafted rules. Additionally, $\gG$-mixup \citep{han2022g} leverages estimated graphons for generating mixed graphs. To our best knowledge, there are currently no learnable model for mixing in Euclidean space while generating new graphs. 

\begin{table}[h]
    \centering
    \vspace{-3mm}
    \caption{Graph mixup results. We compare \encoder~with $\gG$-mixup on multiple tasks from MoleculeNet.}
    \resizebox{0.8 \columnwidth}{!}{
    \begin{tabular}{cccccccc}
        \toprule
        & mixup & HIV $\uparrow$  & BBBP $\uparrow$ & Bace $\uparrow$ & Tox21 $\uparrow$ & ToxCast $\uparrow$ & Sider $\uparrow$ \\
        \midrule
        \multirow{3}{*}{\rotatebox[origin=c]{90}{G-Mix}}   & \XSolidBrush    & 77.1 & 68.4 & 75.9 &       &         &       \\
        & \CheckmarkBold   & 77.1 & 70.2 & 77.8 &       &         &       \\
        & gain  & +0.0 & +1.8 & +1.9 &       &         &       \\ \midrule
        \multirow{3}{*}{\rotatebox[origin=c]{90}{Ours}} & \XSolidBrush    & 79.4 & 72.8 & 83.4 & 76.9  & 65.4    & 68.2  \\
        &\chl \CheckmarkBold   &\chl 79.8 &\chl 73.4 &\chl 85.4 &\chl 77.2  &\chl 65.5    &\chl 68.9  \\
        & \crg  gain  & \crg  +0.4 &\crg  +0.6 &\crg  +2.0 &\crg  +0.3  &\crg  +0.1    &\crg  +0.7 \\ \bottomrule
    \end{tabular}}
    \vspace{-3mm}
    \label{tab:mixup}
\end{table}

\textbf{GraphsGPT~is a Competitive Graph Mixer.} We mixup the learned \graphwords~encoded by \encoder-1W, then generate the mixed graph using \decoder-1W. Formally, the \graphwords~of $\gG_i$ and $\gG_j$ are $\gW_i = \mathrm{Graph2Seq}(\gG_i)$ and $\gW_j = \mathrm{Graph2Seq}(\gG_j)$, and the mixed graph is $\gG_{mix} = \mathrm{GraphGPT}(\lambda \gW_i + (1-\lambda) \gW_j)$. We conduct experiments on MoleculeNet and show the results in Table \ref{tab:mixup}. We observe that the straightforward latent mixup outperforms the elaborately designed $\gG$-mixup proposed in the ICML'22 outstanding paper \citep{han2022g}. 

Due to page limit, more results are moved to the appendix.

% about clustering, decoupled representation, latent graph optimization, and graph mutation

\section{Conclusion}
We propose \graphsgpt, the first framework with pure transformer that converts Non-Euclidean graph into Euclidean representations, while preserving information using an edge-centric GPT-style pretraining task. We show that the \encoder~and \decoder~serve as strong graph learners for representation and generation, respectively. The Euclidean representations offer more opportunities previously known to be challenging. The \graphsgpt~may create a new paradigm of graph modeling.

% \section{Broader Impact}
% The proposed framework provides a new paradigm for graph representation, generation and manipulation. The Non-Euclidean to Euclidean transformation may affect broader downstream graph applications, such as graph translation and optimization. The methodology could be extend to other modalities, such as image and sequence.

% \clearpage

\section{Rebuttal Details}
\label{sec:rebuttal}

\paragraph{Q1} Missing discussion on diffusion-based molecular generative models.
\vspace{-2mm}
\paragraph{R1} We conduct additional experiments following \citep{kong2023autoregressive} to compare \decoder-1W with the diffusion-based methods on ZINC-250K. We follow the same few-shots generation setting described in the Section \ref{sec:fewshot_generation}, where we set $M=10$K for fair comparison. As shown in Table \ref{tab:zinc250k}, we find that \decoder-1W~surpasses these methods in a large margin on various metrics, which can further validate the strong generation ability of \decoder. 

\begin{table}[h]
    \caption{Comparison with diffusion-based methods on ZINC-250K. We use $M=10$K shots and sample the same number of \graphword~under different variance $s$.}
    \resizebox{1.0 \columnwidth}{!}{
    \begin{tabular}{lccccc} \toprule
    Model              & Valid $\uparrow$     & Unique $\uparrow$ & Novel $\uparrow$ & NSPDK $\downarrow$     & FCD $\downarrow$   \\ \hline
    GraphAF  \citep{shi2020graphaf}          & 68.47                & 98.64             & \textbf{100} & 0.044                  & 16.02              \\ 
    GraphDF  \citep{luo2021graphdf}          & 90.61                & 99.63             & \textbf{100} & 0.177                  & 33.55              \\ 
    MoFlow  \citep{zang2020moflow}           & 63.11                & 99.99             & \textbf{100} & 0.046                  & 20.93              \\ 
    EDP-GNN  \citep{niu2020permutation}          & 82.97                & 99.79             & \textbf{100} & 0.049                  & 16.74              \\ 
    GraphEBM  \citep{liu2021graphebm}         & 5.29                 & 98.79             & \textbf{100} & 0.212                  & 35.47              \\ 
    SPECTRE \citep{martinkus2022spectre}           & 90.20                & 67.05             & \textbf{100} & 0.109                  & 18.44              \\ 
    GDSS  \citep{jo2022score}             & 97.01                & 99.64             & \textbf{100} & 0.019                  & 14.66              \\ 
    DiGress \citep{vignac2022digress}           & 91.02                & 81.23             & \textbf{100} & 0.082                  & 23.06              \\ 
    GRAPHARM  \citep{kong2023autoregressive}         & 88.23                & 99.46             & \textbf{100} & 0.055                  & 16.26              \\ \hline
    GraphGPT-1W$_{s=0.25}$ & \textbf{99.67} & 99.95             & 93.0             & \textbf{0.0002} & \textbf{1.78} \\ 
    GraphGPT-1W$_{s=0.5}$  & 99.57                & 99.97             & 93.6             & 0.0003                 & 1.79               \\ 
    GraphGPT-1W$_{s=1.0}$  & 98.44                & \textbf{100}  & 98.0             & 0.0012                 & 2.89               \\ 
    GraphGPT-1W$_{s=2.0}$  & 97.64                & \textbf{100}  & \textbf{100} & 0.0056                 & 8.47      \\ \bottomrule        
    \end{tabular}
    }
    \label{tab:zinc250k}
    \vspace{-4mm}
\end{table}

\paragraph{Q2} How do the method consider the symmetry of graphs? Graph data is invariant to permutation.
\vspace{-2mm}
\paragraph{R2}  In Section \ref{sec:permutation_random_shuffle}, we mention that ``we introduce a random shuffle of the position \codebook''. We should explicitly state that this random shuffle of position vectors is equivalent to randomly shuffling the input order of atoms. This allows the model to learn from the data with random order augmentation. We point that building a permutation-invariant encoder is easy and necessary, however, developing a decoder with permutation invariance poses a significant challenge for auto-regressive generation models. We randomly shuffle the position vectors, allowing the model to learn representations with different orders for molecules.

To further verify the effectiveness of our method in handling the permutation invariance, we conduct an additional experiment. Given an input molecular graph sequence, we randomly permute its order for $1024$ times and encode the shuffled sequences with \encoder, obtaining a set of $1024$ \graphwords. We then decode them back to the graph sequences and observe the consistency, which is defined as the maximum percentage of the decoded sequences that share the same results. Table \ref{tab:consistency} shows the results on $1000$ molecules from the test set, where we find both models are resistant to a certain degree of permutation invariance, i.e., 96.1\% and 97.9\% of the average consistency for \graphsgpt-1W and \graphsgpt-8W, respectively.

In addition, there is a contradiction between permutation-invariant model and auto-regressive model. Previous work (TokenGT \citep{kim2022pure}) focuses on representation learning, therefore, do not suffer from the issue of permutation-invariant. We combine representation with generation tasks in the same model, and propose the technique of randomly shuffling position vectors so that all tasks can work well. We should note that randomly shuffling the position vector \codebook~is more effective than shuffling the atom order itself. Readers can read the 
\href{https://openreview.net/forum?id=zxxSJAVQPc&referrer=%5BAuthor%20Console%5D(%2Fgroup%3Fid%3DICML.cc%2F2024%2FConference%2FAuthors%23your-submissions)}{openreview rebuttal}.

\begin{table}
    \caption{Self-consistency of decoded sequences. ``C@$N$'' denotes the decoded results of $N$ out of the total $1024$ permutations for each molecule are consistent. ``Avg.'' denotes the average consistency of all test data.  }
    \vspace{-3mm}
    \resizebox{1.0 \columnwidth}{!}{
        \begin{tabular}{lcccc|c} 
            \toprule
            Models  & C@$256$  & C@$512$  & C@$768$  & C@$1024$ & \underline{Avg.} \\
            \midrule 
            GraphsGPT-1W  & 100\% & 99.2\% & 94.1\% & 77.3\% & \underline{96.1\%} \\
            GraphsGPT-8W  & 100\% & 99.4\% & 96.5\% & 85.3\% & \textbf{97.9\%} \\
            \bottomrule
        \end{tabular}
    }
    \label{tab:consistency}
    \vspace{-6mm}
\end{table}

% \paragraph{Q3} Thank you for the training details. It is amazing that you can process so much data and pretrain in so few hours. Could you also share dataset statistics? E.g. for your choice of tokenization, what is the distribution of sequence lengths (mean/max) across the datasets that you pretrain on?

% \paragraph{R3} We have computed statistics for the 870M molecules in our dataset, and the distribution of atom/boond/token numbers (mean/max) across the pretraining datasets is summarized as follows:

% \vspace{-2mm}
% \begin{table}[h]
%     \centering
%     \caption{Data Statistics. We show the maximum and average numbers of atoms, bonds, and graph tokens for the samples in pretraining dataset.}
%     \vspace{-2mm}
%     \begin{tabular}{lcc} 
%         \toprule
%         & max & mean \\ \hline
%         atoms & 69 & 25.4 \\
%         bonds & 76 & 27.3 \\
%         tokens & 143 & 52.7 \\ \bottomrule  
%     \end{tabular}
%     \vspace{-5mm}
% \end{table}

\section*{Acknowledgements}
This work was supported by National Science and Technology Major Project (No. 2022ZD0115101), National Natural Science Foundation of China Project (No. U21A20427), Project (No. WU2022A009) from the Center of Synthetic Biology and Integrated Bioengineering of Westlake University and Integrated Bioengineering of Westlake University and Project (No. WU2023C019) from the Westlake University Industries of the Future Research Funding .

\section*{Impact Statement}
This paper presents work whose goal is to advance the field of Machine Learning. There are many potential societal consequences  of our work, none which we feel must be specifically highlighted here. \graphsgpt~provides a new paradigm for graph representation, generation and manipulation. The Non-Euclidean to Euclidean transformation may affect broader downstream graph applications, such as graph translation and optimization. The methodology could be extend to other modalities, such as image and sequence. 

\nocite{langley00}

\bibliography{main}

\begin{thebibliography}{112}
\providecommand{\natexlab}[1]{#1}
\providecommand{\url}[1]{\texttt{#1}}
\expandafter\ifx\csname urlstyle\endcsname\relax
  \providecommand{\doi}[1]{doi: #1}\else
  \providecommand{\doi}{doi: \begingroup \urlstyle{rm}\Url}\fi

\bibitem[Achiam et~al.(2023)Achiam, Adler, Agarwal, Ahmad, Akkaya, Aleman, Almeida, Altenschmidt, Altman, Anadkat, et~al.]{achiam2023gpt}
Achiam, J., Adler, S., Agarwal, S., Ahmad, L., Akkaya, I., Aleman, F.~L., Almeida, D., Altenschmidt, J., Altman, S., Anadkat, S., et~al.
\newblock Gpt-4 technical report.
\newblock \emph{arXiv preprint arXiv:2303.08774}, 2023.

\bibitem[Ahmed et~al.(2013)Ahmed, Shervashidze, Narayanamurthy, Josifovski, and Smola]{ahmed2013distributed}
Ahmed, A., Shervashidze, N., Narayanamurthy, S., Josifovski, V., and Smola, A.~J.
\newblock Distributed large-scale natural graph factorization.
\newblock In \emph{Proceedings of the 22nd international conference on World Wide Web}, pp.\  37--48, 2013.

\bibitem[Alayrac et~al.(2022)Alayrac, Donahue, Luc, Miech, Barr, Hasson, Lenc, Mensch, Millican, Reynolds, et~al.]{alayrac2022flamingo}
Alayrac, J.-B., Donahue, J., Luc, P., Miech, A., Barr, I., Hasson, Y., Lenc, K., Mensch, A., Millican, K., Reynolds, M., et~al.
\newblock Flamingo: a visual language model for few-shot learning.
\newblock \emph{Advances in Neural Information Processing Systems}, 35:\penalty0 23716--23736, 2022.

\bibitem[Bagal et~al.(2021)Bagal, Aggarwal, Vinod, and Priyakumar]{bagal2021molgpt}
Bagal, V., Aggarwal, R., Vinod, P., and Priyakumar, U.~D.
\newblock Molgpt: molecular generation using a transformer-decoder model.
\newblock \emph{Journal of Chemical Information and Modeling}, 62\penalty0 (9):\penalty0 2064--2076, 2021.

\bibitem[Brown et~al.(2019)Brown, Fiscato, Segler, and Vaucher]{brown2019guacamol}
Brown, N., Fiscato, M., Segler, M.~H., and Vaucher, A.~C.
\newblock Guacamol: benchmarking models for de novo molecular design.
\newblock \emph{Journal of chemical information and modeling}, 59\penalty0 (3):\penalty0 1096--1108, 2019.

\bibitem[Brown et~al.(2020)Brown, Mann, Ryder, Subbiah, Kaplan, Dhariwal, Neelakantan, Shyam, Sastry, Askell, et~al.]{brown2020language}
Brown, T., Mann, B., Ryder, N., Subbiah, M., Kaplan, J.~D., Dhariwal, P., Neelakantan, A., Shyam, P., Sastry, G., Askell, A., et~al.
\newblock Language models are few-shot learners.
\newblock \emph{Advances in neural information processing systems}, 33:\penalty0 1877--1901, 2020.

\bibitem[Chanpuriya \& Musco(2020)Chanpuriya and Musco]{chanpuriya2020infinitewalk}
Chanpuriya, S. and Musco, C.
\newblock Infinitewalk: Deep network embeddings as laplacian embeddings with a nonlinearity.
\newblock In \emph{Proceedings of the 26th ACM SIGKDD International Conference on Knowledge Discovery \& Data Mining}, pp.\  1325--1333, 2020.

\bibitem[Chen et~al.(2022)Chen, O’Bray, and Borgwardt]{chen2022structure}
Chen, D., O’Bray, L., and Borgwardt, K.
\newblock Structure-aware transformer for graph representation learning.
\newblock In \emph{International Conference on Machine Learning}, pp.\  3469--3489. PMLR, 2022.

\bibitem[Chen et~al.(2018)Chen, Ma, and Xiao]{chen2018fastgcn}
Chen, J., Ma, T., and Xiao, C.
\newblock Fastgcn: fast learning with graph convolutional networks via importance sampling.
\newblock \emph{arXiv preprint arXiv:1801.10247}, 2018.

\bibitem[Chiang et~al.(2019)Chiang, Liu, Si, Li, Bengio, and Hsieh]{chiang2019cluster}
Chiang, W.-L., Liu, X., Si, S., Li, Y., Bengio, S., and Hsieh, C.-J.
\newblock Cluster-gcn: An efficient algorithm for training deep and large graph convolutional networks.
\newblock In \emph{Proceedings of the 25th ACM SIGKDD international conference on knowledge discovery \& data mining}, pp.\  257--266, 2019.

\bibitem[Dehghani et~al.(2023)Dehghani, Djolonga, Mustafa, Padlewski, Heek, Gilmer, Steiner, et~al.]{dehghani2023scaling}
Dehghani, M., Djolonga, J., Mustafa, B., Padlewski, P., Heek, J., Gilmer, J., Steiner, A.~P., et~al.
\newblock Scaling vision transformers to 22 billion parameters.
\newblock In \emph{ICML}, pp.\  7480--7512. PMLR, 2023.

\bibitem[Devlin et~al.(2018)Devlin, Chang, Lee, and Toutanova]{devlin2018bert}
Devlin, J., Chang, M.-W., Lee, K., and Toutanova, K.
\newblock Bert: Pre-training of deep bidirectional transformers for language understanding.
\newblock \emph{arXiv:1810.04805}, 2018.

\bibitem[Diehl(2019)]{diehl2019edge}
Diehl, F.
\newblock Edge contraction pooling for graph neural networks.
\newblock \emph{arXiv preprint arXiv:1905.10990}, 2019.

\bibitem[Dosovitskiy et~al.(2020)Dosovitskiy, Beyer, Kolesnikov, Weissenborn, Zhai, Unterthiner, Dehghani, Minderer, Heigold, Gelly, et~al.]{dosovitskiy2020image}
Dosovitskiy, A., Beyer, L., Kolesnikov, A., Weissenborn, D., Zhai, X., Unterthiner, T., Dehghani, M., Minderer, M., Heigold, G., Gelly, S., et~al.
\newblock An image is worth 16x16 words: Transformers for image recognition at scale.
\newblock \emph{arXiv preprint arXiv:2010.11929}, 2020.

\bibitem[Dwivedi \& Bresson(2020)Dwivedi and Bresson]{dwivedi2020generalization}
Dwivedi, V.~P. and Bresson, X.
\newblock A generalization of transformer networks to graphs.
\newblock \emph{arXiv preprint arXiv:2012.09699}, 2020.

\bibitem[Gao et~al.(2022{\natexlab{a}})Gao, Tan, and Li]{gao2022pifold}
Gao, Z., Tan, C., and Li, S.~Z.
\newblock Pifold: Toward effective and efficient protein inverse folding.
\newblock In \emph{The Eleventh International Conference on Learning Representations}, 2022{\natexlab{a}}.

\bibitem[Gao et~al.(2022{\natexlab{b}})Gao, Tan, Wu, and Li]{gao2022simvp}
Gao, Z., Tan, C., Wu, L., and Li, S.~Z.
\newblock Simvp: Simpler yet better video prediction.
\newblock In \emph{Proceedings of the IEEE/CVF conference on computer vision and pattern recognition}, pp.\  3170--3180, 2022{\natexlab{b}}.

\bibitem[Gao et~al.(2023)Gao, Tan, Chen, Zhang, Xia, Li, and Li]{gao2023kw}
Gao, Z., Tan, C., Chen, X., Zhang, Y., Xia, J., Li, S., and Li, S.~Z.
\newblock Kw-design: Pushing the limit of protein deign via knowledge refinement.
\newblock In \emph{The Twelfth International Conference on Learning Representations}, 2023.

\bibitem[Grover \& Leskovec(2016)Grover and Leskovec]{grover2016node2vec}
Grover, A. and Leskovec, J.
\newblock node2vec: Scalable feature learning for networks.
\newblock In \emph{Proceedings of the 22nd ACM SIGKDD international conference on Knowledge discovery and data mining}, pp.\  855--864, 2016.

\bibitem[Guo \& Mao(2023)Guo and Mao]{guo2023interpolating}
Guo, H. and Mao, Y.
\newblock Interpolating graph pair to regularize graph classification.
\newblock In \emph{AAAI}, volume~37, pp.\  7766--7774, 2023.

\bibitem[Hamilton et~al.(2017)Hamilton, Ying, and Leskovec]{hamilton2017inductive}
Hamilton, W., Ying, Z., and Leskovec, J.
\newblock Inductive representation learning on large graphs.
\newblock \emph{Advances in neural information processing systems}, 30, 2017.

\bibitem[Han et~al.(2022)Han, Jiang, Liu, and Hu]{han2022g}
Han, X., Jiang, Z., Liu, N., and Hu, X.
\newblock G-mixup: Graph data augmentation for graph classification.
\newblock In \emph{ICML}, pp.\  8230--8248. PMLR, 2022.

\bibitem[Hou et~al.(2022)Hou, Liu, Cen, Dong, Yang, Wang, and Tang]{hou2022graphmae}
Hou, Z., Liu, X., Cen, Y., Dong, Y., Yang, H., Wang, C., and Tang, J.
\newblock Graphmae: Self-supervised masked graph autoencoders.
\newblock In \emph{Proceedings of the 28th ACM SIGKDD Conference on Knowledge Discovery and Data Mining}, pp.\  594--604, 2022.

\bibitem[Hu et~al.(2019)Hu, Liu, Gomes, Zitnik, Liang, Pande, and Leskovec]{hu2020strategies}
Hu, W., Liu, B., Gomes, J., Zitnik, M., Liang, P., Pande, V., and Leskovec, J.
\newblock Strategies for pre-training graph neural networks.
\newblock \emph{arXiv preprint arXiv:1905.12265}, 2019.

\bibitem[Hu et~al.(2020{\natexlab{a}})Hu, Dong, Wang, Chang, and Sun]{hu2020gpt}
Hu, Z., Dong, Y., Wang, K., Chang, K.-W., and Sun, Y.
\newblock Gpt-gnn: Generative pre-training of graph neural networks.
\newblock In \emph{Proceedings of the 26th ACM SIGKDD International Conference on Knowledge Discovery \& Data Mining}, pp.\  1857--1867, 2020{\natexlab{a}}.

\bibitem[Hu et~al.(2020{\natexlab{b}})Hu, Dong, Wang, Chang, and Sun]{hu2020gpt-gnn}
Hu, Z., Dong, Y., Wang, K., Chang, K.-W., and Sun, Y.
\newblock Gpt-gnn: Generative pre-training of graph neural networks.
\newblock In \emph{Proceedings of the 26th ACM SIGKDD International Conference on Knowledge Discovery \& Data Mining}, pp.\  1857--1867, 2020{\natexlab{b}}.

\bibitem[Hu et~al.(2020{\natexlab{c}})Hu, Dong, Wang, and Sun]{hu2020heterogeneous}
Hu, Z., Dong, Y., Wang, K., and Sun, Y.
\newblock Heterogeneous graph transformer.
\newblock In \emph{Proceedings of the web conference 2020}, pp.\  2704--2710, 2020{\natexlab{c}}.

\bibitem[Huang et~al.(2022)Huang, Peng, Ma, and Zhang]{huang20223dlinker}
Huang, Y., Peng, X., Ma, J., and Zhang, M.
\newblock 3dlinker: an e (3) equivariant variational autoencoder for molecular linker design.
\newblock \emph{arXiv preprint arXiv:2205.07309}, 2022.

\bibitem[Hussain et~al.(2021)Hussain, Zaki, and Subramanian]{hussain2021edge}
Hussain, M.~S., Zaki, M.~J., and Subramanian, D.
\newblock Edge-augmented graph transformers: Global self-attention is enough for graphs.
\newblock \emph{arXiv preprint arXiv:2108.03348}, 2021.

\bibitem[Hwang et~al.(2020)Hwang, Park, Kwon, Kim, Ha, and Kim]{hwang2020self}
Hwang, D., Park, J., Kwon, S., Kim, K., Ha, J.-W., and Kim, H.~J.
\newblock Self-supervised auxiliary learning with meta-paths for heterogeneous graphs.
\newblock \emph{Advances in Neural Information Processing Systems}, 33:\penalty0 10294--10305, 2020.

\bibitem[Irwin \& Shoichet(2005)Irwin and Shoichet]{irwin2005zinc}
Irwin, J.~J. and Shoichet, B.~K.
\newblock Zinc- a free database of commercially available compounds for virtual screening.
\newblock \emph{Journal of chemical information and modeling}, 45\penalty0 (1):\penalty0 177--182, 2005.

\bibitem[Jin et~al.(2020)Jin, Derr, Liu, Wang, Wang, Liu, and Tang]{jin2020self}
Jin, W., Derr, T., Liu, H., Wang, Y., Wang, S., Liu, Z., and Tang, J.
\newblock Self-supervised learning on graphs: Deep insights and new direction.
\newblock \emph{arXiv preprint arXiv:2006.10141}, 2020.

\bibitem[Jo et~al.(2022)Jo, Lee, and Hwang]{jo2022score}
Jo, J., Lee, S., and Hwang, S.~J.
\newblock Score-based generative modeling of graphs via the system of stochastic differential equations.
\newblock In \emph{International Conference on Machine Learning}, pp.\  10362--10383. PMLR, 2022.

\bibitem[Kim et~al.(2022)Kim, Nguyen, Min, Cho, Lee, Lee, and Hong]{kim2022pure}
Kim, J., Nguyen, D., Min, S., Cho, S., Lee, M., Lee, H., and Hong, S.
\newblock Pure transformers are powerful graph learners.
\newblock \emph{Advances in Neural Information Processing Systems}, 35:\penalty0 14582--14595, 2022.

\bibitem[Kipf \& Welling(2016{\natexlab{a}})Kipf and Welling]{kipf2016semi}
Kipf, T.~N. and Welling, M.
\newblock Semi-supervised classification with graph convolutional networks.
\newblock \emph{arXiv preprint arXiv:1609.02907}, 2016{\natexlab{a}}.

\bibitem[Kipf \& Welling(2016{\natexlab{b}})Kipf and Welling]{kipf2016variational}
Kipf, T.~N. and Welling, M.
\newblock Variational graph auto-encoders.
\newblock \emph{arXiv preprint arXiv:1611.07308}, 2016{\natexlab{b}}.

\bibitem[Kong et~al.(2023)Kong, Cui, Sun, Zhuang, Prakash, and Zhang]{kong2023autoregressive}
Kong, L., Cui, J., Sun, H., Zhuang, Y., Prakash, B.~A., and Zhang, C.
\newblock Autoregressive diffusion model for graph generation.
\newblock In \emph{International conference on machine learning}, pp.\  17391--17408. PMLR, 2023.

\bibitem[Kreuzer et~al.(2021)Kreuzer, Beaini, Hamilton, L{\'e}tourneau, and Tossou]{kreuzer2021rethinking}
Kreuzer, D., Beaini, D., Hamilton, W., L{\'e}tourneau, V., and Tossou, P.
\newblock Rethinking graph transformers with spectral attention.
\newblock \emph{Advances in Neural Information Processing Systems}, 34:\penalty0 21618--21629, 2021.

\bibitem[Lee et~al.(2019)Lee, Lee, and Kang]{lee2019self}
Lee, J., Lee, I., and Kang, J.
\newblock Self-attention graph pooling.
\newblock In \emph{International conference on machine learning}, pp.\  3734--3743. PMLR, 2019.

\bibitem[Li et~al.(2023{\natexlab{a}})Li, Sun, Ling, and Peng]{li2023survey}
Li, X., Sun, L., Ling, M., and Peng, Y.
\newblock A survey of graph neural network based recommendation in social networks.
\newblock \emph{Neurocomputing}, pp.\  126441, 2023{\natexlab{a}}.

\bibitem[Li et~al.(2023{\natexlab{b}})Li, Gao, Tan, Li, and Yang]{li2023general}
Li, Z., Gao, Z., Tan, C., Li, S.~Z., and Yang, L.~T.
\newblock General point model with autoencoding and autoregressive.
\newblock \emph{arXiv preprint arXiv:2310.16861}, 2023{\natexlab{b}}.

\bibitem[Lin et~al.(2022{\natexlab{a}})Lin, Gao, Xu, Wu, Li, and Li]{lin2022conditional}
Lin, H., Gao, Z., Xu, Y., Wu, L., Li, L., and Li, S.~Z.
\newblock Conditional local convolution for spatio-temporal meteorological forecasting.
\newblock In \emph{Proceedings of the AAAI conference on artificial intelligence}, volume~36, pp.\  7470--7478, 2022{\natexlab{a}}.

\bibitem[Lin et~al.(2022{\natexlab{b}})Lin, Tian, Hou, and Zhao]{lin2022improving}
Lin, Z., Tian, C., Hou, Y., and Zhao, W.~X.
\newblock Improving graph collaborative filtering with neighborhood-enriched contrastive learning.
\newblock In \emph{Proceedings of the ACM Web Conference 2022}, pp.\  2320--2329, 2022{\natexlab{b}}.

\bibitem[Liu et~al.(2023{\natexlab{a}})Liu, Li, Lin, and Zhang]{liu2023gnnrec}
Liu, C., Li, Y., Lin, H., and Zhang, C.
\newblock Gnnrec: Gated graph neural network for session-based social recommendation model.
\newblock \emph{Journal of Intelligent Information Systems}, 60\penalty0 (1):\penalty0 137--156, 2023{\natexlab{a}}.

\bibitem[Liu et~al.(2021{\natexlab{a}})Liu, Yan, Oztekin, and Ji]{liu2021graphebm}
Liu, M., Yan, K., Oztekin, B., and Ji, S.
\newblock Graphebm: Molecular graph generation with energy-based models.
\newblock \emph{arXiv preprint arXiv:2102.00546}, 2021{\natexlab{a}}.

\bibitem[Liu et~al.(2021{\natexlab{b}})Liu, Wang, Liu, Lasenby, Guo, and Tang]{liu2022pretraining}
Liu, S., Wang, H., Liu, W., Lasenby, J., Guo, H., and Tang, J.
\newblock Pre-training molecular graph representation with 3d geometry.
\newblock \emph{arXiv preprint arXiv:2110.07728}, 2021{\natexlab{b}}.

\bibitem[Liu et~al.(2021{\natexlab{c}})Liu, Zhang, Hou, Mian, Wang, Zhang, and Tang]{liu2021self}
Liu, X., Zhang, F., Hou, Z., Mian, L., Wang, Z., Zhang, J., and Tang, J.
\newblock Self-supervised learning: Generative or contrastive.
\newblock \emph{IEEE transactions on knowledge and data engineering}, 35\penalty0 (1):\penalty0 857--876, 2021{\natexlab{c}}.

\bibitem[Liu et~al.(2022)Liu, Jin, Pan, Zhou, Zheng, Xia, and Philip]{liu2022graph}
Liu, Y., Jin, M., Pan, S., Zhou, C., Zheng, Y., Xia, F., and Philip, S.~Y.
\newblock Graph self-supervised learning: A survey.
\newblock \emph{IEEE Transactions on Knowledge and Data Engineering}, 35\penalty0 (6):\penalty0 5879--5900, 2022.

\bibitem[Liu et~al.(2023{\natexlab{b}})Liu, Yang, Zhou, Liu, Wang, Liang, Tu, and Li]{liu2023simple}
Liu, Y., Yang, X., Zhou, S., Liu, X., Wang, S., Liang, K., Tu, W., and Li, L.
\newblock Simple contrastive graph clustering.
\newblock \emph{IEEE Transactions on Neural Networks and Learning Systems}, 2023{\natexlab{b}}.

\bibitem[Liu et~al.(2023{\natexlab{c}})Liu, Yang, Zhou, Liu, Wang, Liang, Tu, Li, Duan, and Chen]{liu2023hard}
Liu, Y., Yang, X., Zhou, S., Liu, X., Wang, Z., Liang, K., Tu, W., Li, L., Duan, J., and Chen, C.
\newblock Hard sample aware network for contrastive deep graph clustering.
\newblock In \emph{Proceedings of the AAAI conference on artificial intelligence}, volume~37, pp.\  8914--8922, 2023{\natexlab{c}}.

\bibitem[Liu et~al.(2021{\natexlab{d}})Liu, Lin, Cao, Hu, Wei, Zhang, Lin, and Guo]{liu2021swin}
Liu, Z., Lin, Y., Cao, Y., Hu, H., Wei, Y., Zhang, Z., Lin, S., and Guo, B.
\newblock Swin transformer: Hierarchical vision transformer using shifted windows.
\newblock In \emph{ICCV}, pp.\  10012--10022, 2021{\natexlab{d}}.

\bibitem[Luo et~al.(2021)Luo, Yan, and Ji]{luo2021graphdf}
Luo, Y., Yan, K., and Ji, S.
\newblock Graphdf: A discrete flow model for molecular graph generation.
\newblock In \emph{International conference on machine learning}, pp.\  7192--7203. PMLR, 2021.

\bibitem[Ma et~al.(2019)Ma, Wang, Aggarwal, and Tang]{ma2019graph}
Ma, Y., Wang, S., Aggarwal, C.~C., and Tang, J.
\newblock Graph convolutional networks with eigenpooling.
\newblock In \emph{Proceedings of the 25th ACM SIGKDD international conference on knowledge discovery \& data mining}, pp.\  723--731, 2019.

\bibitem[Martinkus et~al.(2022)Martinkus, Loukas, Perraudin, and Wattenhofer]{martinkus2022spectre}
Martinkus, K., Loukas, A., Perraudin, N., and Wattenhofer, R.
\newblock Spectre: Spectral conditioning helps to overcome the expressivity limits of one-shot graph generators.
\newblock In \emph{International Conference on Machine Learning}, pp.\  15159--15179. PMLR, 2022.

\bibitem[McInnes \& Healy(2017)McInnes and Healy]{mcinnes2017accelerated}
McInnes, L. and Healy, J.
\newblock Accelerated hierarchical density based clustering.
\newblock In \emph{Data Mining Workshops (ICDMW), 2017 IEEE International Conference on}, pp.\  33--42. IEEE, 2017.

\bibitem[McInnes et~al.(2018)McInnes, Healy, and Melville]{mcinnes2018umap}
McInnes, L., Healy, J., and Melville, J.
\newblock Umap: Uniform manifold approximation and projection for dimension reduction.
\newblock \emph{arXiv preprint arXiv:1802.03426}, 2018.

\bibitem[Mialon et~al.(2021)Mialon, Chen, Selosse, and Mairal]{mialon2021graphit}
Mialon, G., Chen, D., Selosse, M., and Mairal, J.
\newblock Graphit: Encoding graph structure in transformers.
\newblock \emph{arXiv preprint arXiv:2106.05667}, 2021.

\bibitem[Min et~al.(2022)Min, Chen, Bian, Xu, Zhao, Huang, Zhao, Huang, Ananiadou, and Rong]{min2022transformer}
Min, E., Chen, R., Bian, Y., Xu, T., Zhao, K., Huang, W., Zhao, P., Huang, J., Ananiadou, S., and Rong, Y.
\newblock Transformer for graphs: An overview from architecture perspective.
\newblock \emph{arXiv preprint arXiv:2202.08455}, 2022.

\bibitem[Niu et~al.(2020)Niu, Song, Song, Zhao, Grover, and Ermon]{niu2020permutation}
Niu, C., Song, Y., Song, J., Zhao, S., Grover, A., and Ermon, S.
\newblock Permutation invariant graph generation via score-based generative modeling.
\newblock In \emph{International Conference on Artificial Intelligence and Statistics}, pp.\  4474--4484. PMLR, 2020.

\bibitem[Pang et~al.(2022)Pang, Wang, Tay, Liu, Tian, and Yuan]{pang2022masked}
Pang, Y., Wang, W., Tay, F.~E., Liu, W., Tian, Y., and Yuan, L.
\newblock Masked autoencoders for point cloud self-supervised learning.
\newblock In \emph{ECCV}, pp.\  604--621. Springer, 2022.

\bibitem[Park et~al.(2022)Park, Shim, and Yang]{park2022graph}
Park, J., Shim, H., and Yang, E.
\newblock Graph transplant: Node saliency-guided graph mixup with local structure preservation.
\newblock In \emph{Proceedings of the AAAI Conference on Artificial Intelligence}, volume~36, pp.\  7966--7974, 2022.

\bibitem[Peng et~al.(2022)Peng, Luo, Guan, Xie, Peng, and Ma]{peng2022pocket2mol}
Peng, X., Luo, S., Guan, J., Xie, Q., Peng, J., and Ma, J.
\newblock Pocket2mol: Efficient molecular sampling based on 3d protein pockets.
\newblock In \emph{International Conference on Machine Learning}, pp.\  17644--17655. PMLR, 2022.

\bibitem[Peng et~al.(2020{\natexlab{a}})Peng, Dong, Luo, Wu, and Zheng]{peng2020self}
Peng, Z., Dong, Y., Luo, M., Wu, X.-M., and Zheng, Q.
\newblock Self-supervised graph representation learning via global context prediction.
\newblock \emph{arXiv:2003.01604}, 2020{\natexlab{a}}.

\bibitem[Peng et~al.(2020{\natexlab{b}})Peng, Huang, Luo, Zheng, Rong, Xu, and Huang]{peng2020graph}
Peng, Z., Huang, W., Luo, M., Zheng, Q., Rong, Y., Xu, T., and Huang, J.
\newblock Graph representation learning via graphical mutual information maximization.
\newblock In \emph{Proceedings of The Web Conference 2020}, pp.\  259--270, 2020{\natexlab{b}}.

\bibitem[Perozzi et~al.(2014)Perozzi, Al-Rfou, and Skiena]{perozzi2014deepwalk}
Perozzi, B., Al-Rfou, R., and Skiena, S.
\newblock Deepwalk: Online learning of social representations.
\newblock In \emph{Proceedings of the 20th ACM SIGKDD international conference on Knowledge discovery and data mining}, pp.\  701--710, 2014.

\bibitem[Polykovskiy et~al.(2020)Polykovskiy, Zhebrak, Sanchez-Lengeling, Golovanov, Tatanov, Belyaev, Kurbanov, Artamonov, Aladinskiy, Veselov, et~al.]{polykovskiy2020molecular}
Polykovskiy, D., Zhebrak, A., Sanchez-Lengeling, B., Golovanov, S., Tatanov, O., Belyaev, S., Kurbanov, R., Artamonov, A., Aladinskiy, V., Veselov, M., et~al.
\newblock Molecular sets (moses): a benchmarking platform for molecular generation models.
\newblock \emph{Frontiers in pharmacology}, 11:\penalty0 565644, 2020.

\bibitem[Qiu et~al.(2020)Qiu, Chen, Dong, Zhang, Yang, Ding, Wang, and Tang]{qiu2020gcc}
Qiu, J., Chen, Q., Dong, Y., Zhang, J., Yang, H., Ding, M., Wang, K., and Tang, J.
\newblock Gcc: Graph contrastive coding for graph neural network pre-training.
\newblock In \emph{Proceedings of the 26th ACM SIGKDD international conference on knowledge discovery \& data mining}, pp.\  1150--1160, 2020.

\bibitem[Ramachandran et~al.(2017)Ramachandran, Zoph, and Le]{ramachandran2017searching}
Ramachandran, P., Zoph, B., and Le, Q.~V.
\newblock Searching for activation functions.
\newblock \emph{arXiv:1710.05941}, 2017.

\bibitem[Ramp{\'a}{\v{s}}ek et~al.(2022)Ramp{\'a}{\v{s}}ek, Galkin, Dwivedi, Luu, Wolf, and Beaini]{rampavsek2022recipe}
Ramp{\'a}{\v{s}}ek, L., Galkin, M., Dwivedi, V.~P., Luu, A.~T., Wolf, G., and Beaini, D.
\newblock Recipe for a general, powerful, scalable graph transformer.
\newblock \emph{Advances in Neural Information Processing Systems}, 35:\penalty0 14501--14515, 2022.

\bibitem[Rong et~al.(2020)Rong, Bian, Xu, Xie, Wei, Huang, and Huang]{rong2020self}
Rong, Y., Bian, Y., Xu, T., Xie, W., Wei, Y., Huang, W., and Huang, J.
\newblock Self-supervised graph transformer on large-scale molecular data.
\newblock \emph{Advances in Neural Information Processing Systems}, 33:\penalty0 12559--12571, 2020.

\bibitem[Shakibajahromi et~al.(2024)Shakibajahromi, Kim, and Breen]{shakibajahromi2024rimeshgnn}
Shakibajahromi, B., Kim, E., and Breen, D.~E.
\newblock Rimeshgnn: A rotation-invariant graph neural network for mesh classification.
\newblock In \emph{WACV}, pp.\  3150--3160, 2024.

\bibitem[Shi et~al.(2019)Shi, Xu, Zhu, Zhang, Zhang, and Tang]{shi2019graphaf}
Shi, C., Xu, M., Zhu, Z., Zhang, W., Zhang, M., and Tang, J.
\newblock Graphaf: a flow-based autoregressive model for molecular graph generation.
\newblock In \emph{International Conference on Learning Representations}, 2019.

\bibitem[Shi et~al.(2020)Shi, Xu, Zhu, Zhang, Zhang, and Tang]{shi2020graphaf}
Shi, C., Xu, M., Zhu, Z., Zhang, W., Zhang, M., and Tang, J.
\newblock Graphaf: a flow-based autoregressive model for molecular graph generation.
\newblock \emph{arXiv preprint arXiv:2001.09382}, 2020.

\bibitem[St{\"a}rk et~al.(2022)St{\"a}rk, Beaini, Corso, Tossou, Dallago, G{\"u}nnemann, and Li{\`o}]{stark20223d}
St{\"a}rk, H., Beaini, D., Corso, G., Tossou, P., Dallago, C., G{\"u}nnemann, S., and Li{\`o}, P.
\newblock 3d infomax improves gnns for molecular property prediction.
\newblock In \emph{ICML}, pp.\  20479--20502. PMLR, 2022.

\bibitem[Sun et~al.(2019)Sun, Hoffmann, Verma, and Tang]{sun2020infograph}
Sun, F.-Y., Hoffmann, J., Verma, V., and Tang, J.
\newblock Infograph: Unsupervised and semi-supervised graph-level representation learning via mutual information maximization.
\newblock \emph{arXiv preprint arXiv:1908.01000}, 2019.

\bibitem[Suresh et~al.(2021)Suresh, Li, Hao, and Neville]{suresh2021adversarial}
Suresh, S., Li, P., Hao, C., and Neville, J.
\newblock Adversarial graph augmentation to improve graph contrastive learning.
\newblock \emph{Advances in Neural Information Processing Systems}, 34:\penalty0 15920--15933, 2021.

\bibitem[Tan et~al.(2023)Tan, Gao, and Li]{tan2023target}
Tan, C., Gao, Z., and Li, S.~Z.
\newblock Target-aware molecular graph generation.
\newblock In \emph{Joint European Conference on Machine Learning and Knowledge Discovery in Databases}, pp.\  410--427. Springer, 2023.

\bibitem[Tian et~al.(2023)Tian, Dong, Zhang, Zhang, and Chawla]{tian2023heterogeneous}
Tian, Y., Dong, K., Zhang, C., Zhang, C., and Chawla, N.~V.
\newblock Heterogeneous graph masked autoencoders.
\newblock In \emph{Proceedings of the AAAI Conference on Artificial Intelligence}, volume~37, pp.\  9997--10005, 2023.

\bibitem[Veli{\v{c}}kovi{\'c} et~al.(2017)Veli{\v{c}}kovi{\'c}, Cucurull, Casanova, Romero, Lio, and Bengio]{velivckovic2017graph}
Veli{\v{c}}kovi{\'c}, P., Cucurull, G., Casanova, A., Romero, A., Lio, P., and Bengio, Y.
\newblock Graph attention networks.
\newblock \emph{arXiv preprint arXiv:1710.10903}, 2017.

\bibitem[Vignac et~al.(2022)Vignac, Krawczuk, Siraudin, Wang, Cevher, and Frossard]{vignac2022digress}
Vignac, C., Krawczuk, I., Siraudin, A., Wang, B., Cevher, V., and Frossard, P.
\newblock Digress: Discrete denoising diffusion for graph generation.
\newblock \emph{arXiv preprint arXiv:2209.14734}, 2022.

\bibitem[Wang et~al.(2021)Wang, Agarwal, Ham, Choudhury, and Reddy]{wang2021self}
Wang, P., Agarwal, K., Ham, C., Choudhury, S., and Reddy, C.~K.
\newblock Self-supervised learning of contextual embeddings for link prediction in heterogeneous networks.
\newblock In \emph{Proceedings of the web conference 2021}, pp.\  2946--2957, 2021.

\bibitem[Wang et~al.(2022)Wang, Wang, Cao, and Barati~Farimani]{wang2022molecular}
Wang, Y., Wang, J., Cao, Z., and Barati~Farimani, A.
\newblock Molecular contrastive learning of representations via graph neural networks.
\newblock \emph{NMI}, 4\penalty0 (3):\penalty0 279--287, 2022.

\bibitem[Wu et~al.(2019)Wu, Souza, Zhang, Fifty, Yu, and Weinberger]{wu2019simplifying}
Wu, F., Souza, A., Zhang, T., Fifty, C., Yu, T., and Weinberger, K.
\newblock Simplifying graph convolutional networks.
\newblock In \emph{ICML}, pp.\  6861--6871. PMLR, 2019.

\bibitem[Wu et~al.(2021{\natexlab{a}})Wu, Lin, Tan, Gao, and Li]{wu2021self}
Wu, L., Lin, H., Tan, C., Gao, Z., and Li, S.~Z.
\newblock Self-supervised learning on graphs: Contrastive, generative, or predictive.
\newblock \emph{IEEE Transactions on Knowledge and Data Engineering}, 2021{\natexlab{a}}.

\bibitem[Wu et~al.(2022)Wu, Xia, Gao, et~al.]{wu2022graphmixup}
Wu, L., Xia, J., Gao, Z., et~al.
\newblock Graphmixup: Improving class-imbalanced node classification by reinforcement mixup and self-supervised context prediction.
\newblock In \emph{ECML-PKDD}, pp.\  519--535. Springer, 2022.

\bibitem[Wu et~al.(2024{\natexlab{a}})Wu, Huang, Tan, Gao, Hu, Lin, Liu, and Li]{wu2024psc}
Wu, L., Huang, Y., Tan, C., Gao, Z., Hu, B., Lin, H., Liu, Z., and Li, S.~Z.
\newblock Psc-cpi: Multi-scale protein sequence-structure contrasting for efficient and generalizable compound-protein interaction prediction.
\newblock \emph{arXiv preprint arXiv:2402.08198}, 2024{\natexlab{a}}.

\bibitem[Wu et~al.(2024{\natexlab{b}})Wu, Tian, Huang, Li, Lin, Chawla, and Li]{wu2024mape}
Wu, L., Tian, Y., Huang, Y., Li, S., Lin, H., Chawla, N.~V., and Li, S.~Z.
\newblock Mape-ppi: Towards effective and efficient protein-protein interaction prediction via microenvironment-aware protein embedding.
\newblock \emph{arXiv preprint arXiv:2402.14391}, 2024{\natexlab{b}}.

\bibitem[Wu et~al.(2018)Wu, Ramsundar, Feinberg, Gomes, Geniesse, Pappu, Leswing, and Pande]{wu2018moleculenet}
Wu, Z., Ramsundar, B., Feinberg, E.~N., Gomes, J., Geniesse, C., Pappu, A.~S., Leswing, K., and Pande, V.
\newblock Moleculenet: a benchmark for molecular machine learning.
\newblock \emph{Chemical science}, 9\penalty0 (2):\penalty0 513--530, 2018.

\bibitem[Wu et~al.(2021{\natexlab{b}})Wu, Jain, Wright, Mirhoseini, Gonzalez, and Stoica]{wu2021representing}
Wu, Z., Jain, P., Wright, M., Mirhoseini, A., Gonzalez, J.~E., and Stoica, I.
\newblock Representing long-range context for graph neural networks with global attention.
\newblock \emph{NeurIPS}, 34:\penalty0 13266--13279, 2021{\natexlab{b}}.

\bibitem[Xia et~al.(2022{\natexlab{a}})Xia, Wu, Chen, Hu, and Li]{xia2022simgrace}
Xia, J., Wu, L., Chen, J., Hu, B., and Li, S.~Z.
\newblock Simgrace: A simple framework for graph contrastive learning without data augmentation.
\newblock In \emph{Proceedings of the ACM Web Conference 2022}, pp.\  1070--1079, 2022{\natexlab{a}}.

\bibitem[Xia et~al.(2022{\natexlab{b}})Xia, Zhao, Hu, Gao, Tan, Liu, Li, and Li]{xia2022mole}
Xia, J., Zhao, C., Hu, B., Gao, Z., Tan, C., Liu, Y., Li, S., and Li, S.~Z.
\newblock Mole-bert: Rethinking pre-training graph neural networks for molecules.
\newblock In \emph{The Eleventh International Conference on Learning Representations}, 2022{\natexlab{b}}.

\bibitem[Xia et~al.(2023)Xia, Zhao, Hu, Gao, Tan, Liu, Li, and Li]{xia2023mole}
Xia, J., Zhao, C., Hu, B., Gao, Z., Tan, C., Liu, Y., Li, S., and Li, S.~Z.
\newblock Mole-bert: Rethinking pre-training graph neural networks for molecules.
\newblock In \emph{The Eleventh International Conference on Learning Representations}, 2023.

\bibitem[Xiao et~al.(2020)Xiao, Zhao, Zheng, and Song]{xiao2020vertex}
Xiao, W., Zhao, H., Zheng, V.~W., and Song, Y.
\newblock Vertex-reinforced random walk for network embedding.
\newblock In \emph{Proceedings of the 2020 SIAM International Conference on Data Mining}, pp.\  595--603. SIAM, 2020.

\bibitem[Xie et~al.(2022)Xie, Xu, Zhang, Wang, and Ji]{xie2022self}
Xie, Y., Xu, Z., Zhang, J., Wang, Z., and Ji, S.
\newblock Self-supervised learning of graph neural networks: A unified review.
\newblock \emph{IEEE transactions on pattern analysis and machine intelligence}, 45\penalty0 (2):\penalty0 2412--2429, 2022.

\bibitem[Xu et~al.(2018)Xu, Hu, Leskovec, and Jegelka]{xu2018powerful}
Xu, K., Hu, W., Leskovec, J., and Jegelka, S.
\newblock How powerful are graph neural networks?
\newblock \emph{arXiv preprint arXiv:1810.00826}, 2018.

\bibitem[Xu et~al.(2021)Xu, Wang, Ni, Guo, and Tang]{xu2021self}
Xu, M., Wang, H., Ni, B., Guo, H., and Tang, J.
\newblock Self-supervised graph-level representation learning with local and global structure.
\newblock In \emph{International Conference on Machine Learning}, pp.\  11548--11558. PMLR, 2021.

\bibitem[Ying et~al.(2021)Ying, Cai, Luo, Zheng, Ke, He, Shen, and Liu]{ying2021transformers}
Ying, C., Cai, T., Luo, S., Zheng, S., Ke, G., He, D., Shen, Y., and Liu, T.-Y.
\newblock Do transformers really perform badly for graph representation?
\newblock \emph{Advances in Neural Information Processing Systems}, 34:\penalty0 28877--28888, 2021.

\bibitem[Ying et~al.(2018)Ying, You, Morris, Ren, Hamilton, and Leskovec]{ying2018hierarchical}
Ying, Z., You, J., Morris, C., Ren, X., Hamilton, W., and Leskovec, J.
\newblock Hierarchical graph representation learning with differentiable pooling.
\newblock \emph{Advances in neural information processing systems}, 31, 2018.

\bibitem[You et~al.(2020)You, Chen, Sui, Chen, Wang, and Shen]{you2020graph}
You, Y., Chen, T., Sui, Y., Chen, T., Wang, Z., and Shen, Y.
\newblock Graph contrastive learning with augmentations.
\newblock \emph{NeurIPS}, 33:\penalty0 5812--5823, 2020.

\bibitem[You et~al.(2021)You, Chen, Shen, and Wang]{you2021graph}
You, Y., Chen, T., Shen, Y., and Wang, Z.
\newblock Graph contrastive learning automated.
\newblock In \emph{International Conference on Machine Learning}, pp.\  12121--12132. PMLR, 2021.

\bibitem[Yu et~al.(2022)Yu, Tang, Rao, et~al.]{yu2022point}
Yu, X., Tang, L., Rao, Y., et~al.
\newblock Point-bert: Pre-training 3d point cloud transformers with masked point modeling.
\newblock In \emph{CVPR}, pp.\  19313--19322, 2022.

\bibitem[Zang \& Wang(2020)Zang and Wang]{zang2020moflow}
Zang, C. and Wang, F.
\newblock Moflow: an invertible flow model for generating molecular graphs.
\newblock In \emph{Proceedings of the 26th ACM SIGKDD international conference on knowledge discovery \& data mining}, pp.\  617--626, 2020.

\bibitem[Zeng \& Xie(2021)Zeng and Xie]{zeng2021contrastive}
Zeng, J. and Xie, P.
\newblock Contrastive self-supervised learning for graph classification.
\newblock In \emph{AAAI}, volume~35, pp.\  10824--10832, 2021.

\bibitem[Zhang \& Sennrich(2019)Zhang and Sennrich]{zhang2019root}
Zhang, B. and Sennrich, R.
\newblock Root mean square layer normalization.
\newblock \emph{Advances in Neural Information Processing Systems}, 32, 2019.

\bibitem[Zhang et~al.(2023)Zhang, Luo, and Wei]{zhang2023mixupexplainer}
Zhang, J., Luo, D., and Wei, H.
\newblock Mixupexplainer: Generalizing explanations for graph neural networks with data augmentation.
\newblock In \emph{Proceedings of the 29th ACM SIGKDD Conference on Knowledge Discovery and Data Mining}, pp.\  3286--3296, 2023.

\bibitem[Zhang et~al.(2021)Zhang, Liu, Wang, Lu, and Lee]{zhang2021motif}
Zhang, Z., Liu, Q., Wang, H., Lu, C., and Lee, C.-K.
\newblock Motif-based graph self-supervised learning for molecular property prediction.
\newblock \emph{Advances in Neural Information Processing Systems}, 34:\penalty0 15870--15882, 2021.

\bibitem[Zhao et~al.(2021)Zhao, Li, Wen, Wang, Liu, Sun, Xie, and Ye]{zhao2021gophormer}
Zhao, J., Li, C., Wen, Q., Wang, Y., Liu, Y., Sun, H., Xie, X., and Ye, Y.
\newblock Gophormer: Ego-graph transformer for node classification.
\newblock \emph{arXiv preprint arXiv:2110.13094}, 2021.

\bibitem[Zhou et~al.(2020{\natexlab{a}})Zhou, Cui, Hu, Zhang, Yang, Liu, Wang, Li, and Sun]{zhou2020graph}
Zhou, J., Cui, G., Hu, S., Zhang, Z., Yang, C., Liu, Z., Wang, L., Li, C., and Sun, M.
\newblock Graph neural networks: A review of methods and applications.
\newblock \emph{AI open}, 1:\penalty0 57--81, 2020{\natexlab{a}}.

\bibitem[Zhou et~al.(2020{\natexlab{b}})Zhou, Shen, and Xuan]{zhou2020data}
Zhou, J., Shen, J., and Xuan, Q.
\newblock Data augmentation for graph classification.
\newblock In \emph{Proceedings of the 29th ACM International Conference on Information \& Knowledge Management}, pp.\  2341--2344, 2020{\natexlab{b}}.

\bibitem[Zhu et~al.(2020)Zhu, Xu, Yu, Liu, Wu, and Wang]{zhu2020deep}
Zhu, Y., Xu, Y., Yu, F., Liu, Q., Wu, S., and Wang, L.
\newblock Deep graph contrastive representation learning.
\newblock \emph{arXiv preprint arXiv:2006.04131}, 2020.

\bibitem[Zhu et~al.(2021)Zhu, Xu, Yu, et~al.]{zhu2021graph}
Zhu, Y., Xu, Y., Yu, F., et~al.
\newblock Graph contrastive learning with adaptive augmentation.
\newblock In \emph{Proceedings of the Web Conference 2021}, pp.\  2069--2080, 2021.

\bibitem[Zou et~al.(2022)Zou, Wei, Mao, et~al.]{zou2022multi}
Zou, D., Wei, W., Mao, X.-L., et~al.
\newblock Multi-level cross-view contrastive learning for knowledge-aware recommender system.
\newblock In \emph{SIGIR}, pp.\  1358--1368, 2022.

\end{thebibliography}
\bibliographystyle{icml2021}

\onecolumn
\appendix
\section{Representation}

When applying graph mixup, the training samples are drawn from the original data with probability $p_{self}$ and from mixed data with probability $(1-p_{self})$. The mixup hyperparameter $\alpha$ and $p_{self}$ are shown in Table \ref{tab:hyperparameter}.

  \begin{table}[H]
    \centering
    \resizebox{0.8 \columnwidth}{!}{
    \begin{tabular}{llllllllll}
    \toprule
               & Tox21 & ToxCast & Sider & HIV  & BBBP       & BACE & ESOL & FreeSolv & LIPO\\ \midrule
    batch size & 16    & 16      & 16    & 64   & 128        & 16   & 16    & 64    & 16\\
    lr         & 1e-5  & 5e-5    & 1e-4  & 1e-4 & 5e-4       & 1e-5 & 1e-4  & 1e-4  & 5e-5\\
    dropout    & 0.0   & 0.0     & 0.0   & 0.0  & 0.1 or 0.3 & 0.0  & 0.1   & 0.1   & 0.0\\
    epoch      & 50    & 50      & 50    & 50   & 50 or 100  & 50   & 50    & 50    & 50\\ 
    $\alpha$ for mixup & 0.5 & 0.1 & 0.5 &0.5 & 0.5 & 0.5 & 0.5 &0.5 & 0.1\\
    $p_{self}$ for mixup & 0.7 & 0.7 &0.7 &0.5 & 0.5 & 0.7 & 0.7 &0.9 & 0.7\\
    \bottomrule
    \end{tabular}}
    \caption{Hyperparameters for property prediction.}
    \label{tab:hyperparameter}
  \end{table}

\clearpage
\section{Generation}
\label{app:metric}
% 对评估指标的详细描述，对输入属性logp,sa, qed的介绍
\subsection{Few-Shots Generation}
We introduce metrics \citep{bagal2021molgpt} of few-shots generation as follows:
\begin{itemize}
  \item \textbf{Validity}: the fraction of a generated molecules that are valid. We use RDkit for validity check of molecules. Validity measures how well the model has learned the SMILES grammar and the valency of atoms.
  \item \textbf{Uniqueness}: the fraction of valid generated molecules that are unique. Low uniqueness highlights repetitive molecule generation and a low level of distribution learning by the model.
  \item \textbf{Novelty}: the fraction of valid unique generated molecules that are not in the training set. Low novelty is a sign of overfitting. We do not want the model to memorize the training data.
  \item \textbf{Internal Diversity ($\text{IntDiv}_p$)}: measures the diversity of the generated molecules, which is a metric specially designed to check for mode collapse or whether the model keeps generating similar structures. This uses the power ($p$) mean of the Tanimoto similarity ($T$) between the fingerprints of all pairs of molecules $(s1, s2)$ in the generated set ($S$).
\end{itemize}

\begin{equation}
  \text{InvDiv}_p(S) = 1- \sqrt[p]{\frac{1}{|S|^2} \sum_{s1, s2 \in S} T(s1, s2)^p}
\end{equation}

\subsection{Conditional Generation}
\label{app:cond_gen}
% 对评估指标的详细描述，对输入属性logp,sa, qed的介绍
\label{app:conditional_generation}
We provide a detailed description of the conditions used for conditional generation as follows:

\begin{itemize}
    \item \textbf{QED (Quantitative Estimate of Drug-likeness):} a measure that quantifies the ``drug-likeness" of a molecule based on its pharmacokinetic profile, ranging from $0$ to $1$.
    \item \textbf{SA (Synthetic Accessibility):} a score that predicts the difficulty of synthesizing a molecule based on multiple factors. Lower SA scores indicate easier synthesis.
    \item \textbf{logP (Partition Coefficient):} a key parameter in studies of drug absorption and distribution in the body that measuring a molecule's hydrophobicity.
    \item \textbf{Scaffold:} the core structure of a molecule, which typically includes rings and the atoms that connect them. It provides a framework upon which different functional groups can be added to create new molecules.
\end{itemize}

In order to integrate conditional information into our model, we set aside an additional $100$M molecules from the ZINC database for finetuning, which we denote as the dataset $\gD_{\gG}$. For each molecule $\gG \in \gD_{\gG}$, we compute its property values $v_{\text{QED}}$, $v_{\text{SA}}$ and $v_{\text{logP}}$ and normalize them to 0 mean and 1.0 variance, yielding $\bar{v}_{\text{QED}}$, $\bar{v}_{\text{SA}}$ and $\bar{v}_{\text{logP}}$. 

The \encoder~model takes all properties and scaffolds as inputs and transforms them into the \graphword~sequence $\gW=[\vw_1, \vw_2, \cdots, \vw_k]$. The additional property and scaffold information enables \encoder~to encode \graphwords~with conditions. The \graphwords~are then subsequently decoded by \decoder~following the same implementation in Section \ref{sec:decoder}. In summary, the inputs of the \encoder~encoder comprises:
\begin{enumerate}
    \item \textbf{\graphword~Prompts} $[\texttt{[GW 1]}, \cdots, \texttt{[GW k]}]$, which are identical to the word prompts discussed in Section \ref{sec:word_prompts}.
    \item \textbf{Property Token Sequence} $[\texttt{[QED]},\texttt{[SA]},\texttt{[logP]}]$, which is encoded from the normalized property values $\bar{v}_{\text{QED}}$, $\bar{v}_{\text{SA}}$ and $\bar{v}_{\text{logP}}$.
    \item \textbf{Scaffold Flexible Token Sequence} $\texttt{FTSeq}_\text{Scaf}$, representing the sequence of the scaffold for the molecule.
\end{enumerate}

For the sake of comparison, we followed \citet{bagal2021molgpt} and trained a MolGPT model on the GuacaMol dataset \cite{brown2019guacamol} using QED, SA, logP, and scaffolds as conditions for 10 epochs. We compare the conditional generation ability by measuring the MAD (Mean Absolute Deviation), SD (Standard Deviation), validity and uniqueness. Table \ref{tab:condition_complete_molgpt} presents the full results, underscoring the superior control of \decoder-1W-C over molecular properties.

\begin{table}[htbp]
    \centering
    \resizebox{1.0 \columnwidth}{!}{
    \begin{tabular}{cclcccccccccc}
        \toprule
         & Pretrain & Metric & QED=0.5 & QED=0.7 & QED=0.9 & SA=0.7 & SA=0.8 & SA=0.9 & logP=0.0 & logP=2.0 & logP=4.0 & \underline{Avg.} \\
         \midrule
         
        \multirow{3}{*}{\rotatebox[origin=c]{90}{\small{MolGPT}}} &
        \multirow{3}{*}{\XSolidBrush}
         & MAD $\downarrow$ & 0.081 & 0.082 & 0.097 & 0.024 & 0.019 & 0.013 & 0.304 & 0.239 & 0.286 & \underline{0.127} \\
         & & SD $\downarrow$ & 0.065 & 0.066 & 0.092 & 0.022 & 0.016 & 0.013 & 0.295 & 0.232 & 0.258 & \underline{\textbf{0.118}} \\
         & & Validity $\uparrow$ & 0.985 & 0.985 & 0.984 & 0.975 & 0.988 & 0.995 & 0.982 & 0.983 & 0.982 & \underline{0.984} \\
         
        \midrule
        \multirow{6}{*}{\rotatebox[origin=c]{90}{\small{GraphGPT-1W-C}}} &
         \multirow{3}{*}{\XSolidBrush}
         & MAD $\downarrow$ & 0.041 & 0.031 & 0.077 & 0.012 & 0.028 & 0.031 & 0.103 & 0.189 & 0.201 & \underline{0.079} \\
         & & SD $\downarrow$ & 0.079 & 0.077 & 0.121 & 0.055 & 0.062 & 0.070 & 0.460 & 0.656 & 0.485 & \underline{0.229} \\
         & & Validity $\uparrow$ & 0.988 & 0.995 & 0.991 & 0.995 & 0.991 & 0.998 & 0.980 & 0.992 & 0.991 & \underline{0.991} \\
         
         \cdashline{2-13}
         & \multirow{3}{*}{\CheckmarkBold}
         & MAD $\downarrow$ & 0.032 & 0.033 & 0.051 & 0.002 & 0.009 & 0.022 & 0.017 & 0.190 & 0.268 & \underline{\textbf{0.069}} \\
         & & SD $\downarrow$ & 0.080 & 0.075 & 0.090 & 0.042 & 0.037 & 0.062 & 0.463 & 0.701 & 0.796 & \underline{0.261} \\
         & & Validity $\uparrow$ & 0.996 & 0.998 & 0.999 & 0.995 & 0.999 & 0.996 & 0.994 & 0.990 & 0.992 & \underline{\textbf{0.995}} \\
        \bottomrule
    \end{tabular}}
    \caption{Overall comparison between \decoder-1W-C and MolGPT on different properties with scaffold SMILES ``c1ccccc1". ``MAD" denotes the Mean Absolute Deviation of the property value in generated molecules compared to the oracle value. ``SD" denotes the Standard Deviation of the generated property.}
    \label{tab:condition_complete_molgpt}
\end{table}

\begin{figure}[htbp]
  \centering
  \begin{subfigure}{0.48\linewidth}
      \centering
      \includegraphics[width=1.0\linewidth]{figs/scaffold_qed_density0.175.pdf}
      \caption{QED}
      \label{fig:condition_qed}
  \end{subfigure}
  \hfill
  \begin{subfigure}{0.48\linewidth}
      \centering
      \includegraphics[width=1.0\linewidth]{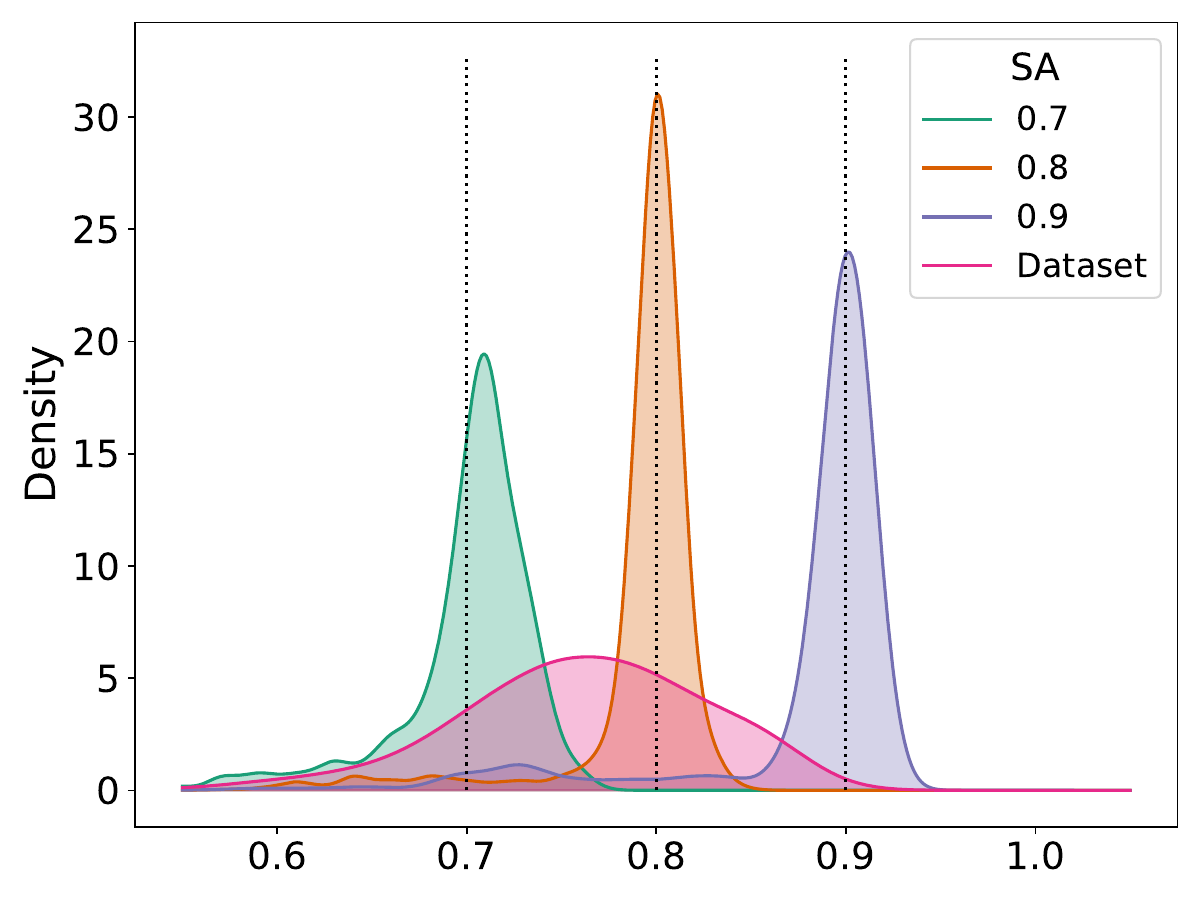}
      \caption{SA}
      \label{fig:condition_sa}
  \end{subfigure}
  \hfill
  \begin{subfigure}{0.48\linewidth}
      \centering
      \includegraphics[width=1.0\linewidth]{figs/scaffold_logp_density0.225.pdf}
      \caption{logP}
      \label{fig:condition_logp}
  \end{subfigure}
  \vspace{-3mm}
  \caption{ Property distribution of generated molecules on different conditions using \decoder-1W-C.  }
  \vspace{-3mm}
\end{figure}

\clearpage
\section{\graphwords}
\label{sec:graph_words}

\subsection{Clustering}
The efficacy of the \encoder~encoder hinges on its ability to effectively map Non-Euclidean graphs into Euclidean latent features in a structured manner. To investigate this, we visualize the latent \graphwords~space using sampled features, encoding 32,768 molecules with \encoder-1W and employing HDBSCAN \cite{mcinnes2017accelerated} for clustering the \graphwords. 

\begin{figure}[h]
    \centering
    \includegraphics[width=1.0\linewidth]{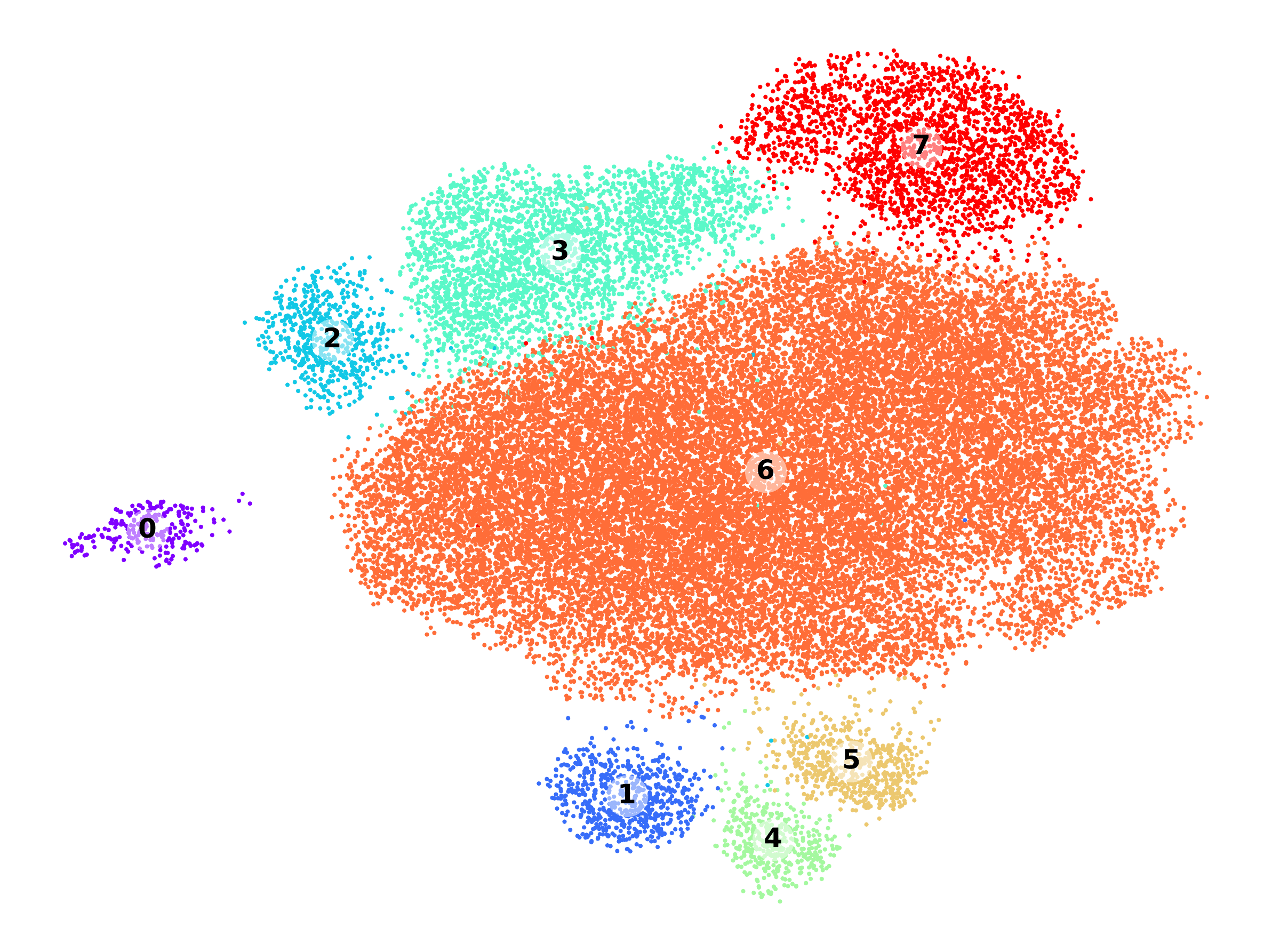}
    \caption{UMAP \cite{mcinnes2018umap} visualization of the clustering result on the \graphwords~of \encoder-1W.}
    \label{fig:clustering}
\end{figure}

\begin{figure}[H]
    \centering
    \resizebox{1.0 \columnwidth}{!}{
    \begin{subfigure}{0.1\linewidth}\includegraphics[width=1.0\linewidth]{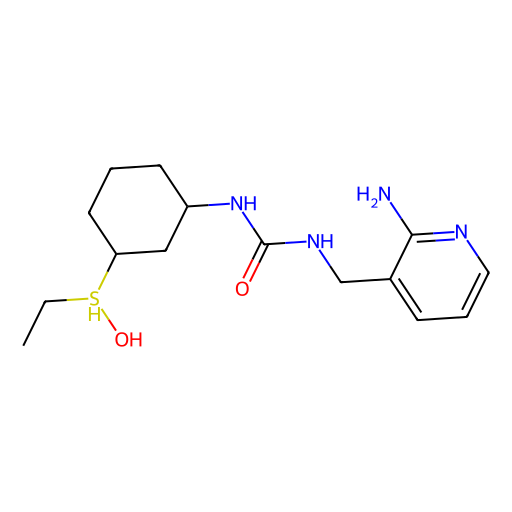}\end{subfigure}\hfill
    \begin{subfigure}{0.1\linewidth}\includegraphics[width=1.0\linewidth]{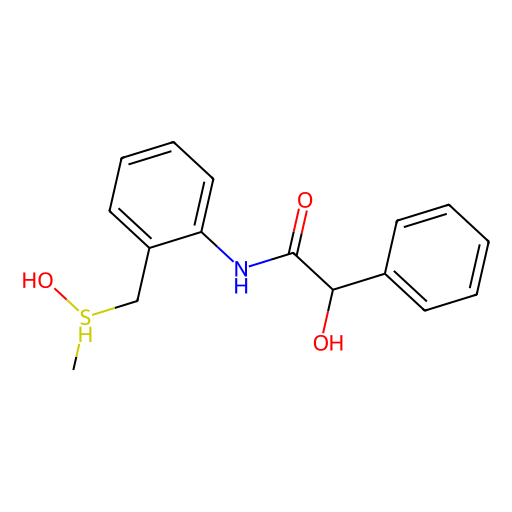}\end{subfigure}\hfill
    \begin{subfigure}{0.1\linewidth}\includegraphics[width=1.0\linewidth]{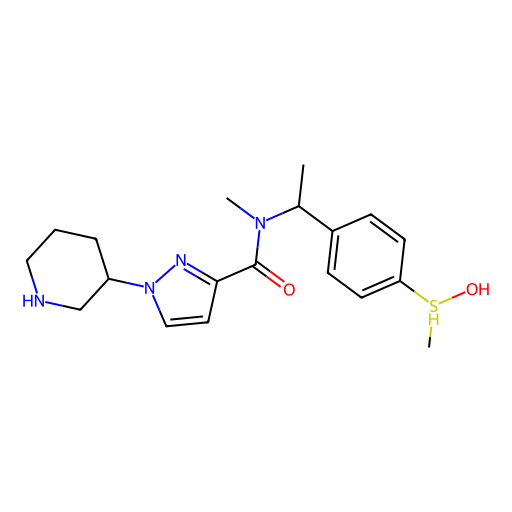}\end{subfigure}\hfill
    \begin{subfigure}{0.1\linewidth}\includegraphics[width=1.0\linewidth]{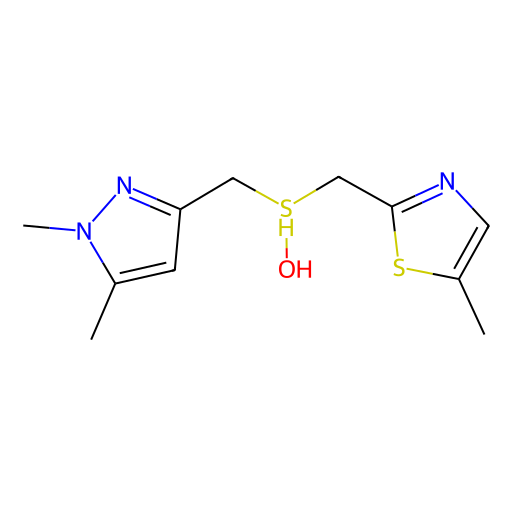}\end{subfigure}\hfill
    \begin{subfigure}{0.1\linewidth}\includegraphics[width=1.0\linewidth]{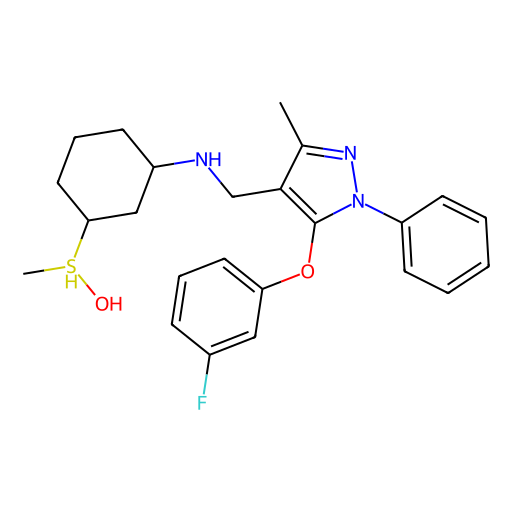}\end{subfigure}\hfill
    \begin{subfigure}{0.1\linewidth}\includegraphics[width=1.0\linewidth]{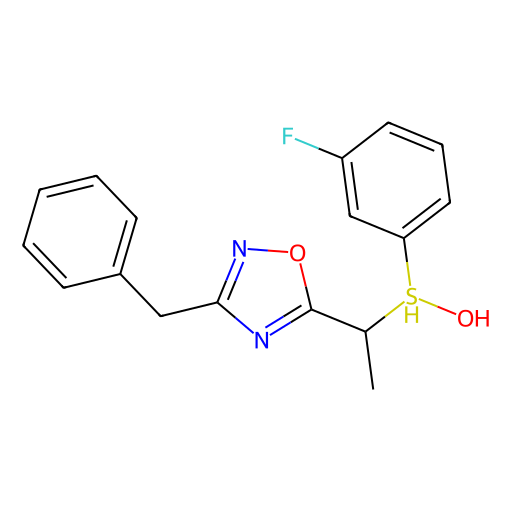}\end{subfigure}\hfill
    \begin{subfigure}{0.1\linewidth}\includegraphics[width=1.0\linewidth]{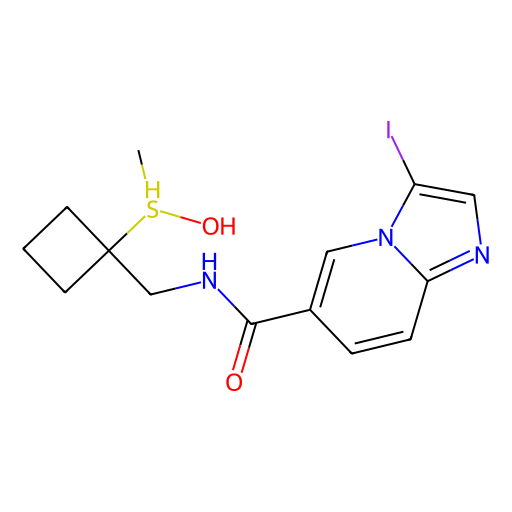}\end{subfigure}\hfill
    \begin{subfigure}{0.1\linewidth}\includegraphics[width=1.0\linewidth]{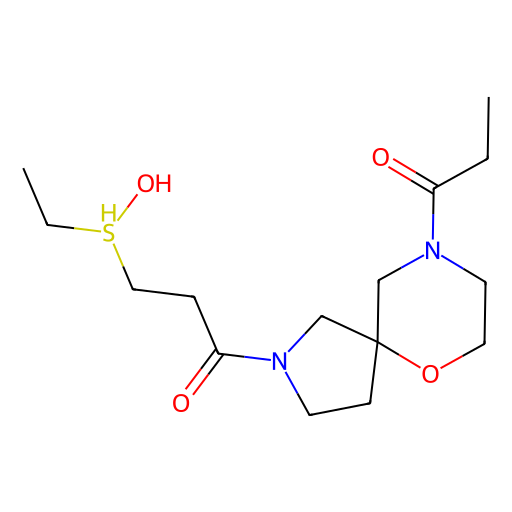}\end{subfigure}
    } \\
    \small{\textbf{(a) }{Cluster 0}} \\
    \resizebox{1.0 \columnwidth}{!}{
    \begin{subfigure}{0.1\linewidth}\includegraphics[width=1.0\linewidth]{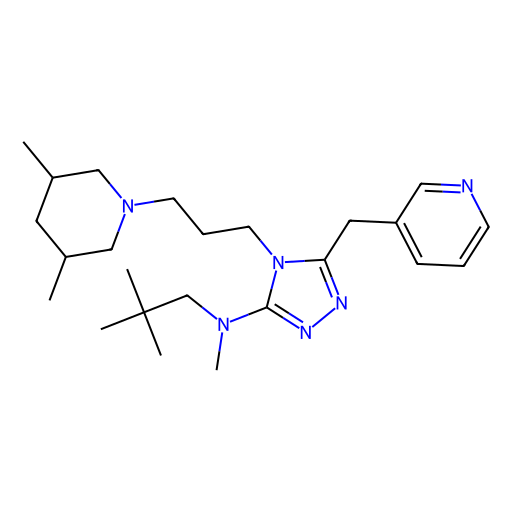}\end{subfigure}\hfill
    \begin{subfigure}{0.1\linewidth}\includegraphics[width=1.0\linewidth]{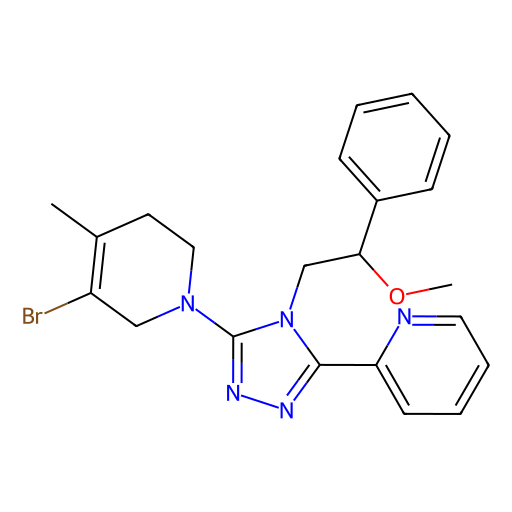}\end{subfigure}\hfill
    \begin{subfigure}{0.1\linewidth}\includegraphics[width=1.0\linewidth]{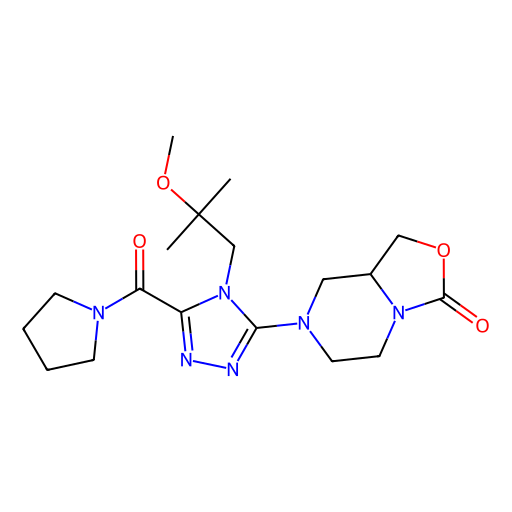}\end{subfigure}\hfill
    \begin{subfigure}{0.1\linewidth}\includegraphics[width=1.0\linewidth]{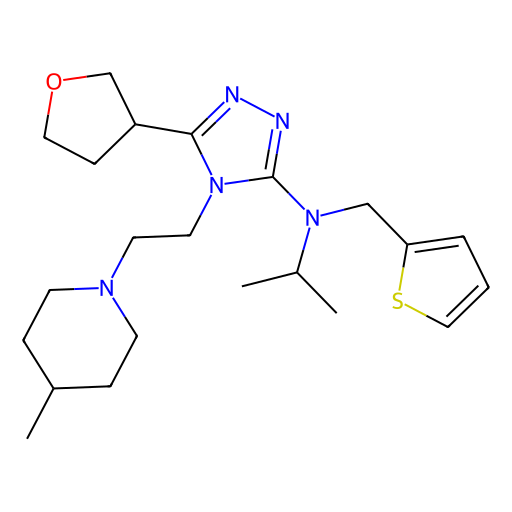}\end{subfigure}\hfill
    \begin{subfigure}{0.1\linewidth}\includegraphics[width=1.0\linewidth]{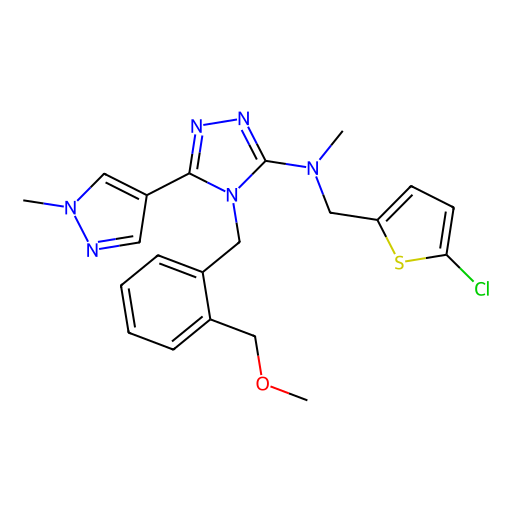}\end{subfigure}\hfill
    \begin{subfigure}{0.1\linewidth}\includegraphics[width=1.0\linewidth]{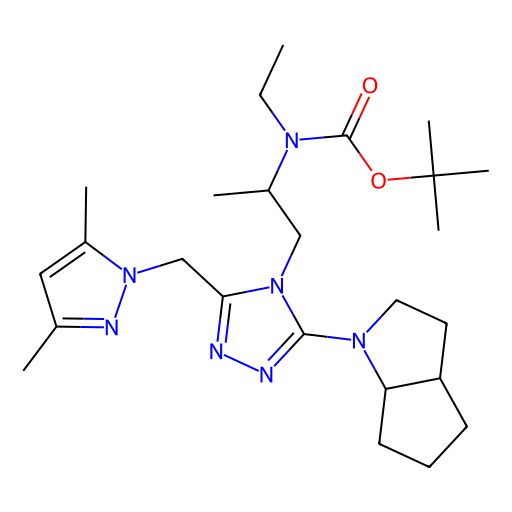}\end{subfigure}\hfill
    \begin{subfigure}{0.1\linewidth}\includegraphics[width=1.0\linewidth]{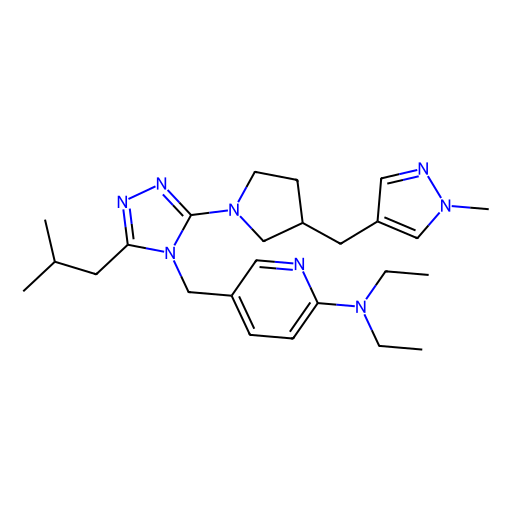}\end{subfigure}\hfill
    \begin{subfigure}{0.1\linewidth}\includegraphics[width=1.0\linewidth]{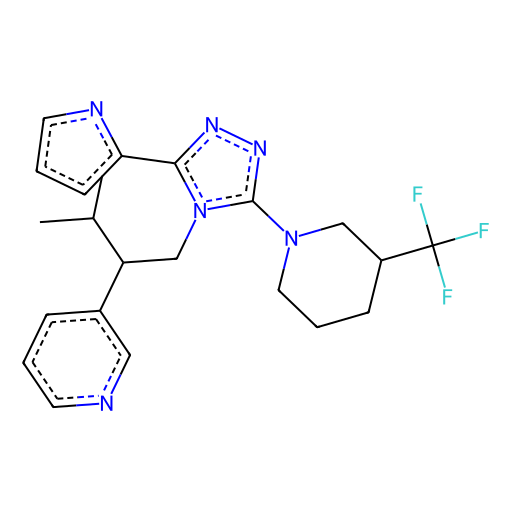}\end{subfigure}
    } \\
    \small{\textbf{(b) }{Cluster 1}} \\
    \resizebox{1.0 \columnwidth}{!}{
    \begin{subfigure}{0.1\linewidth}\includegraphics[width=1.0\linewidth]{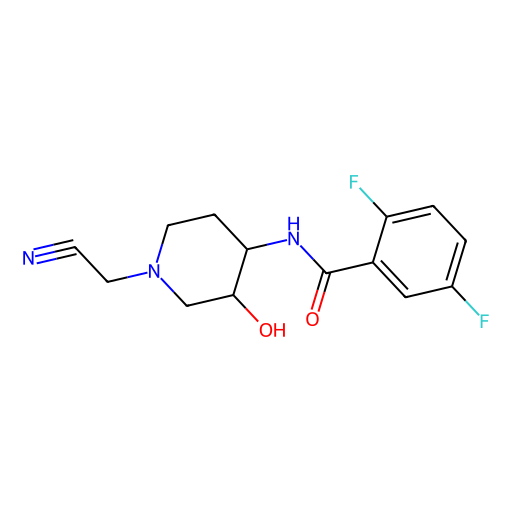}\end{subfigure}\hfill
    \begin{subfigure}{0.1\linewidth}\includegraphics[width=1.0\linewidth]{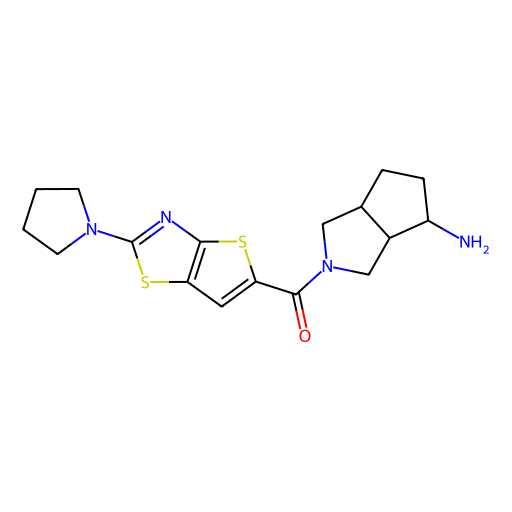}\end{subfigure}\hfill
    \begin{subfigure}{0.1\linewidth}\includegraphics[width=1.0\linewidth]{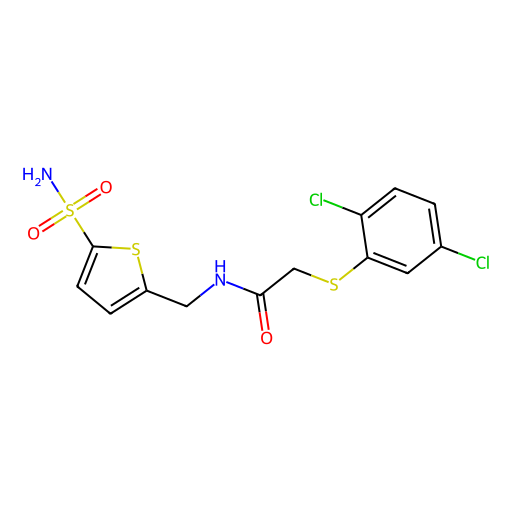}\end{subfigure}\hfill
    \begin{subfigure}{0.1\linewidth}\includegraphics[width=1.0\linewidth]{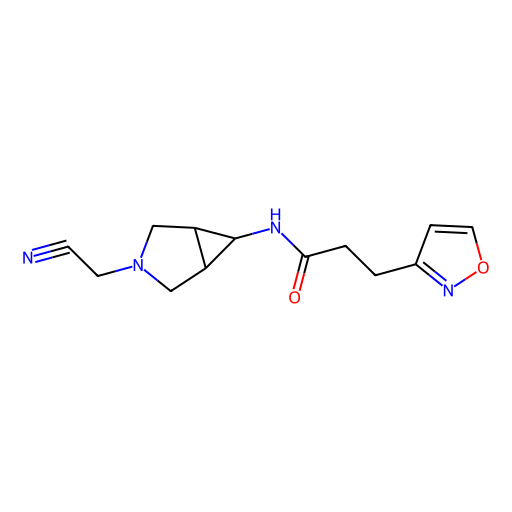}\end{subfigure}\hfill
    \begin{subfigure}{0.1\linewidth}\includegraphics[width=1.0\linewidth]{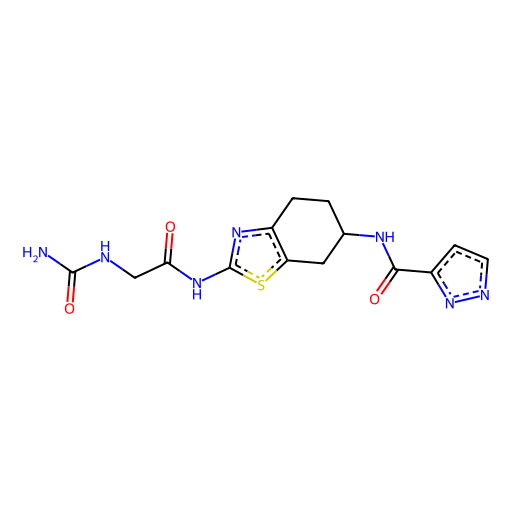}\end{subfigure}\hfill
    \begin{subfigure}{0.1\linewidth}\includegraphics[width=1.0\linewidth]{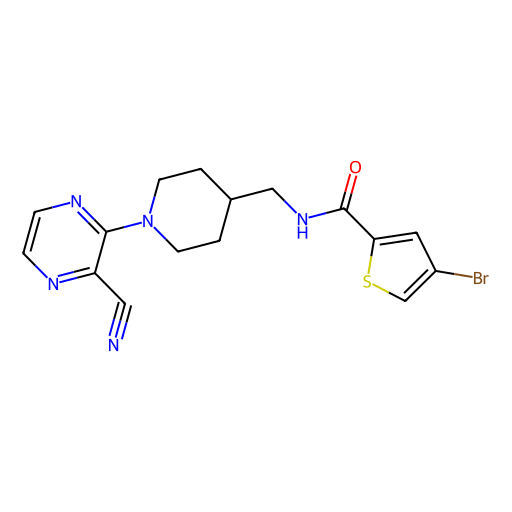}\end{subfigure}\hfill
    \begin{subfigure}{0.1\linewidth}\includegraphics[width=1.0\linewidth]{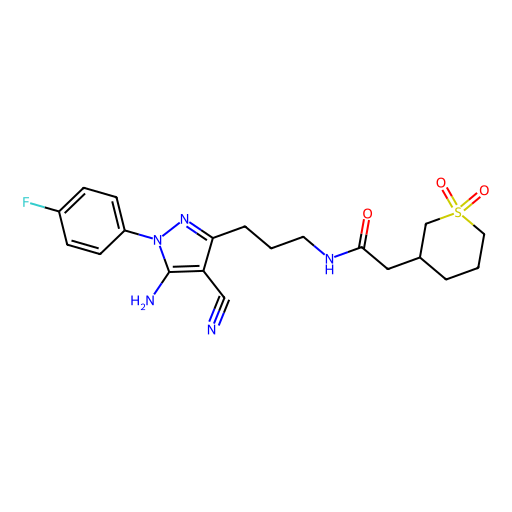}\end{subfigure}\hfill
    \begin{subfigure}{0.1\linewidth}\includegraphics[width=1.0\linewidth]{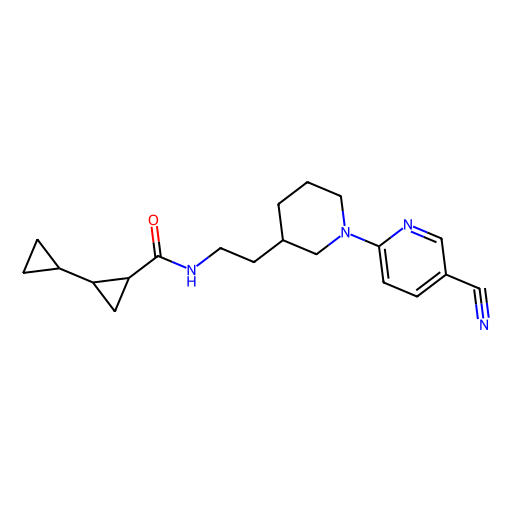}\end{subfigure}
    } \\
    \small{\textbf{(c) }{Cluster 2}} \\
    \resizebox{1.0 \columnwidth}{!}{
    \begin{subfigure}{0.1\linewidth}\includegraphics[width=1.0\linewidth]{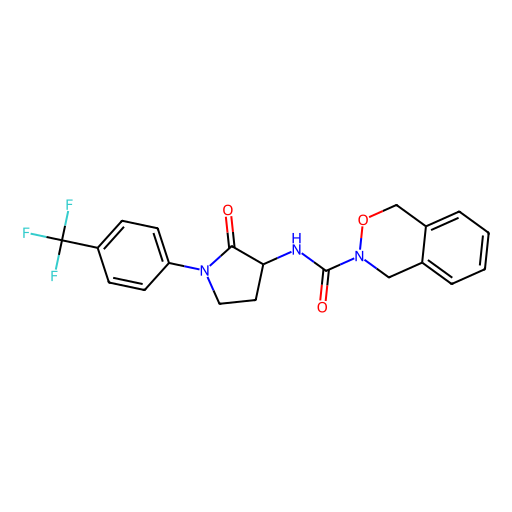}\end{subfigure}\hfill
    \begin{subfigure}{0.1\linewidth}\includegraphics[width=1.0\linewidth]{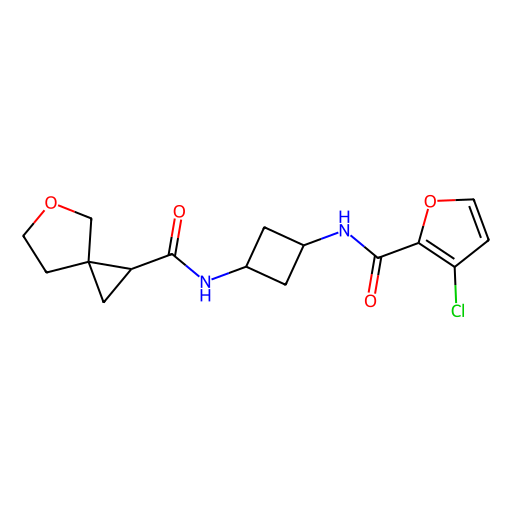}\end{subfigure}\hfill
    \begin{subfigure}{0.1\linewidth}\includegraphics[width=1.0\linewidth]{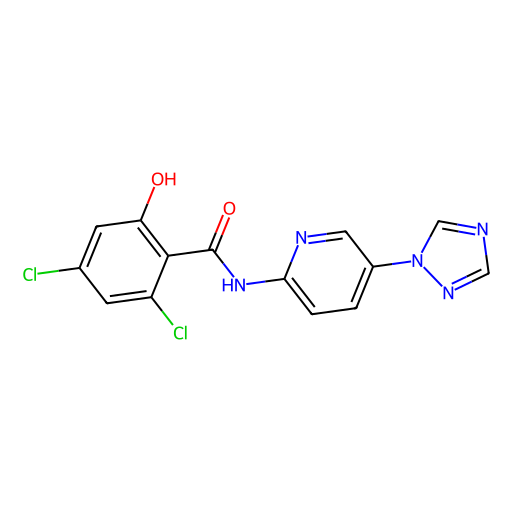}\end{subfigure}\hfill
    \begin{subfigure}{0.1\linewidth}\includegraphics[width=1.0\linewidth]{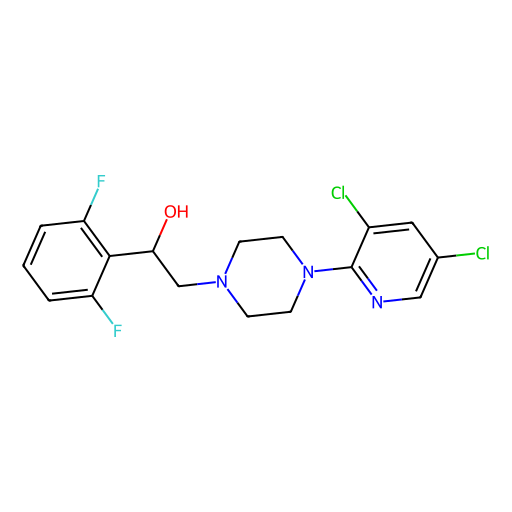}\end{subfigure}\hfill
    \begin{subfigure}{0.1\linewidth}\includegraphics[width=1.0\linewidth]{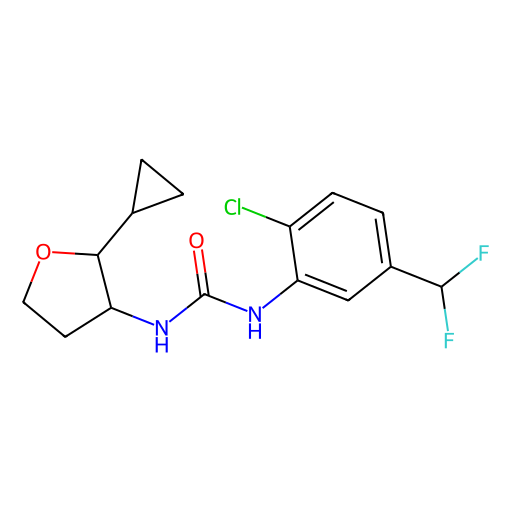}\end{subfigure}\hfill
    \begin{subfigure}{0.1\linewidth}\includegraphics[width=1.0\linewidth]{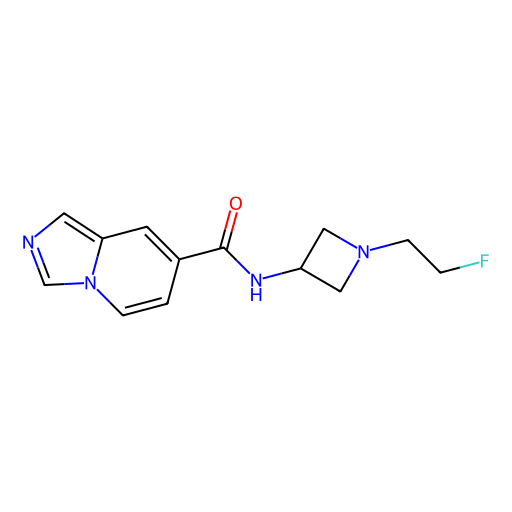}\end{subfigure}\hfill
    \begin{subfigure}{0.1\linewidth}\includegraphics[width=1.0\linewidth]{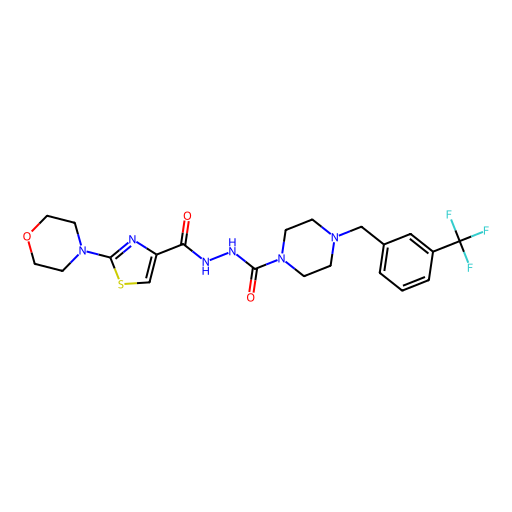}\end{subfigure}\hfill
    \begin{subfigure}{0.1\linewidth}\includegraphics[width=1.0\linewidth]{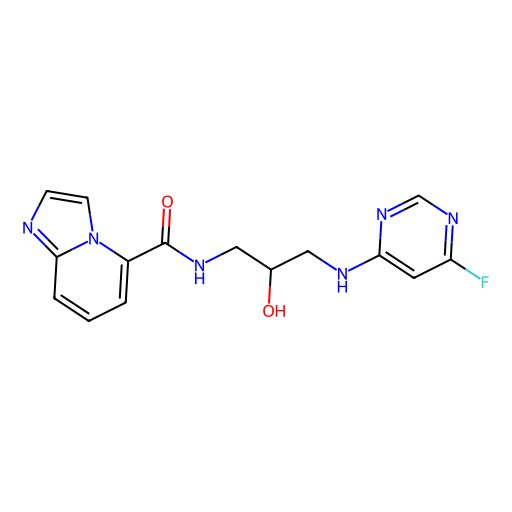}\end{subfigure}
    } \\
    \small{\textbf{(d) }{Cluster 3}} \\
    \resizebox{1.0 \columnwidth}{!}{
    \begin{subfigure}{0.1\linewidth}\includegraphics[width=1.0\linewidth]{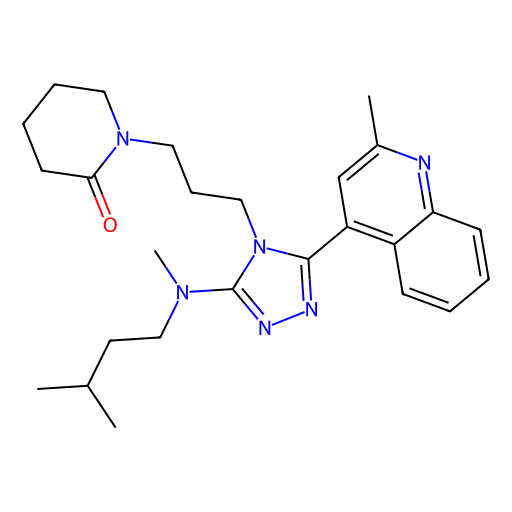}\end{subfigure}\hfill
    \begin{subfigure}{0.1\linewidth}\includegraphics[width=1.0\linewidth]{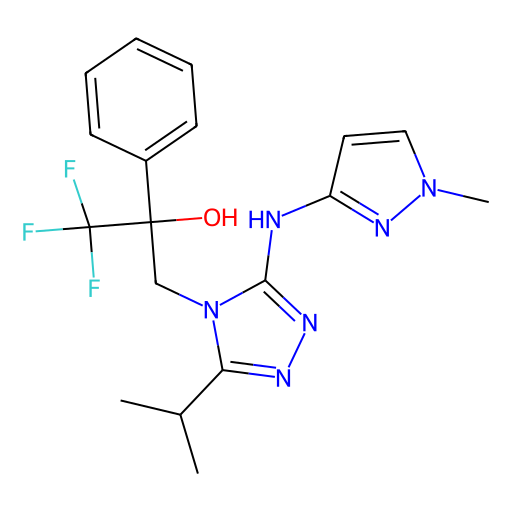}\end{subfigure}\hfill
    \begin{subfigure}{0.1\linewidth}\includegraphics[width=1.0\linewidth]{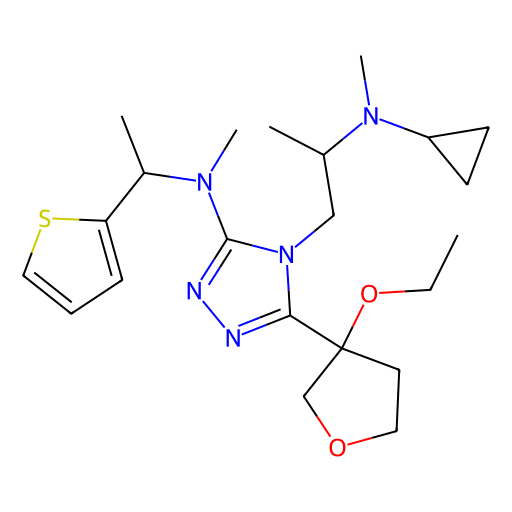}\end{subfigure}\hfill
    \begin{subfigure}{0.1\linewidth}\includegraphics[width=1.0\linewidth]{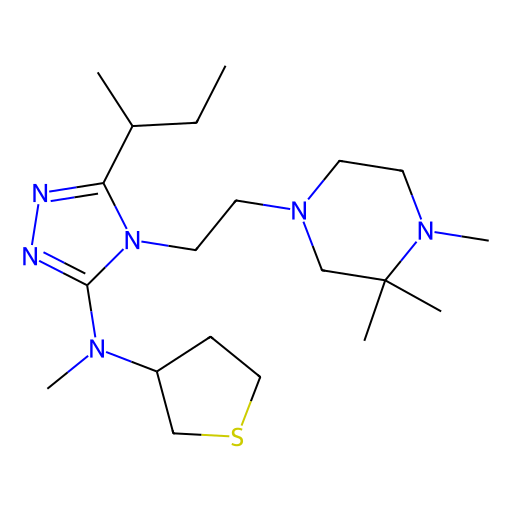}\end{subfigure}\hfill
    \begin{subfigure}{0.1\linewidth}\includegraphics[width=1.0\linewidth]{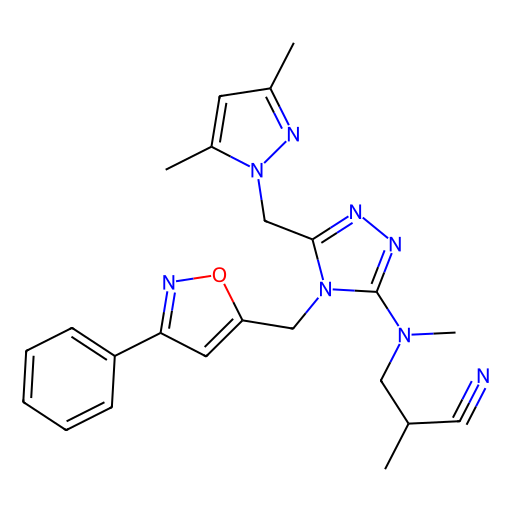}\end{subfigure}\hfill
    \begin{subfigure}{0.1\linewidth}\includegraphics[width=1.0\linewidth]{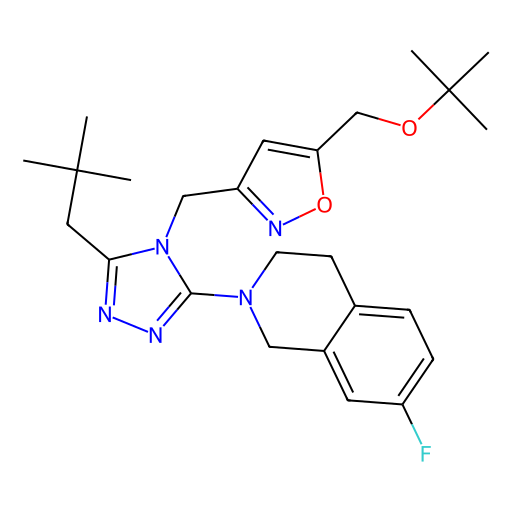}\end{subfigure}\hfill
    \begin{subfigure}{0.1\linewidth}\includegraphics[width=1.0\linewidth]{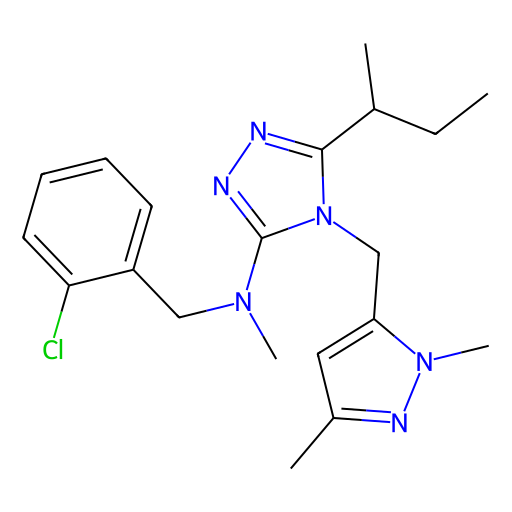}\end{subfigure}\hfill
    \begin{subfigure}{0.1\linewidth}\includegraphics[width=1.0\linewidth]{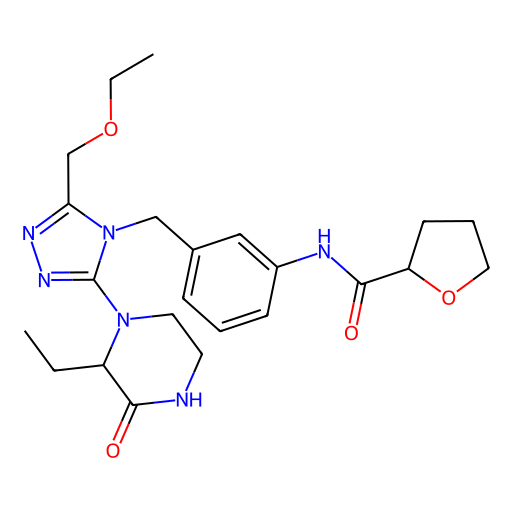}\end{subfigure}
    } \\
    \small{\textbf{(e) }{Cluster 4}} \\
    \resizebox{1.0 \columnwidth}{!}{
    \begin{subfigure}{0.1\linewidth}\includegraphics[width=1.0\linewidth]{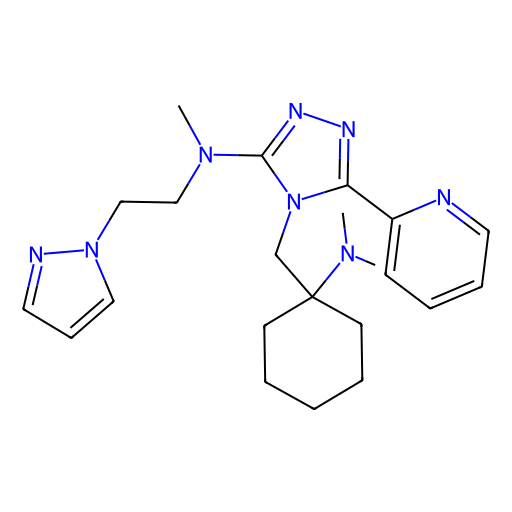}\end{subfigure}\hfill
    \begin{subfigure}{0.1\linewidth}\includegraphics[width=1.0\linewidth]{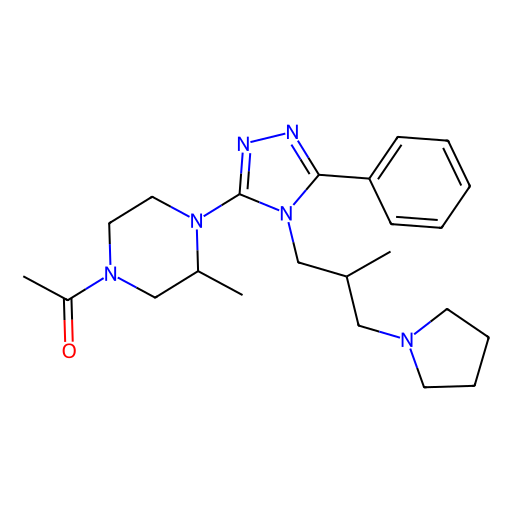}\end{subfigure}\hfill
    \begin{subfigure}{0.1\linewidth}\includegraphics[width=1.0\linewidth]{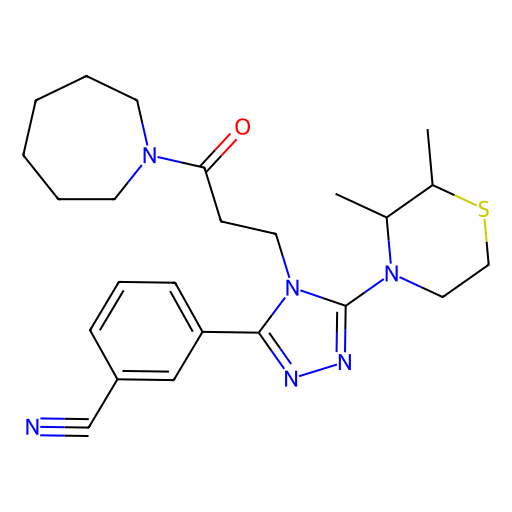}\end{subfigure}\hfill
    \begin{subfigure}{0.1\linewidth}\includegraphics[width=1.0\linewidth]{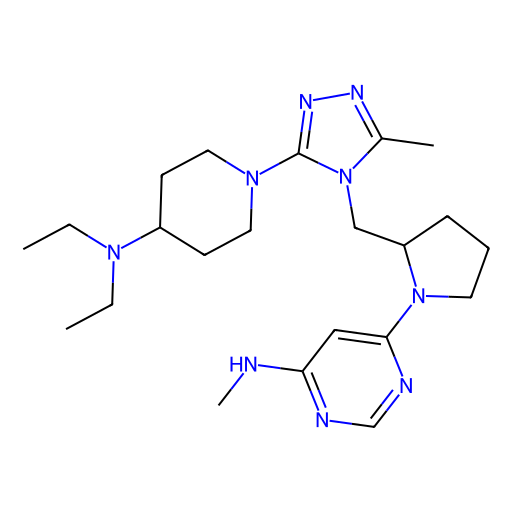}\end{subfigure}\hfill
    \begin{subfigure}{0.1\linewidth}\includegraphics[width=1.0\linewidth]{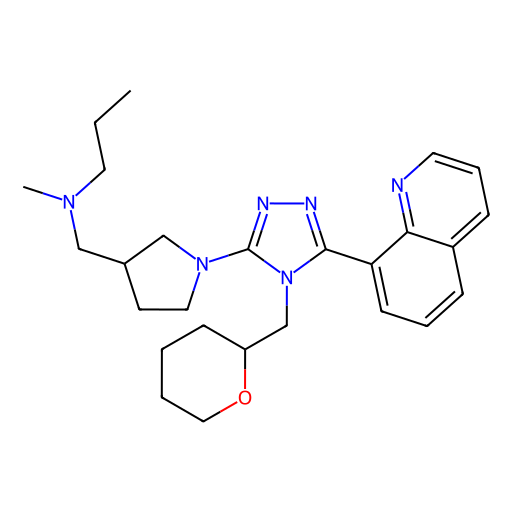}\end{subfigure}\hfill
    \begin{subfigure}{0.1\linewidth}\includegraphics[width=1.0\linewidth]{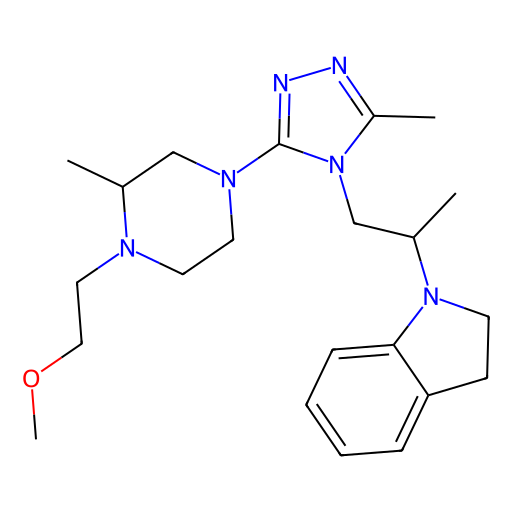}\end{subfigure}\hfill
    \begin{subfigure}{0.1\linewidth}\includegraphics[width=1.0\linewidth]{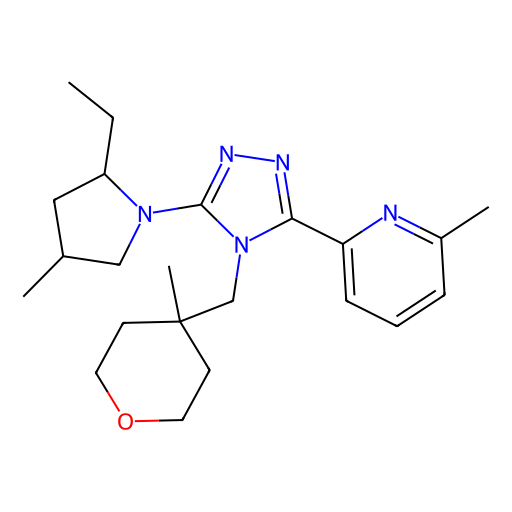}\end{subfigure}\hfill
    \begin{subfigure}{0.1\linewidth}\includegraphics[width=1.0\linewidth]{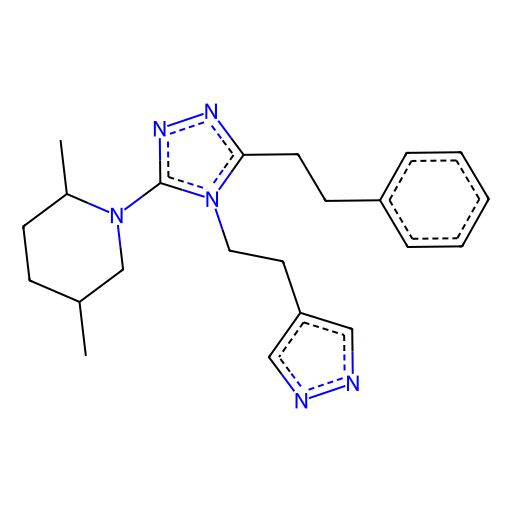}\end{subfigure}
    } \\
    \small{\textbf{(f) }{Cluster 5}} \\
    \resizebox{1.0 \columnwidth}{!}{
    \begin{subfigure}{0.1\linewidth}\includegraphics[width=1.0\linewidth]{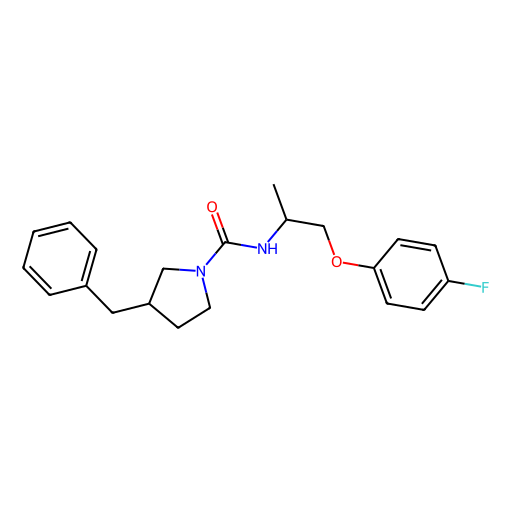}\end{subfigure}\hfill
    \begin{subfigure}{0.1\linewidth}\includegraphics[width=1.0\linewidth]{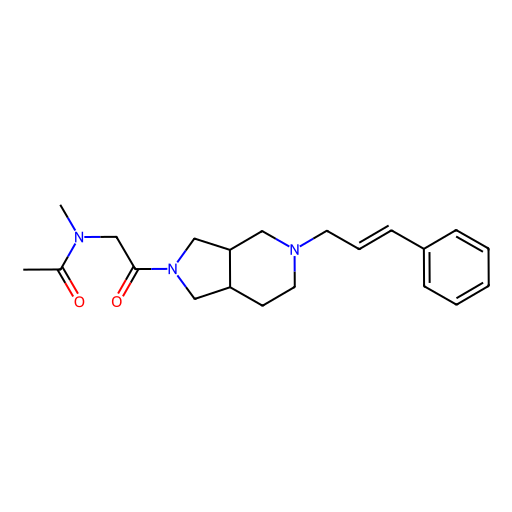}\end{subfigure}\hfill
    \begin{subfigure}{0.1\linewidth}\includegraphics[width=1.0\linewidth]{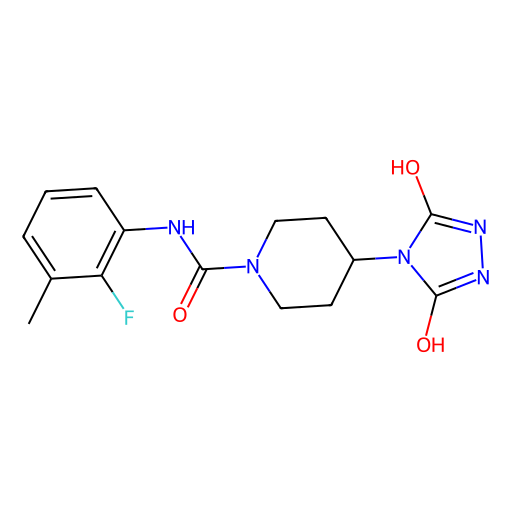}\end{subfigure}\hfill
    \begin{subfigure}{0.1\linewidth}\includegraphics[width=1.0\linewidth]{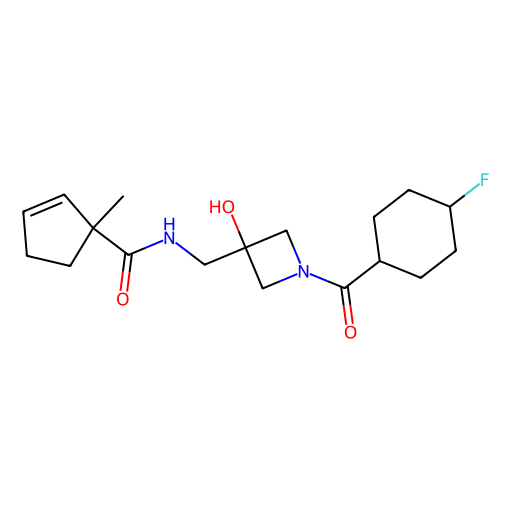}\end{subfigure}\hfill
    \begin{subfigure}{0.1\linewidth}\includegraphics[width=1.0\linewidth]{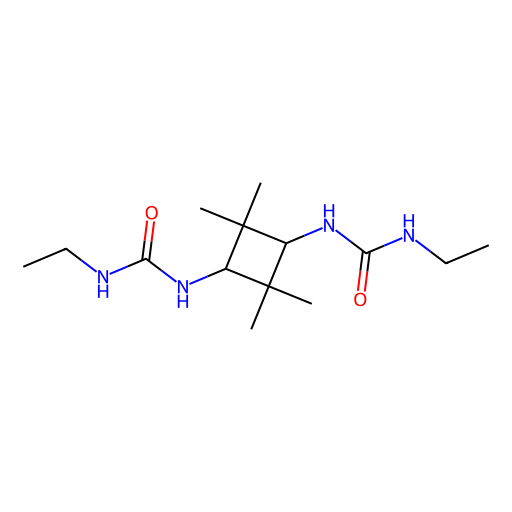}\end{subfigure}\hfill
    \begin{subfigure}{0.1\linewidth}\includegraphics[width=1.0\linewidth]{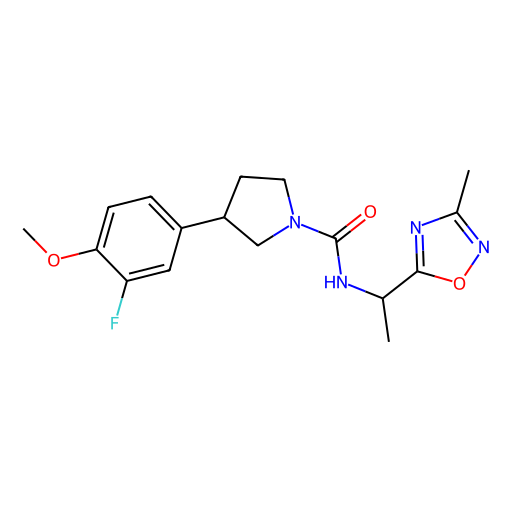}\end{subfigure}\hfill
    \begin{subfigure}{0.1\linewidth}\includegraphics[width=1.0\linewidth]{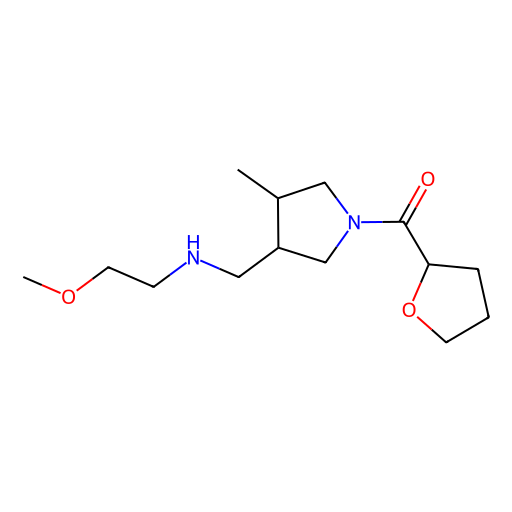}\end{subfigure}\hfill
    \begin{subfigure}{0.1\linewidth}\includegraphics[width=1.0\linewidth]{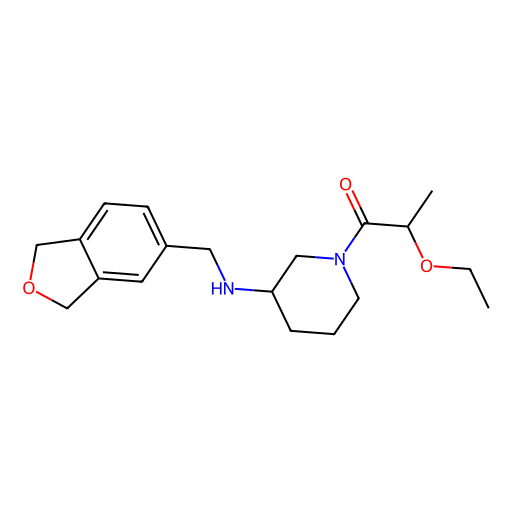}\end{subfigure}
    } \\
    \small{\textbf{(g) }{Cluster 6}} \\
    \resizebox{1.0 \columnwidth}{!}{
    \begin{subfigure}{0.1\linewidth}\includegraphics[width=1.0\linewidth]{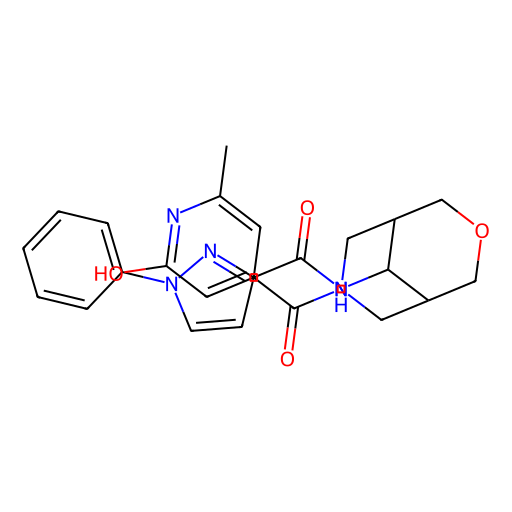}\end{subfigure}\hfill
    \begin{subfigure}{0.1\linewidth}\includegraphics[width=1.0\linewidth]{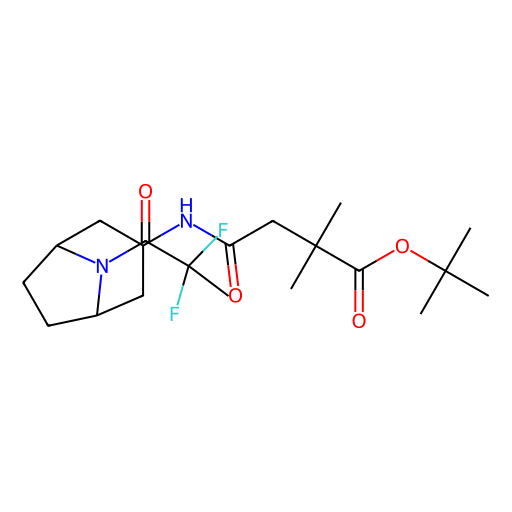}\end{subfigure}\hfill
    \begin{subfigure}{0.1\linewidth}\includegraphics[width=1.0\linewidth]{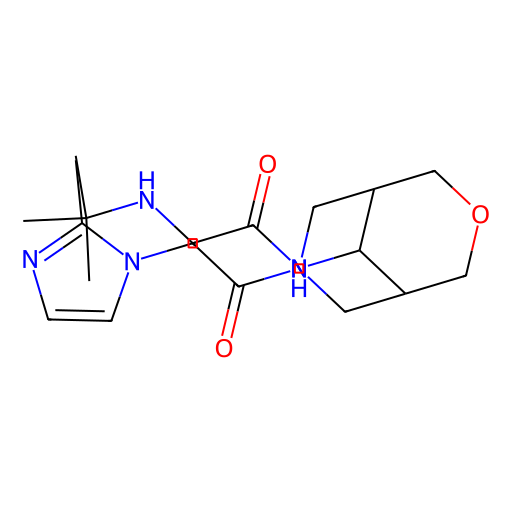}\end{subfigure}\hfill
    \begin{subfigure}{0.1\linewidth}\includegraphics[width=1.0\linewidth]{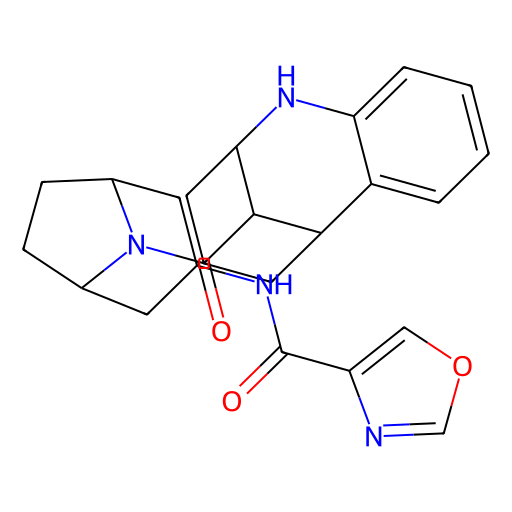}\end{subfigure}\hfill
    \begin{subfigure}{0.1\linewidth}\includegraphics[width=1.0\linewidth]{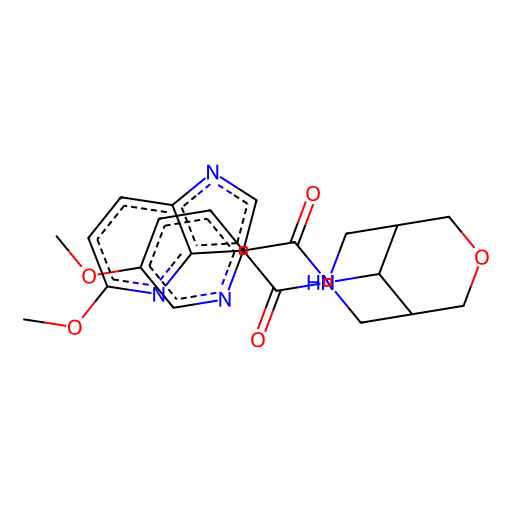}\end{subfigure}\hfill
    \begin{subfigure}{0.1\linewidth}\includegraphics[width=1.0\linewidth]{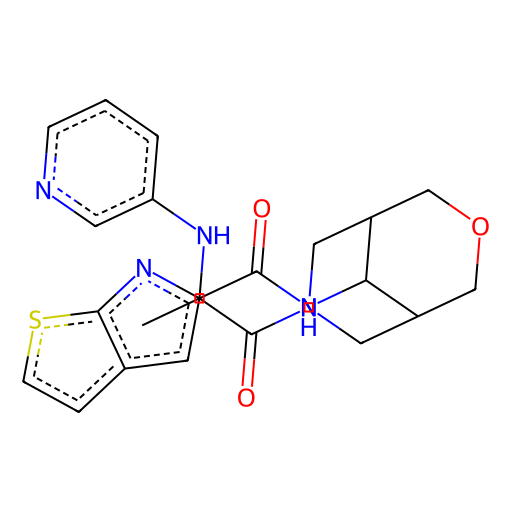}\end{subfigure}\hfill
    \begin{subfigure}{0.1\linewidth}\includegraphics[width=1.0\linewidth]{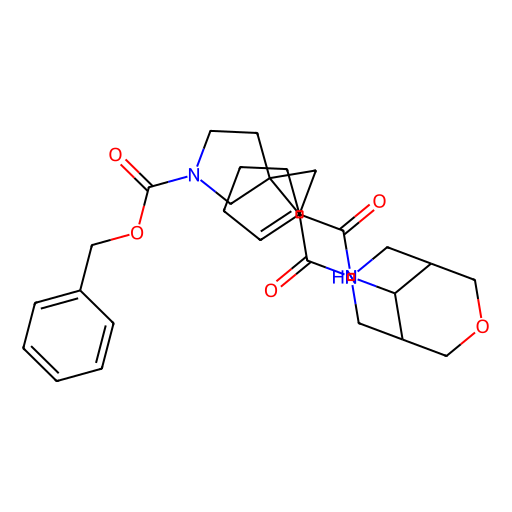}\end{subfigure}\hfill
    \begin{subfigure}{0.1\linewidth}\includegraphics[width=1.0\linewidth]{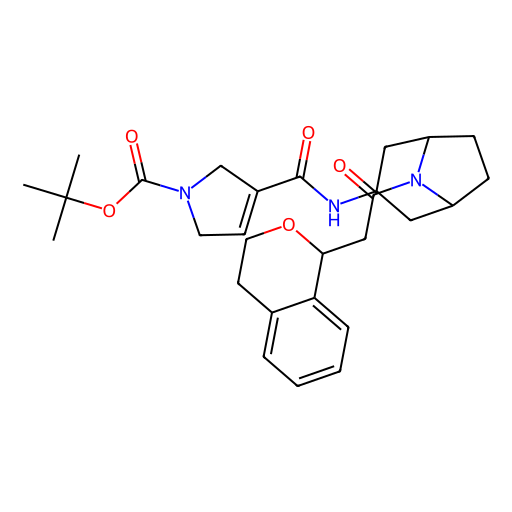}\end{subfigure}
    } \\
    \small{\textbf{(h)}{Cluster 7}} \\
    \caption{Visualization of the molecules in each cluster.}
    \label{fig:clustering_detail}
\end{figure}

Figures \ref{fig:clustering} and \ref{fig:clustering_detail} respectively illustrate the clustering results and the molecules within each cluster. An intriguing observation emerges from these results: the \encoder~model exhibits a propensity to cluster molecules with similar properties (e.g., identical functional groups in clusters 0, 1, 4, 5; similar structures in clusters 2, 3, 7; or similar Halogen atoms in cluster 3) within the latent \graphwords~space. This insight could potentially inform and inspire future research.

\subsection{Graph Translation}
\paragraph{Graph Interpolation.}
In exploit of the Euclidean representation of graphs, we explore the continuity of the latent \graphwords~using interpolation. Consider a source molecule $\gG_s$ and a target molecule $\gG_t$. We utilize \encoder~to encode them into \graphwords, represented as $\gW_s$ and $\gW_t$, respectively. We then proceed to conduct a linear interpolation between these two \graphwords, resulting in a series of interpolated \graphwords: $\gW^\prime_{\alpha_1}, \gW^\prime_{\alpha_2}, \dots, \gW^\prime_{\alpha_k}$, where each interpolated \graphword~is computed as $\gW^\prime_{\alpha_i} = (1-\alpha_i)\gW_s + \alpha_i\gW_t$. These interpolated \graphwords~are subsequently decoded back into molecules using \decoder.

\begin{figure}[htbp]
    \centering
    \begin{subfigure}{0.95\linewidth}
        \centering
        \includegraphics[width=1.0\linewidth]{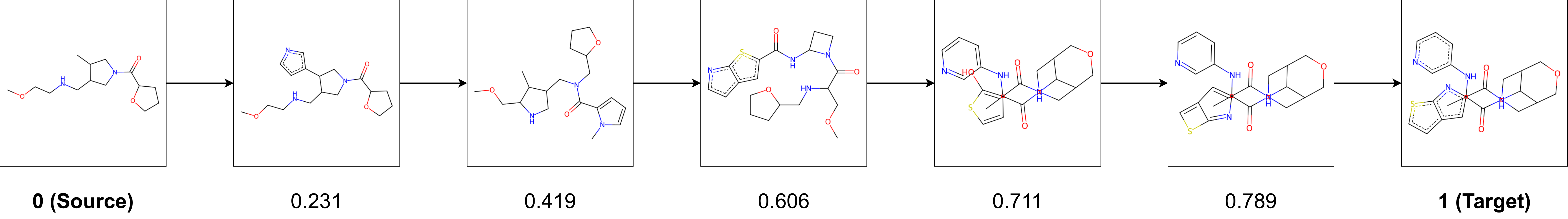}
        \caption{}
    \end{subfigure}
    \hfill
    \begin{subfigure}{0.95\linewidth}
        \centering
        \includegraphics[width=1.0\linewidth]{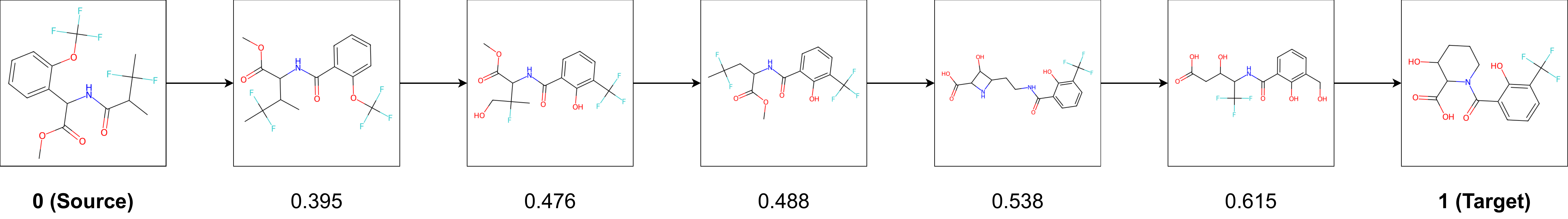}
        \caption{}
    \end{subfigure}
    \hfill
    \begin{subfigure}{0.95\linewidth}
        \centering
        \includegraphics[width=1.0\linewidth]{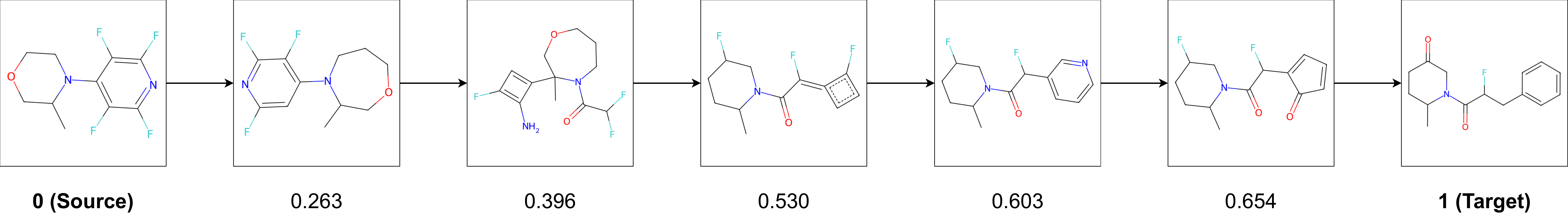}
        \caption{}
    \end{subfigure}
    \hfill
    \begin{subfigure}{0.95\linewidth}
        \centering
        \includegraphics[width=1.0\linewidth]{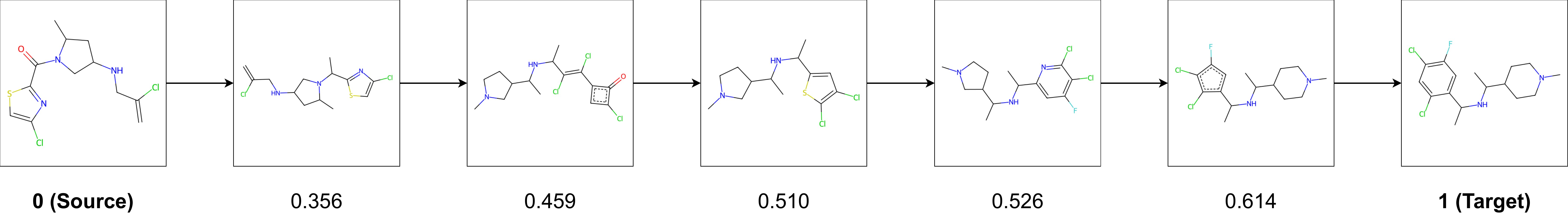}
        \caption{}
    \end{subfigure}
    \caption{Graph interpolation results with different source and target molecules using \graphsgpt-1W. The numbers denote the values of $\alpha$ for corresponding results.}
    \label{fig:interpolation}
\end{figure}

The interpolation results are depicted in Figure \ref{fig:interpolation}. We observe a smooth transition from the source to the target molecule, which demonstrates the model's ability to capture and traverse the continuous latent space of molecular structures effectively. This capability could potentially be exploited for tasks such as molecular optimization and drug discovery.

\paragraph{Graph Hybridization.}
With \encoder, a graph $\gG$ can be transformed into a fixed-length \graphword~sequence $\gW=[\vw_1, \cdots, \vw_k]$, where each \graphword~is expected to encapsulate distinct semantic information. We investigate the representation of \graphwords~by hybridizing them among different inputs. 

Specifically, consider a source molecule $\gG_s$ and a target molecule $\gG_t$, along with their \graphwords~$\gW_s=[{\vw_s}_1, \cdots, {\vw_s}_k]$ and $\gW_t=[{\vw_t}_1, \cdots, {\vw_t}_k]$. Given the indices set $I$ ,we replace a subset of source \graphwords~with the corresponding target \graphwords~${\vw_s}_i \leftarrow {\vw_t}_i, i \in I$, yielding the hybrid \graphwords~$\gW_h=[{\vw_h}_1, \cdots, {\vw_h}_k]$, where:

\begin{equation}
\vw_h=
\begin{cases}
    {\vw_t}_i,\quad i\in I \\
    {\vw_s}_i,\quad i\notin I
\end{cases}.
\end{equation}

\begin{figure}[h]
    \centering
    \includegraphics[width=0.8\linewidth]{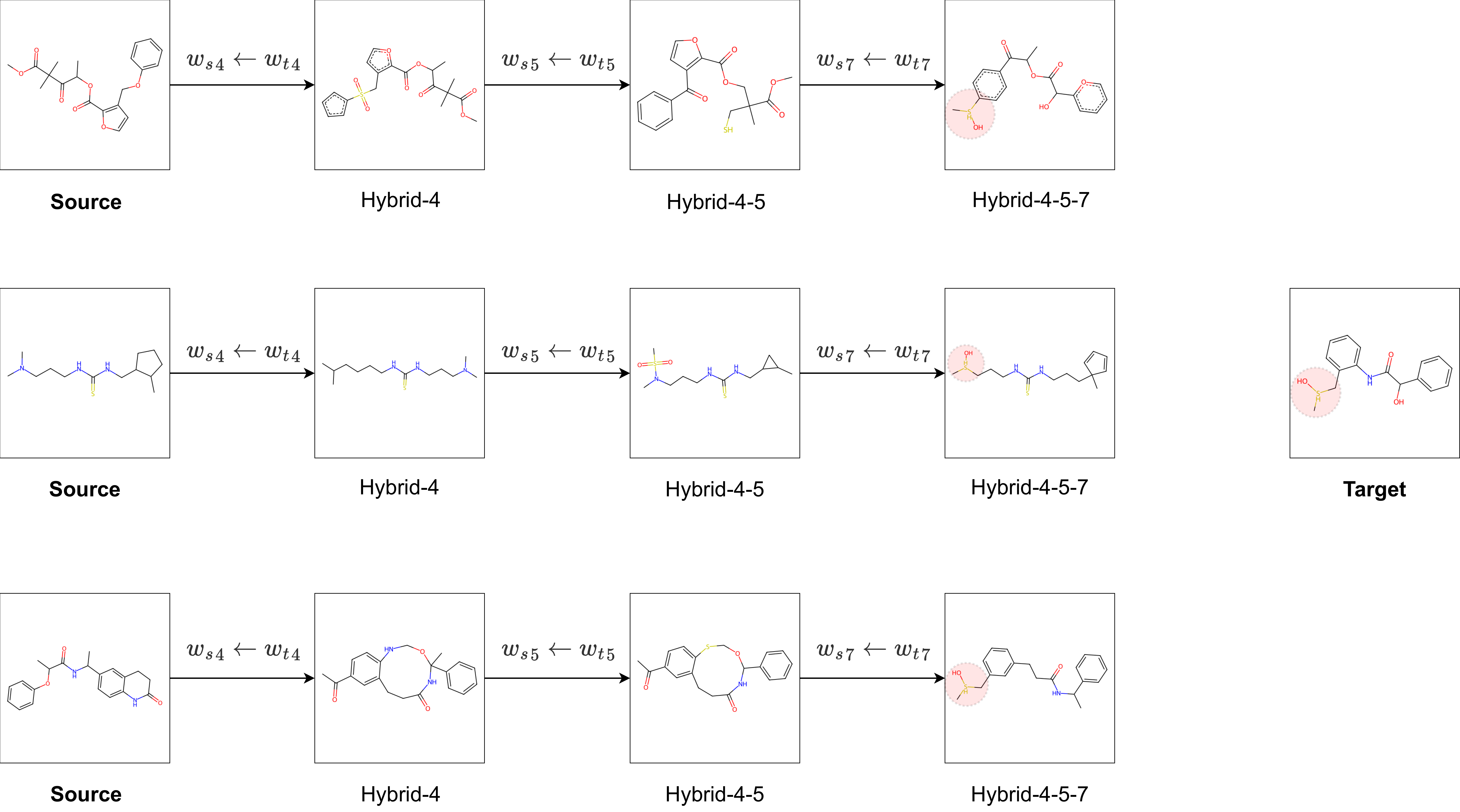}
    \caption{Hybridization results of \graphwords. The figure shows the changes in the source molecule after hybridizing specific \graphwords~from the target molecule. We use \graphsgpt-8W which has 8 \graphwords~in total.}
    \label{fig:hybrid}
\end{figure}

We then decode $\gW_h$ using \decoder~back into the graph and observe the changes on the molecules. The results are depicted in Figure \ref{fig:hybrid}. From these results, we observe that hybridizing specific \graphwords~can lead to the introduction of certain features from the target molecule into the source molecule, such as the Sulfhydryl functional group. This suggests that \graphwords~could potentially be used as a tool for manipulating specific features in molecular structures, which could have significant implications for molecular design and optimization tasks.

\end{document}